\pdfoutput=1

\documentclass[11pt]{article}

\usepackage[final]{acl}

\usepackage{times}
\usepackage{latexsym}
\usepackage{subfig}
\usepackage{stfloats}
\usepackage{listings}
\usepackage{multirow}
\usepackage{lscape}
\usepackage{amsmath} 
\usepackage{multicol}
\usepackage{multirow}
\usepackage{arydshln}
\newcommand{\squishlist}{
   \begin{list}{\small $\bullet$}
    { \setlength{\itemsep}{0pt}      \setlength{\parsep}{2pt}
      \setlength{\topsep}{1pt}       \setlength{\partopsep}{1pt}
     \setlength{\leftmargin}{1.2em} \setlength{\labelwidth}{1em}
      \setlength{\labelsep}{0.5em} } }
\newcommand{\squishend}{  \end{list}  }

\usepackage[T1]{fontenc}

\usepackage[utf8]{inputenc}

\usepackage{microtype}

\usepackage{inconsolata}

\usepackage{graphicx}

\title{How Does the Disclosure of AI Assistance Affect\\ the Perceptions of Writing?}

\author{
Zhuoyan Li\textsuperscript{1}, Chen Liang\textsuperscript{2}, Jing Peng\textsuperscript{2}, Ming Yin\textsuperscript{1} \\
\textsuperscript{1}Purdue University \\
\textsuperscript{2}University of Connecticut  \\
\texttt{\{li4178, mingyin\}@purdue.edu}\\ \texttt{\{chenliang, jing.peng\}@uconn.edu} \\
}

\begin{document}
\maketitle
\begin{abstract}
Recent advances in generative AI technologies like large language models have boosted the incorporation of AI assistance in writing workflows, leading to the rise of a new paradigm of human-AI co-creation in writing. To understand how people perceive writings that are produced under this paradigm, in this paper, we conduct an experimental study to understand whether and how the disclosure of the level and type of AI assistance in the writing process would affect  people's perceptions of the writing on various aspects, including their evaluation on the quality of the writing and their ranking of different writings.
Our results suggest that disclosing the AI assistance in the writing process, especially if AI has provided assistance in generating new content, decreases the average quality ratings for both argumentative essays and creative stories. This decrease in the average quality ratings often comes with an increased level of variations in different individuals' quality evaluations of the same writing. Indeed, factors such as an individual's writing confidence and familiarity with AI writing assistants are shown to moderate the impact of AI assistance disclosure on their writing quality evaluations. We also find that disclosing the use of AI assistance may significantly reduce the proportion of writings produced with AI's content generation assistance among the top-ranked writings. 
\end{abstract}

\section{Introduction}
Recent advances in generative artificial intelligence (AI) technologies like large language models (LLMs) have significantly expanded the scope and depth of tasks where AI tools can collaborate with humans. For instance, GPT-4~\cite{openai2023gpt4} have demonstrated remarkable abilities in language understanding and generation, highlighting the potential of incorporating LLM-powered assistants in human writing workflows to enhance their productivity and creativity~\cite{noy2023experimental,clark2018creative}. Unlike traditional writing support systems that only offer basic grammar and spelling checks, LLM-powered writing assistants can provide a diverse range of support to human writers, ranging from polishing human-written texts to drafting written content from scratch, all tailored to human instructions.  
As such, a new paradigm of human-AI co-creation in writing has emerged, and we expect to see a growing amount of writings that are generated with some degree of AI assistance in the next many years to come. 

A critical question that arises with the emergence of the human-AI co-writing paradigm is how to present writings appropriately to readers in the future. While the disclosure of AI assistance in the writing process has been advocated for transparency and accountability considerations,  empirical understandings on the implications of AI assistance disclosure on 
people's perceptions of writings are still lacking. For example, does the disclosure of AI assistance during the writing process change people's evaluation on the quality of the writings? In other words, in an era of human-AI collaboration, will people
evaluate the quality of writings solely based on the final written content, or also consider the generation process of the writings?
If disclosing the involvement of AI assistance does affect people's perceptions of writing quality, does the effect vary with the type of AI assistance 
and writing task? How will the disclosure of AI assistance change people's ranking among writings that are produced with different levels of AI assistance? 

To answer these questions, in this paper, we present an experimental study  to understand how the disclosure of AI assistance 
influences people's perceptions of writings on various aspects. 
Our study was divided into two phases. In Phase 1, we collected writing samples on two types of tasks (i.e., argumentative essay or creative story), and during their writing process, the human writers used different levels or types of AI assistance powered by ChatGPT, a state-of-the-art LLM. Specifically, participants were recruited to complete their writing tasks under one of the three writing modes: 
(a) \textbf{{\em Independent}}, where participants completed the writing task on their own without receiving any assistance from ChatGPT; (b) \textbf{{\em AI editing}}, where participants took the primary responsibility of drafting
the article while they received only text editing and polishing assistance from ChatGPT; and (c) \textbf{{\em AI (content) generation}}, where ChatGPT took the primary responsibility of drafting the initial
version of the article while participants provided feedback and directed the subsequent revisions of the article through conversational interactions with ChatGPT. After 
collecting writing samples, in Phase 2, we designed a randomized experiment in which participants were recruited to review writing samples collected from Phase 1 on various aspects (e.g., overall quality, organization, creativity). Depending on the 
treatment that participants were assigned, the level and type of AI assistance during the writing process was either {\em disclosed} or {\em not disclosed} to them when they reviewed these writings. 

Our experimental results indicate that disclosing AI assistance in the writing process, especially when AI provides content generation assistance as that in the ``AI generation'' writing mode, significantly decreases people's evaluation on the overall quality of the writing for both argumentative essays and creative stories. Similar patterns have been observed on people's willingness to shortlist a piece of writing for performance-based bonus, and people's detailed evaluations on the writing on aspects like organization and originality. The disclosure of AI assistance often increases the variations in quality evaluations provided by different individuals for the same writing as well. This indicates that with the knowledge of the AI involvement in the writing process, the quality evaluations of writings become unpredictable and highly susceptible to variability depending on who evaluates them.  Further examinations 
reveal that factors such as an individual's own writing confidence and their familiarity with ChatGPT might play a role in moderating the effects of AI assistance disclosure on people's evaluation of the writing quality. For instance, individuals with higher confidence in their writing skills are more likely to lower their quality ratings when AI use is disclosed, compared to those with lower writing confidence.  Moreover, we find that disclosing AI assistance would also significantly reduce the proportion of top-ranked writings produced with AI's content generation assistance, but this is only true for argumentative essays and not for creative stories. 

Together, our study offers important experimental evidence regarding the impact of the disclosure of AI use on human perceptions of the human-AI co-created content.
We conclude by discussing the design implications of our findings, and outline limitations and future work.

\section{Related Work}
\paragraph{\textbf{Assisting human writing using large language models.}} Large language models (LLMs)~\cite{brown2020language}, a specific type of generative AI technologies, have demonstrated exceptional abilities in language understanding and generation~\cite{openai2023gpt4,li2023synthetic}. This opens up exciting possibilities of actively incorporating LLMs into human writing processes to enhance human productivity and creativity~\cite{noy2023experimental,clark2018creative,wasi2024ink,wasi2024llms,li2024value,lee2024design,lee2022evaluating,jiang2023personallm,piller2023ethics}. A recent line of research has investigated into the range of assistance that LLMs can offer to humans during their writing~\cite{10.1145/3637875}. They found that LLMs can provide assistance in diverse areas such as creative content  ideation~\cite{10.1145/3586183.3606800,suh2023structured}, 
tailored content generation and completion~\cite{10.1145/3544548.3580969,10.1145/3490099.3511105,buschek2021impact}, and  advanced text revisions beyond the traditional grammar or spelling checks~\cite{10.1145/3490099.3511105}. LLM-powered writing assistants can also be applied to different writing tasks including stories~\cite{chung2022talebrush,10.1145/3490099.3511105,singh2023hide,yang2022ai,clark2018creative,lee2022coauthor}, advertisement slogans~\cite{chen2023large,clark2018creative}, argumentative essays~\cite{lee2022coauthor}, emails~\cite{buschek2021impact,fu2023comparing}, and scripts~\cite{mirowski2023co}.

\paragraph{\textbf{Human-centered evaluations of texts generated or edited by large language models.}}
With their unparalleled generative capabilities, LLMs can rapidly generate a large volume of texts. 
Recent research has started to evaluate how humans perceive LLM-generated texts. 
For example, it was found that distinguishing LLM-generated texts from human-generated texts is quite challenging, in terms of both objective textual patterns such as text fluency and perplexity~\cite{ippolito2019automatic}, and subjective human judgment~\cite{dou2021gpt,clark2021all}. On
online platforms like Airbnb, it was found that AI-generated profiles may be perceived as less trustworthy by users~\cite{jakesch2019ai}.
Interestingly, recent research has also showed that when LLMs are used to assist humans in writing, the LLM-generated texts can impact the human writer, consciously or unconsciously.
For instance,
one study found that text suggestions provided by LLMs can change the sentiment of human-generated
texts~\cite{hohenstein2020ai}. In addition, writing assistants powered by opinionated LLMs can not only alter the opinions expressed in
the human-generated text but also subtly influence the perspectives and beliefs of the human writers themselves~\cite{jakesch2023co}. 
Different from these prior works, in this paper, we focus on examining whether and how the {\em disclosure} of varying types of assistance from LLMs would affect people's perceptions of writings.


\section{Study Design}
To understand how the disclosure of AI assistance would influence people's perceptions of writings, 
we divide our study into two phases. In the first phase, we collect writing samples on different types of writing tasks, while the human writers use different levels or types of AI assistance during their writing processes. The second phase is our focal experiment, in which we recruit participants to evaluate the writing samples collected. They will be randomly assigned to groups informed or uninformed about AI assistance used in the writing process and the specific type of assistance used. This study was approved by the Institutional Review Board of the authors' institution.

\subsection{Phase 1: Collection of Writing Samples}

In preparation for our focal experiment in Phase 2, we first undertake our Phase 1 study to collect humans' writing samples when they complete their writing tasks with different levels and types of AI assistance.  

\paragraph{Writing tasks.}
Participants in Phase 1 were asked to write a 200--250 word article within 45 minutes. As different writing tasks 
vary on their purposes and 
required skills, we asked our participants to work on one of the two types of writing tasks:
\squishlist
\item \textbf{Argumentative essay writing}: Participants were provided with a statement that was randomly sampled from a set of statements used in the TOEFL writing exam topic pool (e.g., ``{\em Nowadays it is easier to maintain good health than it was in the past.}''). Participants were asked to write an essay to discuss the argument in the statement. In their essays, participants could either support or oppose the argument in the statement. 

\item \textbf{Creative story writing}: Participants were given a prompt (e.g., {\em Someone saying ``Let's go for a walk.''}), and they were asked to write a story that includes the prompt. Prompts we used in these tasks were adapted from the popular Reedsy’s Short Story Contest\footnote{https://blog.reedsy.com/creative-writing-prompts/terms/.}.
\squishend

\paragraph{Writing modes.} To reflect varying degrees and types of assistance writers may receive from a state-of-the-art LLM, ChatGPT\footnote{We used the GPT-3.5-turbo model in our study.}, we considered three different writing modes in Phase 1: 
\squishlist
    \item \textbf{Independent}: In this mode, participants completed the writing task independently without any assistance from ChatGPT.
    
    \item \textbf{AI editing}: In this mode, participants were primarily responsible for writing the article. Meanwhile, participants could send any part of their drafts to ChatGPT for editing and polishing, and then they could decide how to integrate the polished texts into their writing. Note that in this mode, we configured ChatGPT in a way such that it can only provide editing assistance to participants and can not generate new content from scratch.
    
    \item \textbf{AI (content) generation}: In this mode, ChatGPT took the lead in drafting the initial version of the article. Participants could then provide feedback and direct the subsequent revisions of the article through conversational interactions with ChatGPT. In the end, participants decided how to compose the final article based on different versions of the drafts that ChatGPT generated, and they also had the option to incorporate some of their own writing into it.
\squishend

\paragraph{Procedure.}
We opened our Phase 1 study to U.S. workers whose primary language is English on Prolific, and each worker was only allowed to participate in this study once. Upon arrival, participants were first asked to report some demographic information (e.g., gender, age, confidence in various writing tasks). 
Then, participants were randomly assigned to one of the two writing tasks (i.e., argumentative essay or creative story) and one of the two conditions: ``{\em Independent vs. AI editing}'' or ``{\em Independent vs. AI generation}.'' Each condition includes two writing modes at different levels of compensation. 
Participants were required to choose a preferred writing mode, 
and complete their writing task under that writing mode\footnote{Note that for some participants, the writing mode with AI assistance came with higher payments than independent writing, while for others the payments for independent writing were higher. We collected human writers' writing samples by asking them to select into their preferred writing modes, because Phase 1 data was used to estimate the value that people attach to different types of AI assistance, which has been reported in~\citeauthor{li2024value}~\shortcite{li2024value}. Due to this design, in our analysis, we focus on the impacts of AI assistance disclosure on raters' perceptions of writings 
across different scenarios. However, we do not intend to draw any causal conclusions based solely on the direct comparisons of 
raters' perceptions of writings 
across the three writing modes. \label{design_footnote}}. 
After completing the writing task,
participants were asked to fill out an exit survey to report their familiarity and usage frequency of ChatGPT, their future confidence in writing under the chosen mode, and their perceptions of their writing experience. To filter out inattentive participants, we included two attention check questions in our study. The participants must pass both attention checks for their writings to be used in our subsequent Phase 2 study.

\begin{table}[t]
\resizebox{\linewidth}{!}{
\begin{tabular}{cccc}
\hline
                    & \textbf{Independent} & \textbf{AI editing} & \textbf{AI generation} \\ \hline
\textbf{Argumentative essay} & 89          & 49       & 68         \\
\textbf{Creative story}      & 69          & 58       & 74         \\ \hline
\end{tabular}
}
\caption{The number of articles collected in Phase 1 for each writing mode across the two types of writing tasks.}
\label{count}
\vspace{-10pt}
\end{table}

\paragraph{Data Collection Results.}
In total, 407 Prolific workers successfully completed our Phase 1 study and passed the attention checks. On average, participants spent 27.1 minutes on the study and received an average hourly payment of \$13.9. Table~\ref{count} shows the
number of articles we collected from Phase 1 for each writing mode across the two types of writing tasks. For more details about the design of our Phase 1 study, please see Appendix~\ref{design1}.

\subsection{Phase 2: Experimental Design}

After collecting writing samples that human writers generated using different levels and types of AI assistance, in Phase 2, we designed a randomized experiment to evaluate how people perceive these writing samples, when the usage of AI assistance during the writing process was or was not revealed.  

Specifically, in our Phase 2 study, a separate set of participants (distinct from those in Phase 1) were each asked to review a random set of six articles that we collected from Phase 1. Participants were told that the articles they needed to review were submitted by crowd workers in a previous study. For each article, participants first evaluated its overall quality on a scale from 1 to 5 stars, with a granularity of 0.5, and indicated whether they were willing to shortlist the article for granting its author an additional performance-based bonus. They were then required to justify their evaluations in a few sentences. Following this, participants provided their detailed evaluation of this article on five specific aspects, including the article's (a) grammar and vocabulary, (b) organization, (c) originality, (d) creativity, and (e) emotional authenticity. Each aspect was again evaluated on a scale from 1 to 5 stars.  We considered two experimental treatments in Phase 2:
\squishlist
    \item \textbf{Non-Disclose}: In this treatment, participants evaluated their assigned articles without being informed about whether or how the authors of these articles used AI assistance during their writing.

    \item\textbf{Disclose}: In this treatment,
    we informed participants about whether and how AI assistance was used by the author when writing the article, before they were asked to evaluate it. For example, when the author of an article completed their writing under the {\em AI generation} mode, our Phase 2 participants were told that ``the draft of this article was generated by ChatGPT, and the crowd worker had prompted ChatGPT to revise and improve the content'' before they started to evaluate this article. 
\squishend

\subsection{Phase 2: Experimental Procedure}
Our Phase 2 study was open to U.S. workers only on Prolific. We excluded the workers who had participated in our Phase 1 study, and each worker was allowed to only take this Phase 2 study once. Each participant went through a few steps in the study, as detailed below.
\paragraph{\textbf{Background assessment.}}
Upon arrival, participants were first asked to fill out a questionnaire to report their demographic information (e.g., gender, age, education, and race). We then asked  participants to indicate how confident they were in their own writing skills and how often they engaged in writing activities on a 5-point Likert scale from 1 (very low) to 5 (very high). 

\paragraph{\textbf{Main rating tasks.}} Subsequently, participants were randomly assigned into one of the two experimental treatments, {\em Disclose} or {\em Non-Disclose}. They then needed to evaluate six articles sequentially. Each article was randomly sampled from the pool of articles we collected in our Phase 1 study, although 
for participants in the ``{\em Disclose}'' treatment, we also ensured that all six articles were generated under the same writing mode. 
In other words, participants in the ``{\em Disclose}'' treatment were limited to evaluating articles from the same writing mode, thereby mitigating the potential interference between different writing modes in the evaluation. As described earlier, for each article, participants needed to provide evaluation on its overall quality and on five specific aspects, and they could also choose to shortlist the article for bonus.

\paragraph{\textbf{Exit Survey.}} After the completion of the main rating tasks, participants were asked to complete an exit survey. In this survey, participants needed to report their familiarity with ChatGPT and their frequency of using ChatGPT on a 5-point Likert scale. We also included a few questions to understand participants' perceived authorship of the final articles to the human authors. 
Similar as that in Phase 1, we also included one attention check question in our Phase 2 study. Only the data collected from participants who passed the attention check in Phase 2 were considered as valid, and would be used for the analysis.

\section{Results}
In total, 786 workers from Prolific took our Phase 2 study and passed the attention check ({\em Non-Disclose}: 380, {\em Disclose}: 406; see Appendix~\ref{demographics} for participants' demographic backgrounds). 
The average amount of time participants spent on the Phase 2 study was 20 minutes, resulting in an average hourly wage of \$7.5.  
On average, each article received 5.6 evaluations from participants in the ``{\em Non-Disclose}'' treatment and 5.9 evaluations from participants in the ``{\em Disclose}'' treatment. 
Below, we analyze the impact of disclosing AI assistance in the writing process on people's assessment of writings 
based on the data we collected in Phase 2. Due to the space limit, we focus on sharing the results on the impact of AI disclosure on the perceived quality and ranking of articles in the main text. See Appendix~\ref{sec:authorship} for the impact of AI disclosure on the perceived authorship of articles.

\begin{figure}[t]
  \centering
  \subfloat[Argumentative essay]{\includegraphics[width=0.24\textwidth]{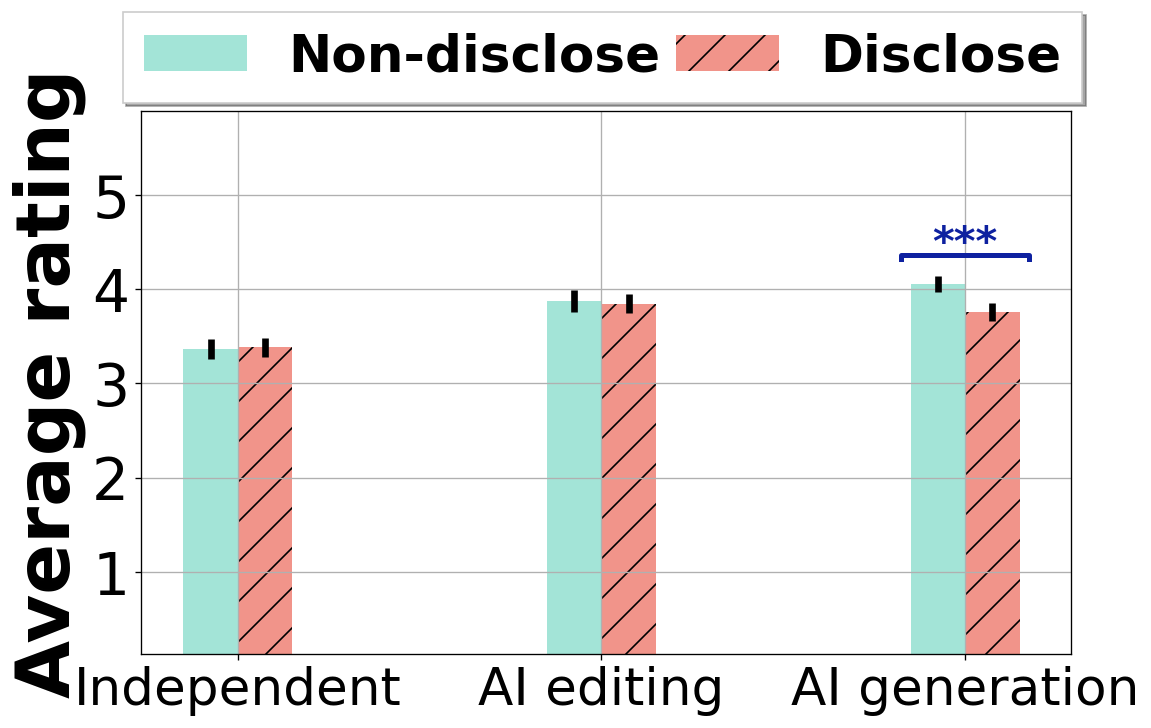}\label{fig:rating_statement}}
  \hfill
  \subfloat[Creative story]{\includegraphics[width=0.24\textwidth]{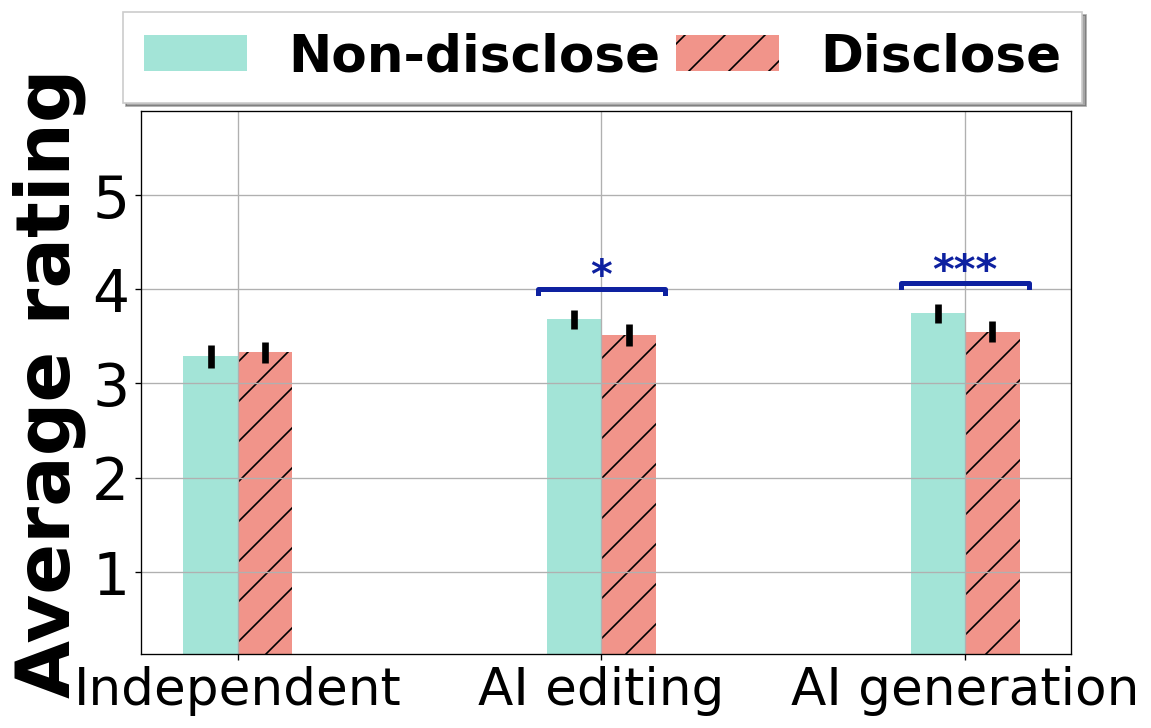}\label{fig:rating_story}}
  \vspace{-5pt}
  \caption{Comparing {\em average} ratings of the overall quality of articles generated under the independent, AI editing, or AI generation writing modes, with and without disclosure of the use and type of AI assistance during the writing process. Error bars represent the 95\% confidence intervals of the mean values. $\textsuperscript{*}$ and $\textsuperscript{***}$ denote significance levels of $0.05$ and $0.001$, respectively.}
  \label{rating_treatment}
  \vspace{-15pt}
\end{figure}

\subsection{Impacts of Disclosing AI Assistance on the Quality Evaluation of Writing}
\label{sec:avg}

Participants' average ratings on the overall quality of the argumentative essays and creative stories, grouped by writing modes, are presented in Figures~\ref{fig:rating_statement} and~\ref{fig:rating_story}, respectively. Visually, it appears that when the articles are written by humans independently without using any AI assistance, whether knowing this information or not does not significantly change people's perceived quality of the articles. In contrast, when the writers use some degree of AI assistance in their writing process, the disclosure of this information often decreases people's perceived quality of the articles. 

To examine whether these differences are statistically significant, we conducted regression analyses. Specifically, the focal independent variable was the treatment that a participant was assigned to in our Phase 2 study (i.e., {\em Disclose} vs. {\em Non-Disclose}), while the dependent variable was the participant's rating on the overall quality of an article. This regression analysis was done separately for articles generated under each of the three writing modes. To minimize the impact of potential confounding variables, we also accounted for a set of covariates in our regressions, such as the demographic background of both the Phase 1 article writers (e.g., age, gender, the payment received for the selected writing mode, confidence in the assigned writing task, etc.) and the Phase 2 raters (e.g., age, gender, general writing confidence, writing activity frequency, 
frequency of ChatGPT use, 
etc.).

Our regression results suggest that for articles produced under the ``{\em AI generation}'' writing mode, informing people about the inclusion of ChatGPT's content generation assistance during the writing process significantly decreases people's perceived quality of the writing, for both argumentative essays and creative stories ($p<0.001$).  However, when ChatGPT only provides editing assistance to humans during the writing process, the disclosure of AI's editing assistance only significantly decreases people's perceived quality of creative stories ($p=0.011$) but not that of argumentative essays.  We found similar patterns when examining how the disclosure of AI assistance affects people's willingness to shortlist an article,  and people's detailed evaluations on the article's grammar and vocabulary, organization, originality, creativity, and emotional authenticity. For more details, please see Appendix~\ref{quality_sm}.

\subsection{Impacts of Disclosing AI Assistance on the Variation in the Quality Evaluation 
}
\label{sec:var}

In Section~\ref{sec:avg}, we have found that the disclosure of AI assistance in the writing process may lead to a decrease in the average rating of writing quality. To understand whether it also affects the dispersion in quality evaluation, we examine how the disclosure of AI assistance affects the {\em variance} in ratings among different participants for the same article. 
Specifically, for each of the 407 articles used in Phase 2, given a particular aspect of evaluation (e.g., overall quality), we gathered all the ratings on this aspect for this article, and then computed the variance within the ratings provided by participants in the {\em Disclose} and {\em Non-Disclose} treatments, separately.

 \begin{figure}[t]
  \centering
  \subfloat[Argumentative essay]{\includegraphics[width=0.24\textwidth]{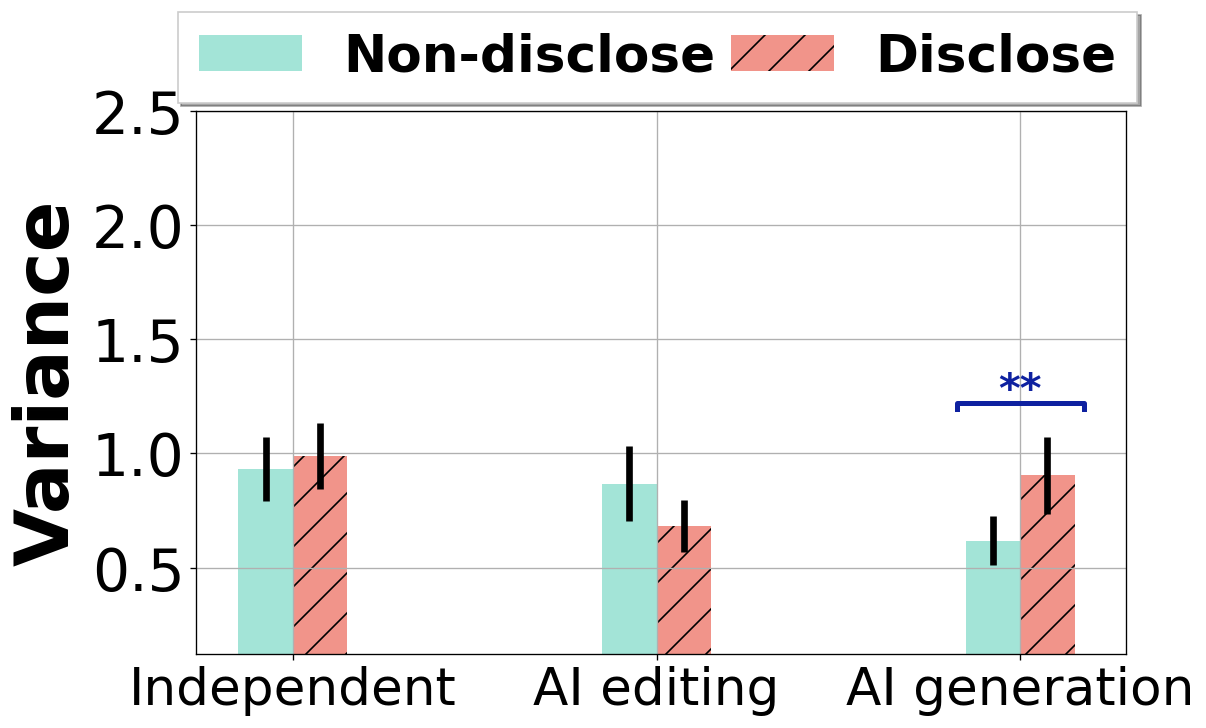}\label{fig:variance_statement}}
  \hfill
  \subfloat[Creative story]{\includegraphics[width=0.24\textwidth]{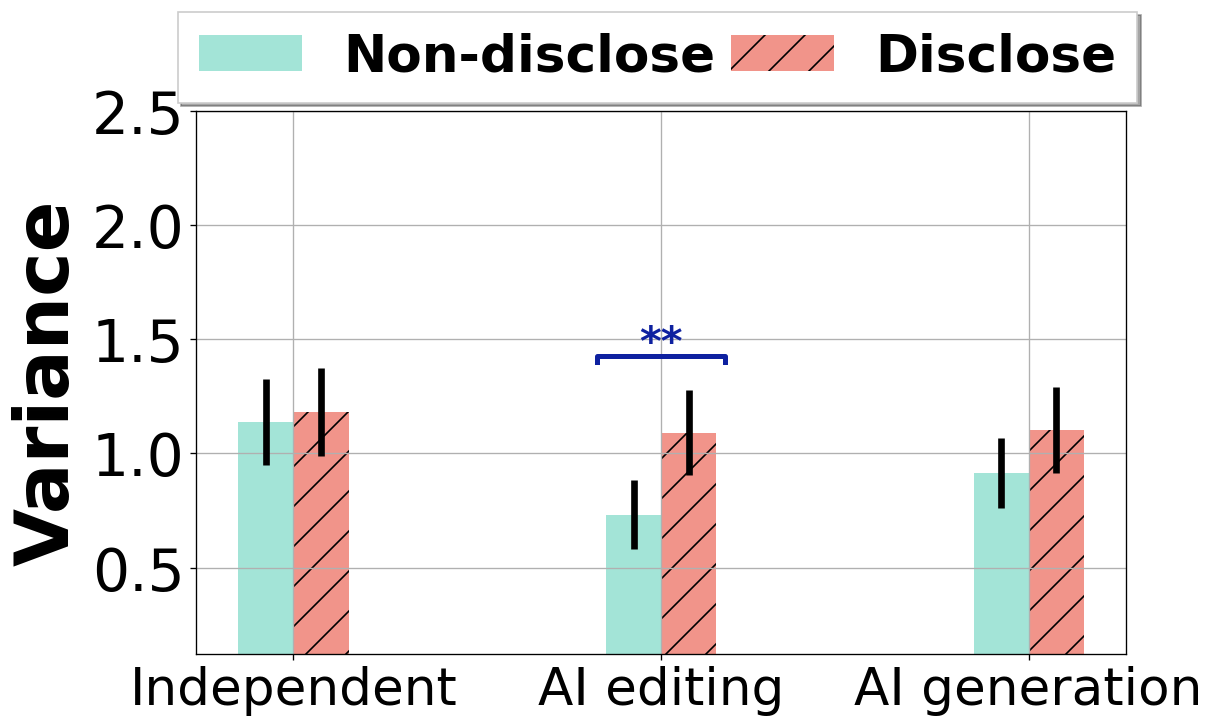}\label{fig:variance_story}}
  \vspace{-5pt}
  \caption{Comparing the {\em variance} in the overall quality ratings of articles generated under the independent, AI editing, or AI generation writing modes, with and without disclosure of the use and type of AI assistance during the writing process. Error bars represent the 95\% confidence intervals of the variance. $\textsuperscript{**}$ denotes the significance level of $0.01$. }
 \label{rating_variance}
  \vspace{-15pt}
\end{figure}

Figure~\ref{rating_variance} compares the variance in the overall quality ratings given to the same argumentative essays or the same creative stories, when the use and type of AI assistance was or was not revealed to Phase 2 participants. Interestingly, for the three settings where we previously found the disclosure of AI assistance results in significant decreases in people's perceived overall quality of the articles (i.e., when AI provides content generation assistance for both arguments and stories, and when AI provides editing assistance for stories; see Figure~\ref{rating_treatment}), the variance in people's overall quality ratings for the same article also appears to increase. 
Focusing on articles written under each of the three writing modes separately, we then fitted regression models to predict the variance in the overall quality ratings for each article based on the disclosure of AI assistance. The regression results suggest that compared to participants in the ``{\em Non-Disclose}'' treatment who were unaware of how the article was written, those participants who were informed about the AI's content generation assistance showed a significantly larger variation in their overall quality evaluations of the same argumentative essay ($p=0.004$). Similarly, the disclosure of AI's editing assistance also makes participants significantly diverges on their evaluations of the same creative story ($p=0.005$). 
Further analyses on people's detailed evaluations on various aspects of the articles (e.g., organization, originality) also show that disclosing AI's content generation assistance consistently results in a significantly higher level of variation ($p<0.05$) in people's evaluations on almost all aspects of an argumentative essay (except for its grammar and vocabulary; see the Appendix~\ref{dispersion_sm} for more details). 

These results suggest that the disclosure of people's usage of AI assistance not only has a general tendency to decrease the average quality evaluation of the writings, but also substantially increase the 
uncertainty in the evaluation as it becomes more unpredictable and highly susceptible to variability depending on
{\em who} is evaluating the writing.

\subsection{Individual Heterogeneity in the Impacts of AI Disclosure on the Quality Evaluation 
}

\begin{figure}[t]
  \centering
  \subfloat[Argumentative essay]{\includegraphics[width=0.24\textwidth]{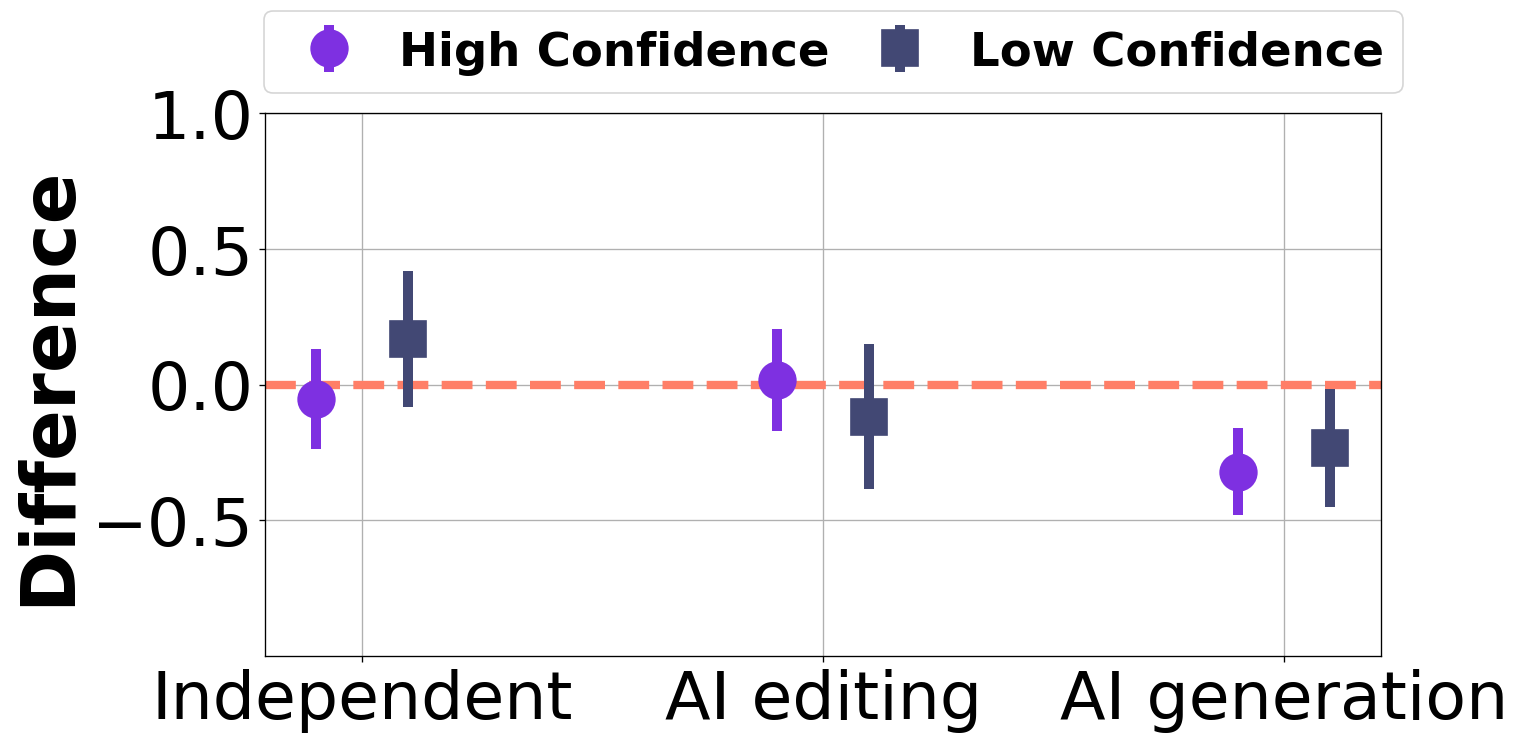}\label{fig:confidence_rating_statement}}
  \hfill
  \subfloat[Creative story]{\includegraphics[width=0.24\textwidth]{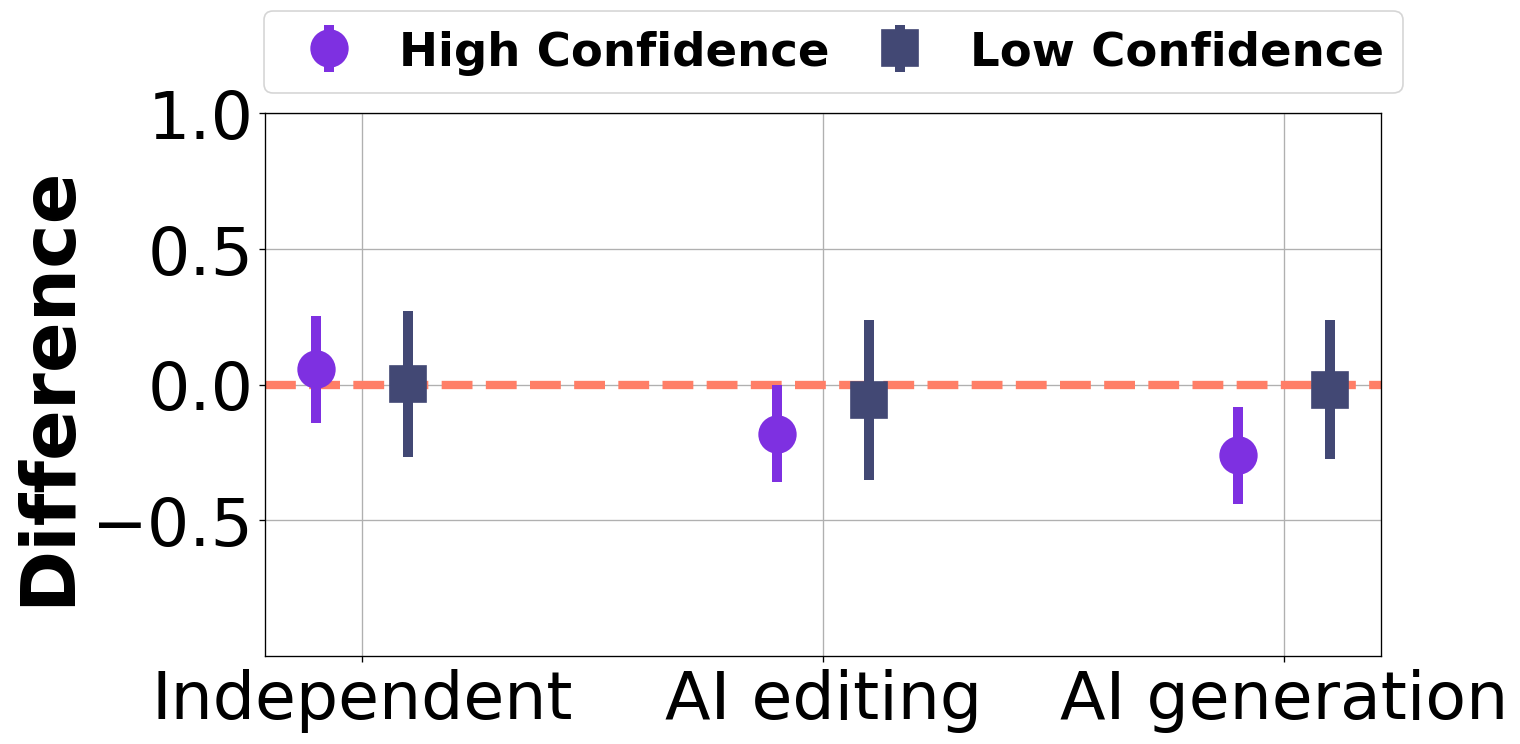}\label{fig:confidence_rating_story}}
  \vspace{-5pt}
  \caption{The average {\em difference} between an article's overall quality ratings in the ``{\em Disclose}'' and ``{\em Non-Disclose}'' treatments, among raters with high versus low {\em confidence in their own writing skills}.  Error bars represent the 95\% bootstrap confidence intervals of the rating difference. An interval below zero means the corresponding group of raters significantly decrease their ratings when the use and type of AI assistance in the writing process was revealed to them. }
\label{confidence_rating}
  \vspace{-15pt}
\end{figure}

In this section, we focus on examining whether and how a rater's own characteristics may moderate the effects of the disclosure of AI assistance, and we identified two characteristics as potential moderating factors.  We begin with considering how disclosing the use and type of AI assistance in the writing process would affect the evaluation of raters who had different levels of confidence in their own writing skills. To investigate into this, we first divided all participants in our Phase 2 study into two groups based on a median split of their self-reported confidence in writing. Within each group of participants, we could compute the {\em difference} in the overall quality ratings given by participants assigned to the ``{\em Disclose}'' and ``{\em Non-Disclose}'' conditions, separately for articles generated under each writing mode. We then used bootstrapping ($R=10,000$) to compute the 95\% bootstrap confidence intervals of these differences, and the results are visualized in Figure~\ref{confidence_rating}. Intuitively, if the 95\% bootstrap confidence interval of the rating difference between with and without disclosure is below zero, it means that raters significantly decrease their evaluation of articles when they become aware of the use and type of AI assistance during the writing process. 

As shown in Figure~\ref{confidence_rating}, compared to those who have low confidence in writing, people who are more confident in writing themselves are more likely to lower their evaluation upon the disclosure of AI assistance in the writing process. For example, on average, raters with high writing confidence decreased their overall quality ratings on argumentative essays  by $0.32$ (95\% $\text{CI}=[-0.48, -0.16]$) once they knew  ChatGPT assisted the authors in generating some content in these essays. Similarly, disclosing the use of AI's editing assistance (or content generation assistance) in the creative story writing process also resulted in a decrease of $0.18$, 95\% $\text{CI}=[-0.35, -0.01]$ (or $0.26$, 95\% $\text{CI}=[-0.44, -0.08]$) in the overall quality ratings given by raters with high writing confidence.  
In contrast, for raters with low confidence in their own writing skills, their evaluation of the articles was seldomly influenced by the disclosure of AI assistance significantly---the only exception was observed when AI's content generation assistance during the argumentative essay writing process was revealed ($\Delta=-0.23 [-0.44, -0.02]$)\footnote{We conducted a validation check by dividing participants into low, medium, and high groups based on a three-quantile split of their self-reported writing confidence. We observed that the decrease in ratings upon disclosing AI assistance in the writing process primarily originated from participants with high writing confidence.}. 

\begin{figure}[t]
  \centering
  \subfloat[Argumentative essay]{\includegraphics[width=0.24\textwidth]{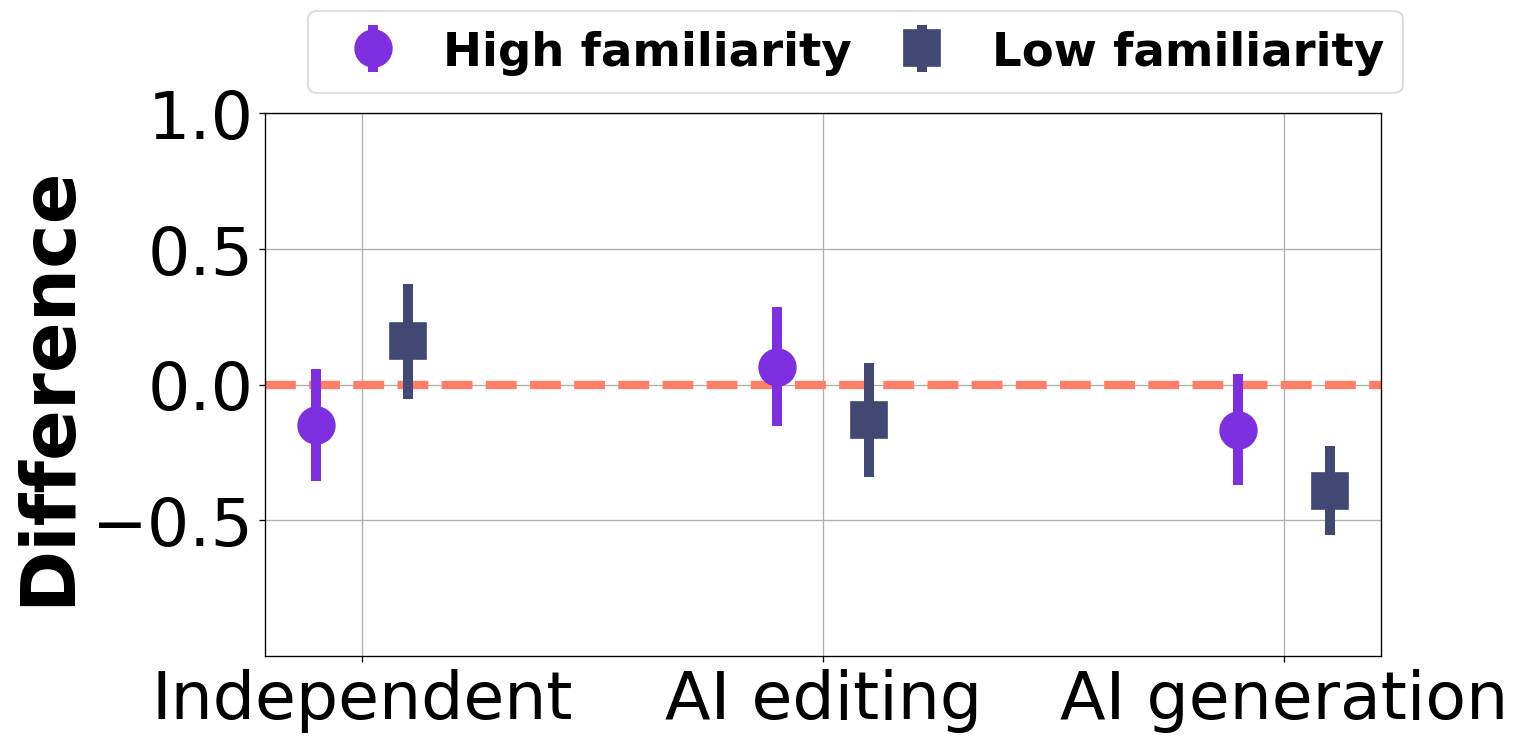}\label{fig:fam_rating_statement}}
  \hfill
  \subfloat[Creative story]{\includegraphics[width=0.24\textwidth]{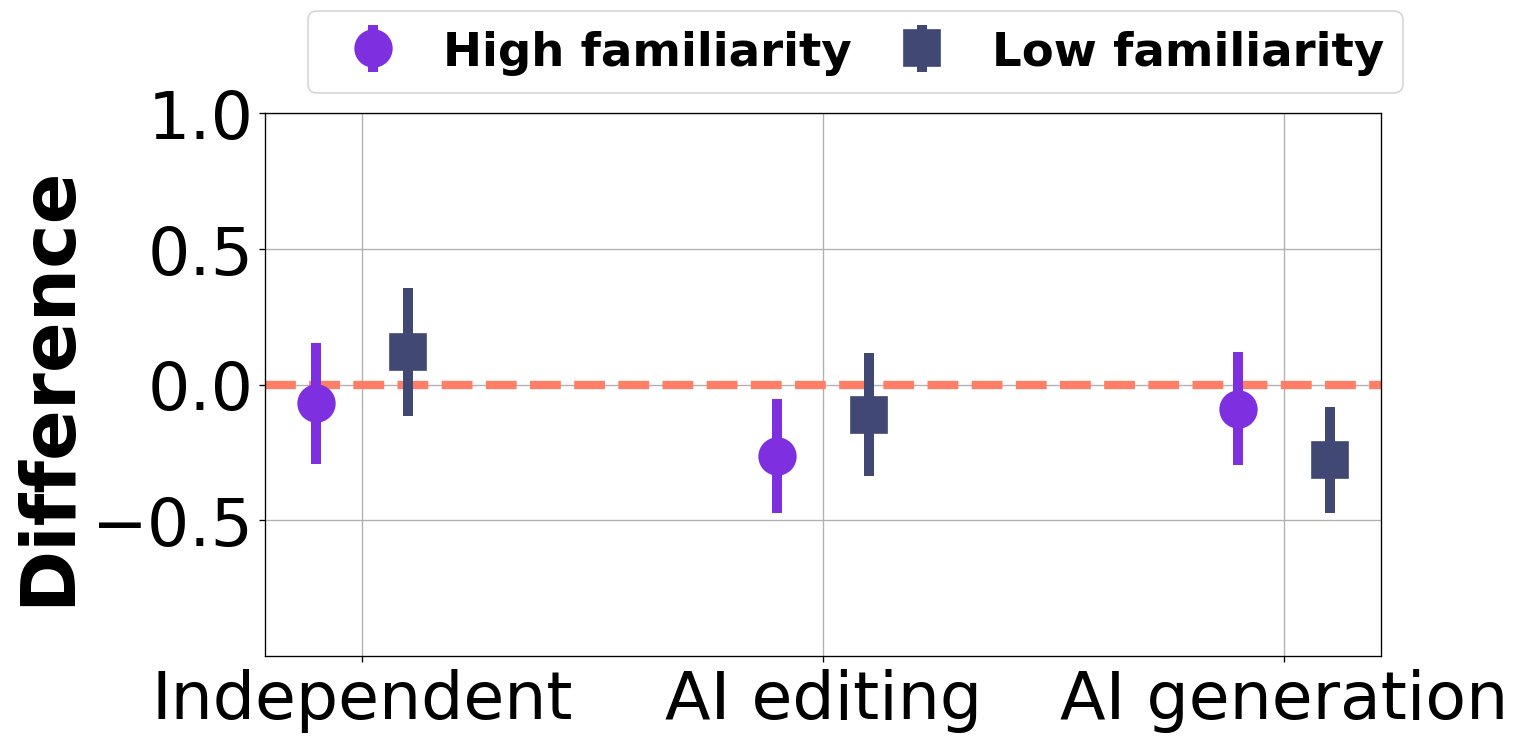}\label{fig:fam_rating_story}}
  \vspace{-5pt}
  \caption{The average {\em difference} between an article's overall quality ratings in the ``{\em Disclose}'' and ``{\em Non-Disclose}'' treatments, among raters with high versus low {\em familiarity with ChatGPT}. Error bars represent the 95\% bootstrap confidence intervals of the rating difference. An interval below zero means the corresponding group of raters significantly decrease their ratings when the use and type of AI assistance in the writing process was revealed to them.}
\label{fam_rating}
  \vspace{-15pt}
\end{figure}

Another possible moderating factor is raters' familiarity with ChatGPT, which could influence their perception of finished articles generated with the use of AI assistance. Again, we divided all participants in our Phase 2 study into two groups based on a median split of their self-reported familiarity with ChatGPT, and Figure~\ref{fam_rating} shows the difference in the overall quality ratings between the two treatments, among participants with high familiarity with ChatGPT and those with low familiarity, separately. From the figure, it is clear that the decrease in the evaluation of writings after the disclosure of AI's content generation assistance was mainly driven by raters with low familiarity with ChatGPT (on argumentative essays: $\Delta=-0.39[-0.55, -0.23]$; on creative stories: $\Delta=-0.28[-0.47, -0.09]$). 
Interestingly, we also noticed that when AI's editing assistance during the creative story's writing process was revealed to participants, the decrease in the rating of the story primarily came from those participants with high familiarity with ChatGPT ($\Delta=-0.26[-0.47, -0.05]$).

Finally, we found that raters' own writing confidence and familiarity with ChatGPT also moderate the impacts of AI assistance disclosure on their detailed evaluations of different aspects of the writings (e.g., originality and creativity). See Appendix~\ref{heter_sm} for detailed results.


\subsection{Impacts of Disclosing AI Assistance on the Ranking of Writing}

In reality, crowd workers frequently contend with each other for opportunities to be hired or rewarded based on the reviews they receive from employers, while regular employees may compete for chances to be promoted based on performance evaluations from managers. Consequently, it is crucial to understand how the disclosure of AI assistance affects the performance ranking of workers. 
In this section, we aim to understand how the disclosure of the use and type of AI assistance during the writing process may influence the ranking of writings generated under different writing modes.  

\begin{figure}[t]
  \centering
  \subfloat[Argumentative essay]{\includegraphics[width=0.24\textwidth]{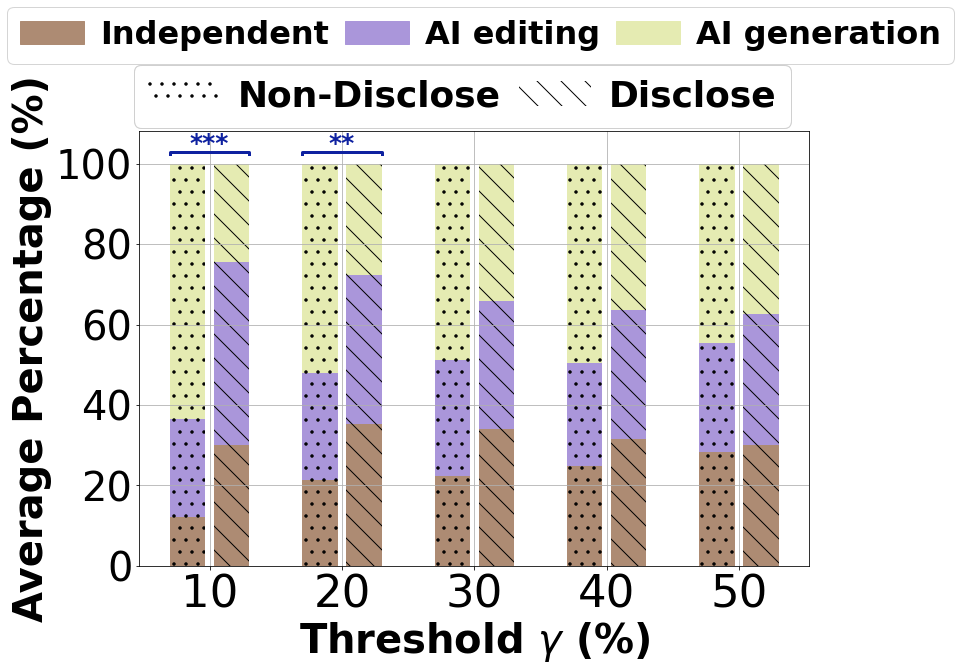}\label{fig:per_statement}}
  \hfill
  \subfloat[Creative story]{\includegraphics[width=0.24\textwidth]{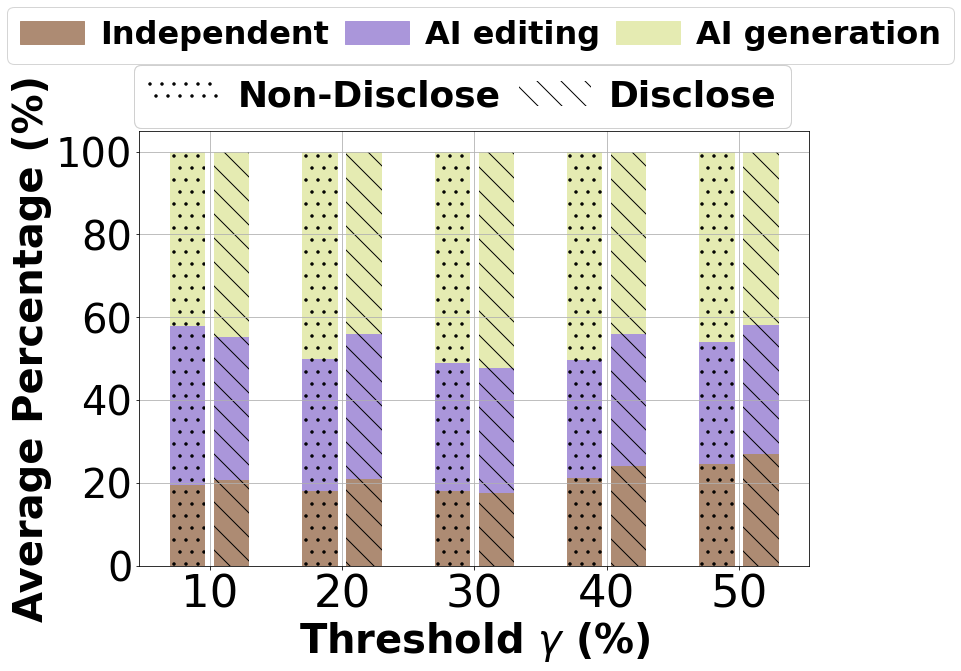}\label{fig:per_story}}
  \vspace{-5pt}
  \caption{Within the top $\gamma \%$ of articles for the same writing task (ranked by articles' average overall quality ratings), the percentages of articles that were written in each of the three writing modes, with and without disclosing the use and type of AI assistance. $\textsuperscript{**}$ and $\textsuperscript{***}$ denote the significance level of $0.01$ and $0.001$, respectively.}
\label{percentage}  
\vspace{-15pt}
\end{figure}

Suppose the articles' average overall quality ratings are used to determine their rankings. Given all the articles written on the same topic, we look into that within the top $\gamma$\% of the articles for this topic, what proportions of the articles are written in the {\em independent}, {\em AI editing}, and {\em AI generation} writing modes, respectively, and how these proportions change after we informed participants about the use and type of AI assistance during the writing process of these articles. 

Figure~\ref{percentage} compares  the average percentages of articles generated under the three writing modes whose overall quality ratings fall within the top $\gamma \%$ ($\gamma\in\{10, 20, \cdots, 50\}$) threshold,  when the use of AI assistance was or was not revealed to raters. For the argumentative essay task, it is clear that disclosing the use and type of AI assistance during the writing process results in a decrease in the proportion of articles generated under the ``{\em AI generation}'' mode and an increase in the proportion of articles generated under the ``{\em AI editing}'' or ``{\em Independent}'' modes
within the highly ranked articles. 
This is especially true when we focus on the most top-ranked articles (i.e., when $\gamma$ is small). 
Chi-square tests suggest that the compositions of the article's writing modes are significantly different between the ``{\em Disclose}'' treatment and the ``{\em Non-Disclose}'' treatment within the top 10\% ($p<0.001$) and top 20\% ($p=0.002$) of argumentative essays.
In contrast, for the creative story task, the disclosure of the use and type of AI assistance during the writing process appears to have minimal impacts on the ranking of different articles. The analysis, based on the ranking criteria for an article's probability of being shortlisted by raters, revealed similar findings---revealing AI assistance significantly decreases the proportions of highly-ranked articles that are generated with ChatGPT's content generation assistance for argumentative essays, while the impacts on the ranking of creative stories are minimal. See Appendix~\ref{rank_sm} for detailed results.

\section{Discussions}
\vspace{-5pt}
\noindent \textbf{Possible explanations for our findings.}
In our study, we made a few interesting observations. For example, we find that disclosing the usage of AI assistance may affect people's evaluations of the writing quality to a larger extent when AI provides assistance in generating new content than when AI provides editing assistance.
This observation may be caused by raters' perceived human effort contributing to the final writings. In particular, when AI provides content generation assistance (i.e., as in the ``AI generation'' writing mode), AI directly generates sentences or paragraphs, reducing the writer's effort significantly (we found that writers under the AI generation writing mode spend significantly less time than writers under other modes). As such, when this type of AI assistance is disclosed, raters might be resistant to giving high scores to the quality of the writings, as they may give less credit to the writers who use AI generation assistance compared to those who do not. Raters may also significantly raise their expectation for the writing quality when knowing that the writers can utilize AI assistance to generate content on behalf of them.

Moreover, we also notice that disclosing the usage of AI assistance appears to have a greater impact on people's perceived ranking of argumentative writing than on creative writing. 
We attribute this to the distinct nature and objectives of these two writing tasks. Argumentative writing requires constructing a logical structure and identifying appropriate evidence, tasks for which LLMs can easily provide support and generate logical structures. When AI usage is not disclosed, this can lead to higher ratings, as raters attribute the quality solely to the writer. However, when AI usage is disclosed, raters may partially credit the quality to the AI assistance, resulting in noticeably lower ratings, as reflected in the significant shift in rankings. In contrast, creative writing relies heavily on imagination to create compelling characters and plots, an area where LLMs still struggle to produce truly novel content. As a result, even when AI's content generation assistance is not disclosed, the perceived quality of the best AI-assisted articles in creative writing do not substantially outperform the best ones written in other modes, especially those written with AI's editing assistance.   
As such, after AI's content assistance is disclosed, 
which causes the perceived quality for both articles written with AI's content generation assistance and editing assistance to decrease, we do not see dramatic changes in the the rankings of the creative writings.

\vspace{5pt}
\noindent \textbf{Design implications.} Our findings have important design implications. 
For example, for online platforms that often determine the ordering of posts based on historical users' evaluation of posts (e.g., ratings and upvotes), one challenge they face is that the prevalence of AI-assisted writings may ``crowd out'' and disengage writers who continue to create content without the use of AI assistance.  
Results of our study suggest that these platforms may benefit from introducing content labels 
to mark those content that are partly generated by AI, as such labels may change users' evaluations of these content in relative to other content that are generated independently by humans, especially when these content have an argumentative flavor.

In addition, the fact that disclosing AI assistance tends to decrease the quality perceptions of writings implies a tension that writers face between being transparent and maintaining the perceived quality of their writing. From platforms' point of view, this means that they need to design appropriate incentives for writers to be willing to disclose the AI assistance usage, and ensure no unintended disadvantage or ``discrimination'' for those who do. From the writers' point of view, they should still recognize their ethical responsibility to inform readers about how the writing is generated to maintain readers' trust, as misleading readers about AI involvement in the writing may erode trust in the long run.

\vspace{5pt}
\noindent \textbf{Limitations.} Our study has a few limitations. For example, the type of AI assistance and writing tasks we considered do not capture the full range of AI assistance and writing tasks in the real world. 
Both the collection and evaluation of writing samples in our study are conducted on an online platform. Participants, primarily motivated by financial payments, may not accurately reflect professional writers and evaluators in the real-world. We also limited our study to U.S. workers to control for the potential variability in English language habits or preferences that differ across countries, limiting the generalizability of findings to non-English speaking populations. 
Future research should look into whether our findings generalize to a broader spectrum of AI writing assistance, a wider range of writing tasks, people with more diverse demographic and cultural backgrounds, and in contexts that more closely mimic real-world writing and evaluation scenarios.

It is also worthwhile to note that in our study, we observed that articles completed in the AI-assisted modes, particularly those under the ``{\em AI generation}'' mode, received significantly higher quality ratings from raters compared to articles completed independently by participants. Thus, even after the AI assistance is disclosed, participants who used the AI's content generation assistance may still receive a higher average rating of their writing quality than participants who wrote independently.  
However, as we noted in footnote \ref{design_footnote}, due to the design of the Phase 1 study (i.e., participants
self-selected into their preferred writing mode), a comparison across the three writing modes does not allow for a causal interpretation. Future work should conduct a rigorously designed study to see whether this observation still holds true.

\vspace{-5pt}
\section{Conclusions}
\vspace{-4pt}
In this paper, we ask the question of whether and how the disclosure of AI assistance would influence people's perceptions of writings. 
Our experimental results suggest that when people are informed of the use of AI content generation assistance in the writing process, there is a significant decrease in the quality evaluation of the writing across different types of writing tasks. Furthermore, the disclosure of AI assistance often leads to an increased level of variation in the perceived writing quality. We identify potential factors, such as the rater's confidence in writing and their familiarity with ChatGPT, that might moderate the effect of disclosing AI assistance on the evaluation of writing quality. Additionally, our findings suggest that disclosing the use of AI assistance would also significantly reduce the proportion of AI-assisted writings among top-ranked argumentative essays.


\bibliography{custom}

\appendix

\newpage
\counterwithin{figure}{section}
\counterwithin{table}{section}

\newpage

\section{Demographic Information of Raters}
\label{demographics}

\begin{table*}
\centering
\caption{Details of the demographic backgrounds of raters in our study.}
\label{tab:demo}
\begin{tabular}{cccc} 
\hline
\multicolumn{2}{c}{Demographics}                       & \begin{tabular}[c]{@{}c@{}}Non-Disclose\\($N=380$)\end{tabular} & \begin{tabular}[c]{@{}c@{}}Disclose\\($N=406$)\end{tabular}  \\ 
\hline
\multirow{3}{*}{Gender}    & Male                      & 48.3\%                                                          & 50.9\%                                                       \\
                           & Female                    & 49.3\%                                                          & 47.3\%                                                       \\
                           & Other                    & 2.4\%                                                           & 1.8\%                                                        \\ 
\hline
\multirow{3}{*}{Age}       & Below 35                  & 40.2\%                                                          & 39.8\%                                                       \\
                           & 35--44                    & 28.7\%                                                          & 26.3\%                                                       \\
                           & 45 or above               & 31.1\%                                                          & 33.9\%                                                       \\ 
\hline
\multirow{4}{*}{Race}      & White                     & 72.7\%                                                          & 68.4\%                                                       \\
                           & Black                     & 13.3\%                                                          & 14.9\%                                                       \\
                           & Hispanic                  & 7.6\%                                                           & 6.6\%                                                        \\
                           & Other                    & 6.4\%                                                           & 10.1\%                                                       \\ 
\hline
\multirow{4}{*}{Education} & High school or lower      & 14.1\%                                                          & 14.1\%                                                       \\
                           & Some college              & 30.5\%                                                          & 29.8\%                                                       \\
                           & Bachelor Degree           & 39.2\%                                                          & 40.2\%                                                       \\
                           & Graduate school or higher & 16.2\%                                                          & 15.9\%                                                       \\ 
\hline
\multicolumn{2}{c}{Average writing confidence}         & 3.84                                                            & 3.88                                                         \\ 
\hline
\multicolumn{2}{c}{Average writing frequency}          & 2.24                                                            & 2.21                                                         \\ 
\hline
\multicolumn{2}{c}{Average frequency of ChatGPT use}   & 3.72                                       & 3.52                                  \\
\hline
\end{tabular}
\end{table*}

In total, 786 workers from Prolific took our study and passed the attention check. Among them, 380 were allocated to the {\em Non-Disclose} treatment, while the remaining 406 were assigned to the {\em Disclose} treatment. The full demographic information of participants is shown in ~\ref{tab:demo}. The writing confidence, writing frequency, and the frequency of ChatGPT use are measured on a 5-point Likert scale from 1 (very low) to 5 (very high).

\section{The Design of Phase 1 Study (Additional Details)}
\label{design1}

As discussed in the main paper, beyond enabling the collection of writing samples produced under different writing modes, our Phase 1 study serves the purpose of estimating the financial value that people attach to different types of AI assistance in their writing. Below, we provide more details of the design of this study.

\subsection{Writing Task}
Participants of our Phase 1 study were asked to write a 200--250 word article within 45 minutes. To examine how people's value of AI assistance in writing 
may vary with the nature of the writing task, 
we considered two kinds of writing jobs in our experiment:
\begin{itemize}
\item \textbf{Argumentative essay writing}: Participants were provided with a statement, and they were asked to write an essay to discuss the argument in the statement. In their essays, participants had the freedom to either support or oppose the argument in the statement. We considered three statements that were sampled from the pool of TOEFL writing exam topics: 
\begin{enumerate}
\item ``Some people think that if companies prohibit sending emails to staff on weekend or during other time out of office hours, staff's dissatisfactions with their companies will decrease. Others think this will not reduce the overall dissatisfactions among staff.''
\item ``Govenrment should put higher tax on junk food to reduce consumption.''
\item ``Nowadays it is easier to maintain good health than it was in the past.''
\end{enumerate} 

\item \textbf{Creative story writing}: Participants were given a prompt, and they were asked to write a story that includes the prompt. Drawing from the array of popular creative writing tasks featured on Reedsy’s Short Story Contest\footnote{ \url{https://blog.reedsy.com/creative-writing-prompts/terms/}.}, we curated three writing prompts:

\begin{enumerate}
    \item (Someone) ``realizes they're on the wrong path.''
    \item Including the line ``We're just too different.''
    \item Someone saying ``Let’s go for a walk.''
\end{enumerate}
\end{itemize}
 To ensure that the argumentative essay and creative story writing tasks chosen for our Phase 1 study have a comparable and reasonable difficulty level, we conducted a pilot study to test the difficulty of different argumentative essay statements and creative story prompts.      
In this pilot study, participants were asked to complete the writing task on their own. 
For the final set of 3 statements and 3 prompts selected for our Phase 1 study, as detailed above, our pilot study results suggested that participants could successfully complete the essay/story writing job within the time limit, yielding articles of satisfactory quality.

\subsection{Writing Modes}
In our study, we included three different writing modes, each reflecting varying degrees of assistance from and collaboration with a state-of-the-art LLM, i.e., ChatGPT. These modes encompass: 
\begin{itemize}
    \item \textbf{Independent}: In this mode, participants completed the writing task independently without any assistance from ChatGPT.
    
    \item \textbf{AI editing}: In this mode, participants were primarily responsible for writing the article. Meanwhile, participants could send any part of their drafts to ChatGPT for editing and polishing, and then they could decide how to integrate the polished texts into their writing. Note that in this mode, to ensure that ChatGPT would only provide editing assistance to participants, we covertly crafted a prompt employing the OpenAI API by appending the following instructions ahead of the text that participants sought to polish: ``You should only edit or polish the texts I send to you. Please do not write any new content.''
    
    \item \textbf{AI generation}: In this mode, ChatGPT took the lead in drafting the initial version of the article. Participants could then provide feedback and direct the subsequent revisions of the article through conversational interactions with ChatGPT. In the end, participants decided how to compose the final article based on different versions of the drafts that ChatGPT generated, and they could also add some of their own writing into it.
\end{itemize}

\subsection{Experimental Treatments}
To quantify how much financial value that people attach to
different kinds of writing assistance that LLM-powered assistants could offer, following the classical methods in economics for estimating the ``willingness to pay/accept'' ~\cite{rosen1986theory,mas2017valuing,liang2023hidden}, we created two experimental treatments:
\begin{itemize}
    \item \textbf{Independent vs. AI editing}: In this treatment, after the topic of the writing job (i.e., the statement for writing an argumentative essay or the prompt for writing a creative story) was revealed to participants, we presented participants with two job offers: The first offer paid the participant \$3 to complete the writing job in the \textbf{\em independent} writing mode, while the second offer paid the participant \$$x$ to complete the writing job in the \textbf{\em AI editing} mode, where $x$ was randomly sampled from the set $\{1.5, 2, 2.25, 2.5, 2.75, 3, 3.25, 3.5, 3.75, 4, 4.5\}$. Participants were asked to make a selection between these two offers, and subsequently complete the writing job in accordance with the writing mode specified by the selected offer\footnote{We conducted a pilot test to validate the appropriateness of the lower and upper bounds, \$1.5 and \$4.5, respectively, for estimating participants’ willingness to pay for AI assistance. This consideration was made given that a majority of participants in our pilot study would 
    prefer not  to choose the job with AI assistance at the \$1.5 wage level, while conversely, most participants would select the job with AI assistance at the \$4.5 wage level.}.
    
    \item \textbf{Independent vs. AI generation}: In this treatment, after the topic of the writing job was revealed to participants, we presented participants with two job offers: The first offer compensated the participant with \$3 for completing the writing job in the \textbf{independent} writing mode, 
    while the second offer compensated the participant with \$$x$ for completing the writing job in the \textbf{AI generation} mode, where $x$ was again randomly sampled from the set of values as that in the previous treatment. Participants were asked to make a selection between these two offers, and would then complete the writing job in the writing mode specified by the offer that they chose. 
   
\end{itemize}

Note that to ensure participants could make an informed selection between the two job offers, in both treatments, participants would be initially directed to watch a 2-minute video that introduces to them the type of writing assistance that they could receive from ChatGPT and acquaints them with the writing interface they would use upon selecting the job offer with AI assistance (i.e., the offers associated with \textbf{AI editing } or \textbf{ AI generation} writing modes). Participants could only make their job offer selection after finish watching this video. 

\subsection{Experimental Procedure}

Our study was opened only to U.S. workers whose primary language is English on Prolific, and each worker was only allowed to participate in our study once. Each participant went through a few stages in our study, as detailed below.

\paragraph{\textbf{Background assessment.}}
Upon arrival of the study, participants were first asked to fill out a questionnaire to report their demographic information (e.g., gender, age, education). We then asked  participants to indicate how confident they were in completing six types of writing tasks, including creative writing (e.g., stories, novels), writing argumentative essays, writing emails or letters, writing product or book reviews, writing business reports or proposals, and writing blogs. For each type of writing tasks, participants reported their confidence on a 5-point Likert scale from 1 (very low) to 5 (very high). 

\paragraph{\textbf{Treatment assignment and writing mode selection.}} Subsequently, participants were randomized into one of the two treatments, ``{\em Independent vs. AI editing}'' or ``{\em Independent vs. AI generation}''. They were then presented with their writing task, which could be either writing an argumentative essay or writing a creative story, and the statement/prompt used for the writing task was also selected at random from the candidate pool. 
Next, depending on the experimental treatment the participant was assigned, they would be presented with the corresponding two job offers, and the random payment value \$$x$ ($1.5\le x\le 4.5$) of the offer that provided AI assistance to participants would be realized from its 
set of candidate values. 
Participants were told that their final payment from the study would consist of three parts: (1) a {\em base payment} of \$2; (2) the {\em writing job payment} as specified in the job offer that they would choose; and (3) an (optional) {\em performance-based payment} of \$2. We informed participants that their submitted articles would be sent to other crowd workers for review. If the average rating of their article would rank within the top 10\% of the articles written for the same topic, they would receive the performance-based payment.
Once they were clear on the compensation structure, participants were required to watch an introductory video elucidating how they could potentially collaborate with ChatGPT through the designated interface to accomplish the writing task should they choose the AI-assisted writing mode (i.e., ``{\em AI editing}'' or ``{\em AI generation}''). 
With all this information, participants then selected their preferred job offer. 

\paragraph{\textbf{Main writing task.}} After the job offer was selected, participants proceeded to the main writing task. They were asked to complete this writing task using the writing mode specified in their chosen offer, and they had a maximum duration of 45 minutes for finishing the writing task. Note that the time required for ChatGPT to respond to participants’ prompts would not be included in the allocated time limit. 

\paragraph{\textbf{Exit survey.}} After completing the main writing task, participants were asked to complete an exit survey. In this survey, participants were again asked to indicate their confidence in 
completing the same six types of writing tasks (e.g., creative writing, argumentative essay, emails/letters, etc.) as we surveyed at the beginning of the study, should they have the chance in the future to complete those tasks {\em in the same writing mode} as they had experienced in our study. 
We also asked a series of survey questions to guage participants' perceptions of their writing experience in our study. 
For example, 
the NASA Task Load Index (NASA TLX)~\cite{hart1988development} was used to measure the cognitive load that participants experienced during the writing task, including their mental demand, time pressure, amount of effort taken, and frustration level. 
To understand participants' perceptions of the overall writing processes, we presented the following statements to participants and asked them to rate how much they agreed with each statement on a 5-point Likert scale from 1 (strong disagree) to 5 (strongly agree): 
\begin{itemize}
\item \textbf{(Satisfaction)}: ``I am satisfied with the writing process.''
\item \textbf{(Enjoyment)}: ``I enjoy the writing process.''
\item \textbf{(Ease)}: ``I find it easy to complete the writing process.''
\item \textbf{(Ability of self-expression)}: ``I was able to express my creative goals during the writing process.'' 
\end{itemize}

Similarly, we asked participants to rate their agreement with the following statements, again on a 5-point Likert scale, to understand their perceptions of the final writing outcome:
\begin{itemize}
\item \textbf{(Quality)}: ``I am satisfied with the quality of the final article.''
\item \textbf{(Ownership)}: ``I feel ownership over the final article.''
\item \textbf{(Pride)}: ``I'm proud of the final article.''
\item \textbf{(Uniqueness)}: ``The article I submitted feels unique.''
\end{itemize}

In addition, to understand participants' accountability should their article be criticized for various issues during the evaluation process, 
we asked participants to rate their agreement in the following statements, on a 5-point Likert scale:
\begin{itemize}
\item \textbf{(Deceptive content)}: ``I'm willing to take the responsibility if my article is criticized for containing deceptive content (e.g., misinformation).''
\item \textbf{(Plagiarism)}: ``I'm willing to take the responsibility if my article is criticized for containing content that is highly similar to someone else's writing.''
\item \textbf{(Privacy invasion)}: ``I'm willing to take the responsibility if my article is criticized for containing content that invades someone else's privacy.''
\item \textbf{(Discrimination)}: ``I'm willing to take the responsibility if my article is criticized as exhibiting bias and discrimination.''
\end{itemize}

Finally, participants reported their familiarity with ChatGPT (1: very unfamiliar; 5: very familiar) and their frequency of using ChatGPT in their daily life or work (1: never; 5: very frequently--more than once a day). 

\paragraph{\textbf{Attention check.}} To filter out inattentive participants, we included two attention check questions in our study. The first attention check question was presented to participants right before they took the exit survey, and it asked the participant to select again which job offer they had previously chosen in the study. The second attention check question was included in the exit survey, where participants were required to select a randomly pre-specified option in the question. We considered only the data from participants who passed both attention check questions as valid data.

\section{Impacts of Disclosing AI Assistance on the Quality Evaluation of Writing (Additional Results)} \label{quality_sm}

\begin{figure}[t]
  \centering
  \subfloat[Argumentative essay]{\includegraphics[width=0.24\textwidth]{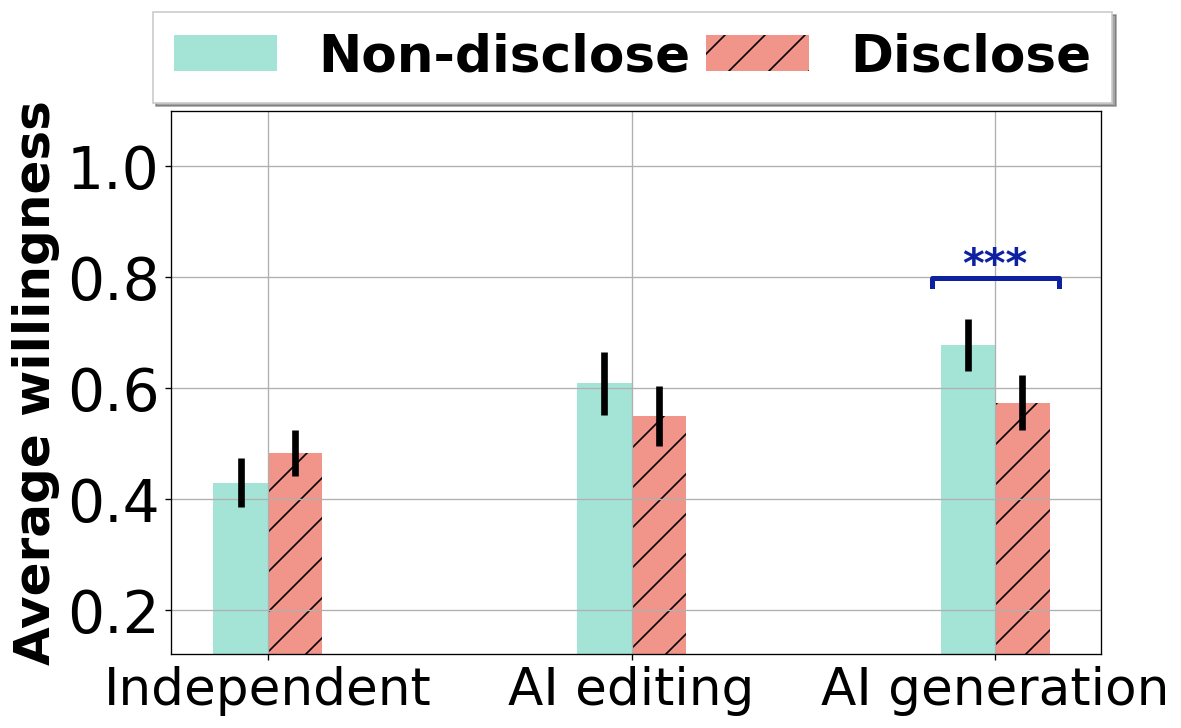}\label{fig:organization_statement}}
  \hfill
  \subfloat[Creative story]{\includegraphics[width=0.24\textwidth]{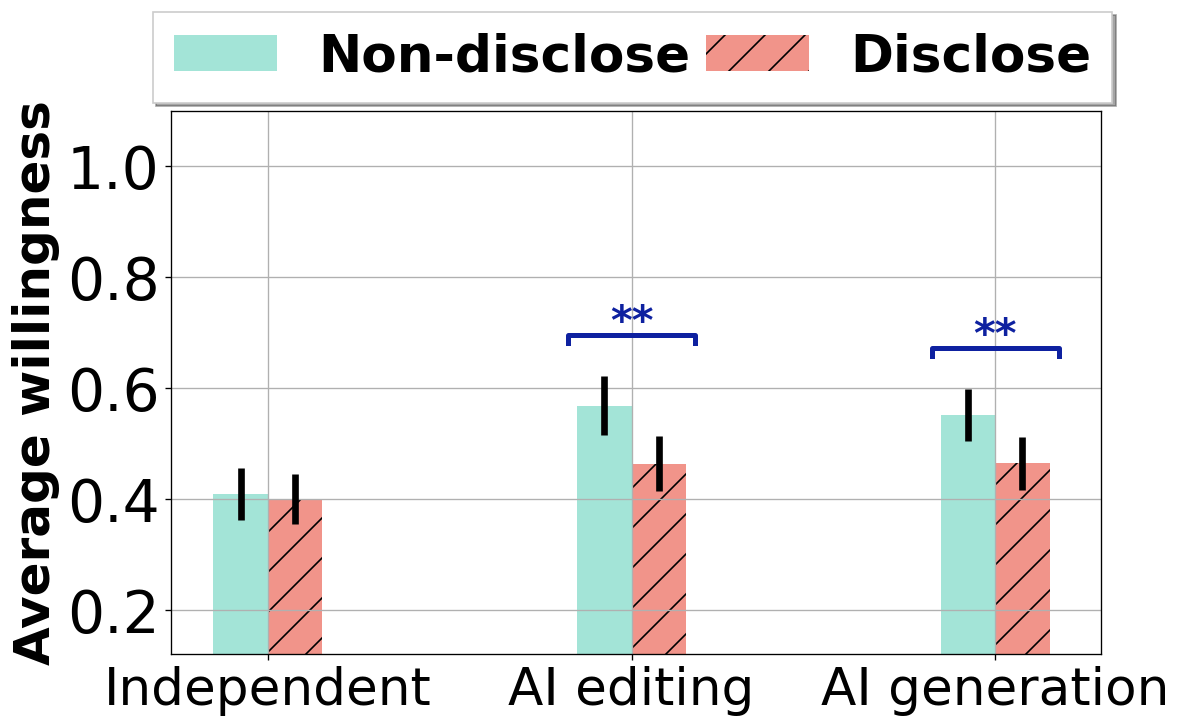}\label{fig:fam_rating_story}}
  \vspace{-5pt}
  \caption{ Comparing the {\em probability} of shortlisting an article that was generated under the independent, AI editing, or AI generation writing modes, when the use and type of AI assistance during the writing process was or was not revealed to raters.
 Error bars represent the 95\% confidence intervals of the probabilities. 
$\textsuperscript{**}$ and $\textsuperscript{***}$ denote significance levels of $0.01$ and $0.001$, respectively.}
\label{recommend_treatment}
  \vspace{-15pt}
\end{figure}

\begin{figure}[t]
  \centering
  \subfloat[Argumentative essay]{\includegraphics[width=0.24\textwidth]{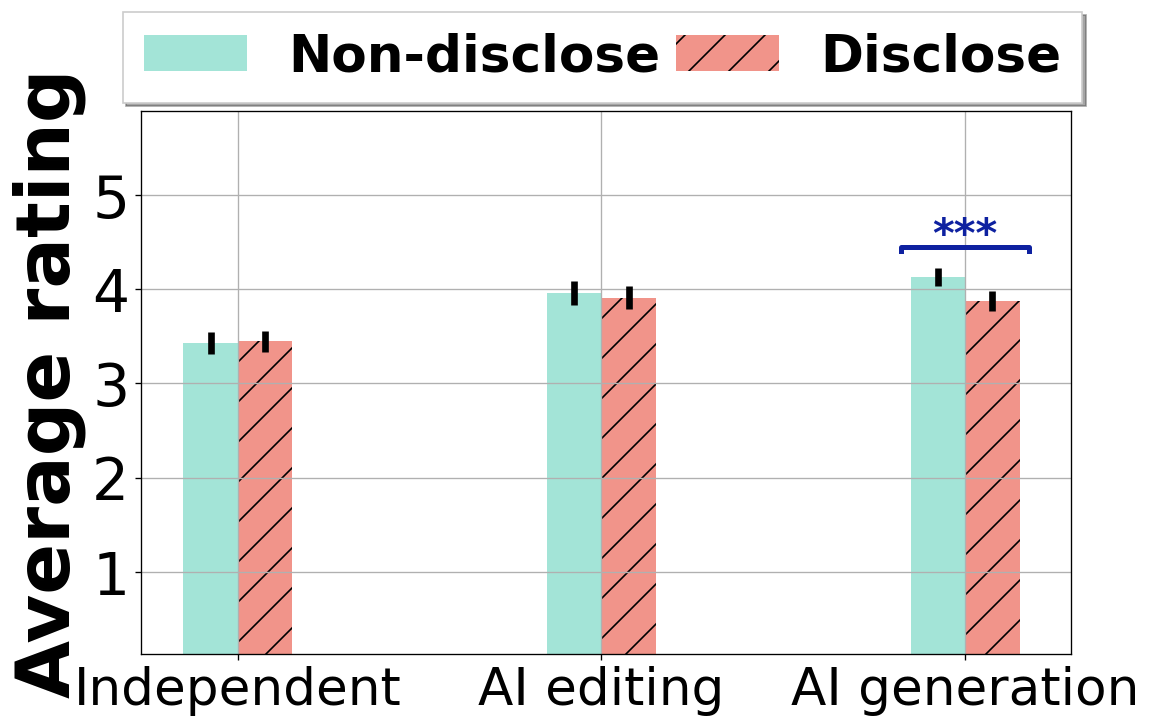}\label{fig:rec_statement}}
  \hfill
  \subfloat[Creative story]{\includegraphics[width=0.24\textwidth]{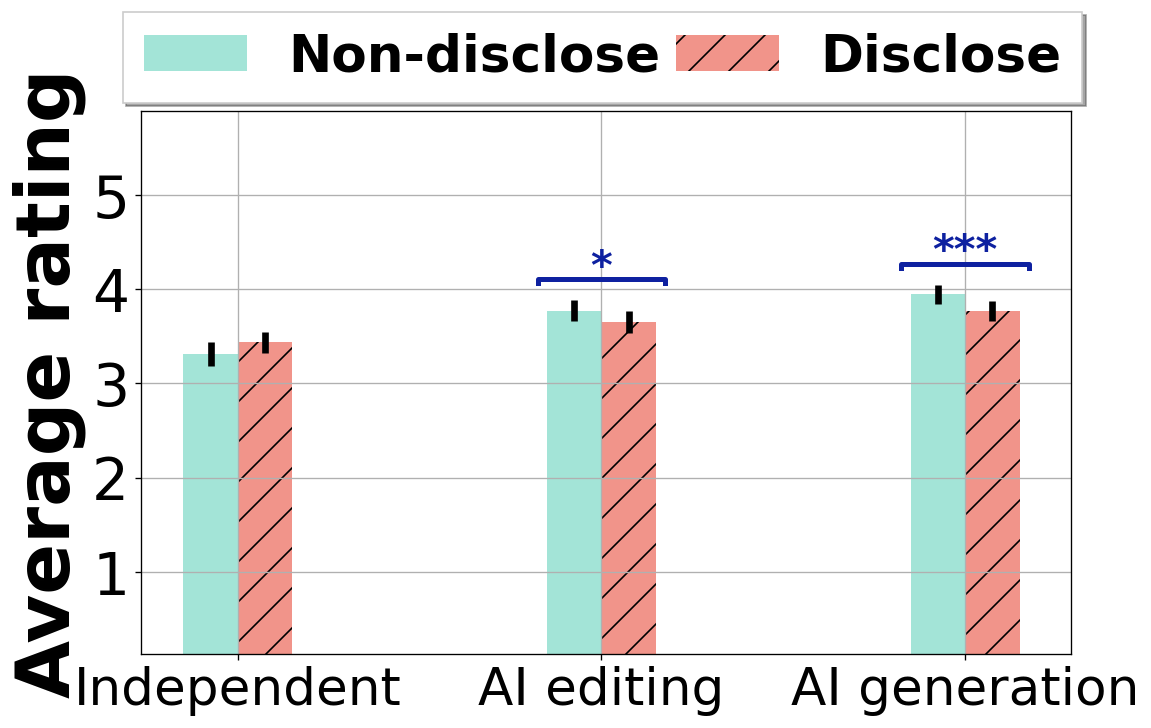}\label{fig:organization_story}}
  \vspace{-5pt}
  \caption{Comparing the {\em average ratings} of organization of articles generated under the independent, AI editing, or AI generation writing modes, when the use and type of AI assistance during the writing process was or was not revealed to raters. Error bars represent the 95\% confidence intervals of the mean values. 
$\textsuperscript{*}$ and $\textsuperscript{***}$ denote significance levels of $0.05$ and $0.001$, respectively.}
\label{organ_treatment}
  \vspace{-15pt}
\end{figure}

\begin{figure}[t]
  \centering
  \subfloat[Argumentative essay]{\includegraphics[width=0.24\textwidth]{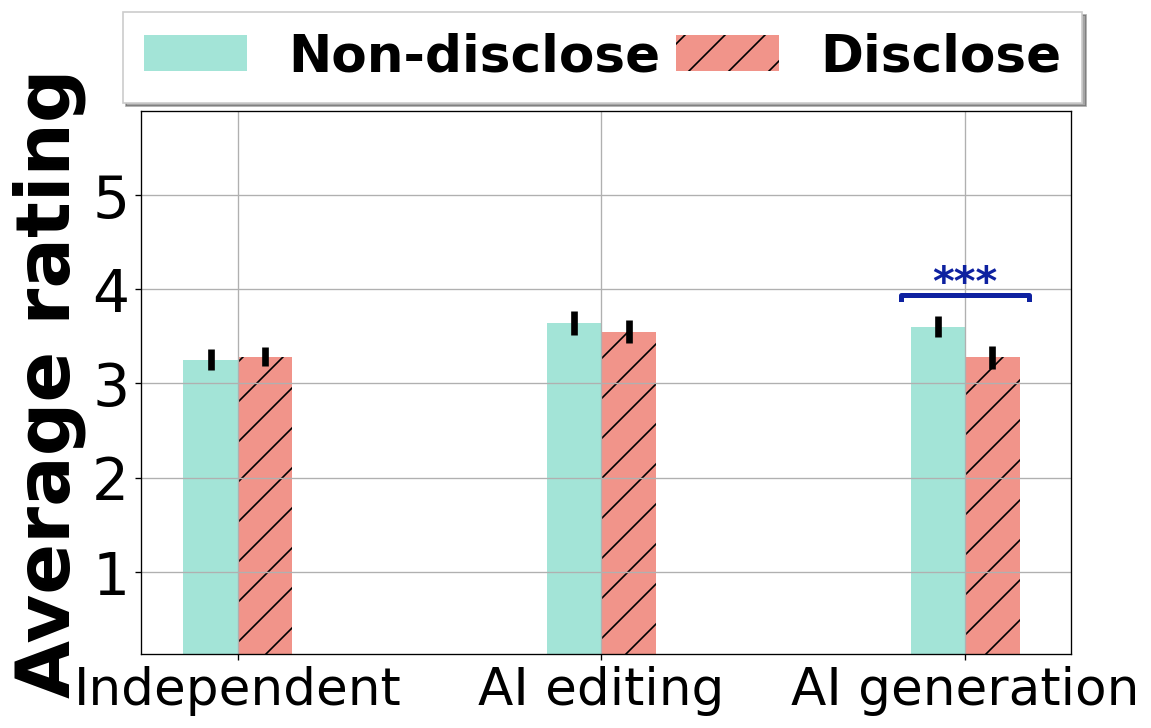}\label{fig:origin_statement}}
  \hfill
  \subfloat[Creative story]{\includegraphics[width=0.24\textwidth]{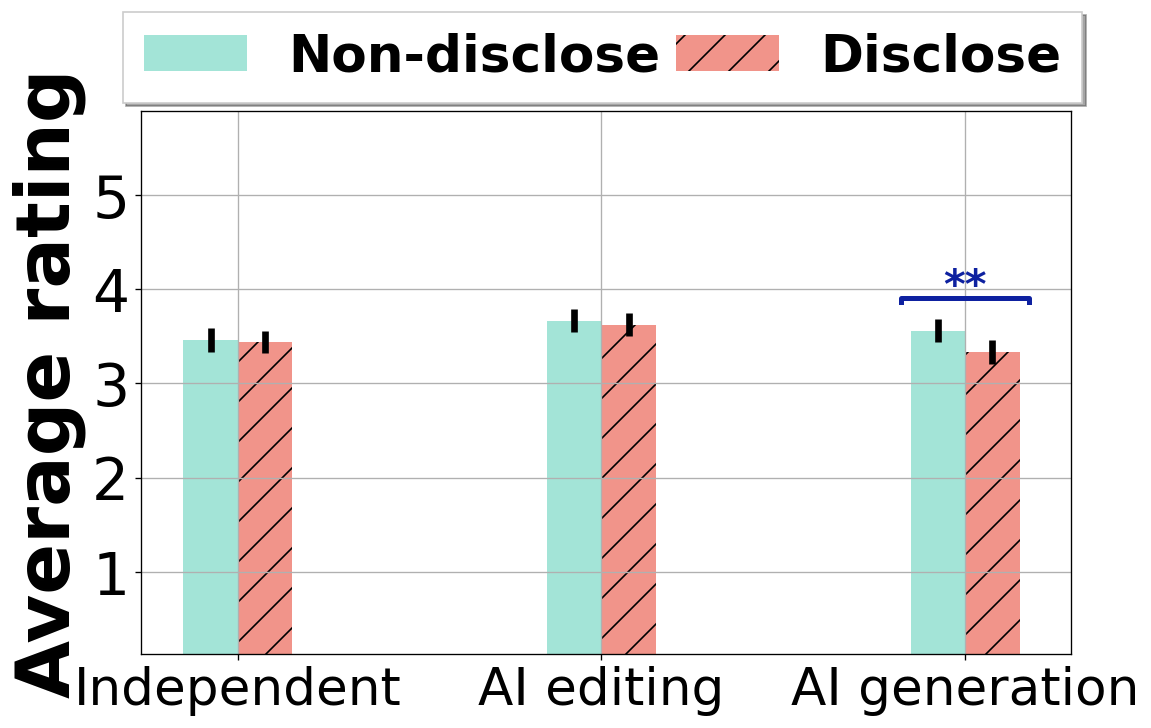}\label{fig:origin_story}}
  \vspace{-5pt}
  \caption{Comparing the {\em average ratings} of originality of articles generated under the independent, AI editing, or AI generation writing modes, when the use and type of AI assistance during the writing process was or was not revealed to raters.
Error bars represent the 95\% confidence intervals of the mean values. 
$\textsuperscript{**}$ and $\textsuperscript{***}$ denote significance levels of $0.01$ and $0.001$, respectively.}
\label{origin_treatment}
  \vspace{-15pt}
\end{figure}

\begin{figure}[t]
  \centering
  \subfloat[Argumentative essay]{\includegraphics[width=0.24\textwidth]{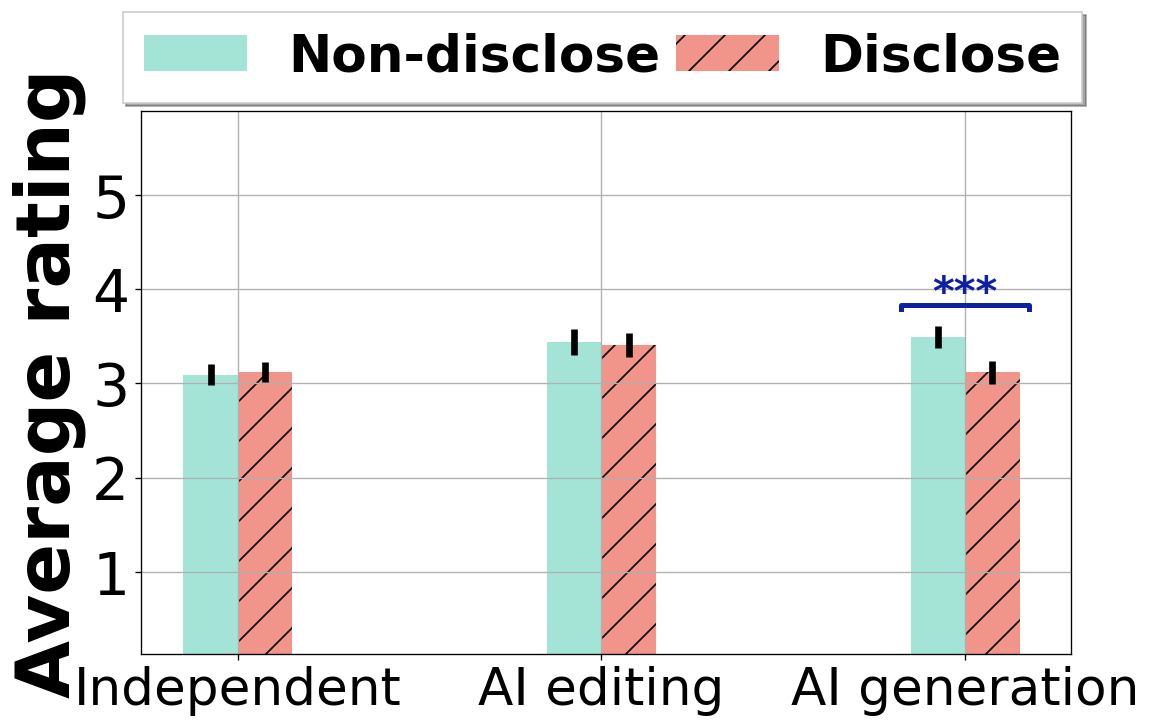}\label{fig:creat_statement}}
  \hfill
  \subfloat[Creative story]{\includegraphics[width=0.24\textwidth]{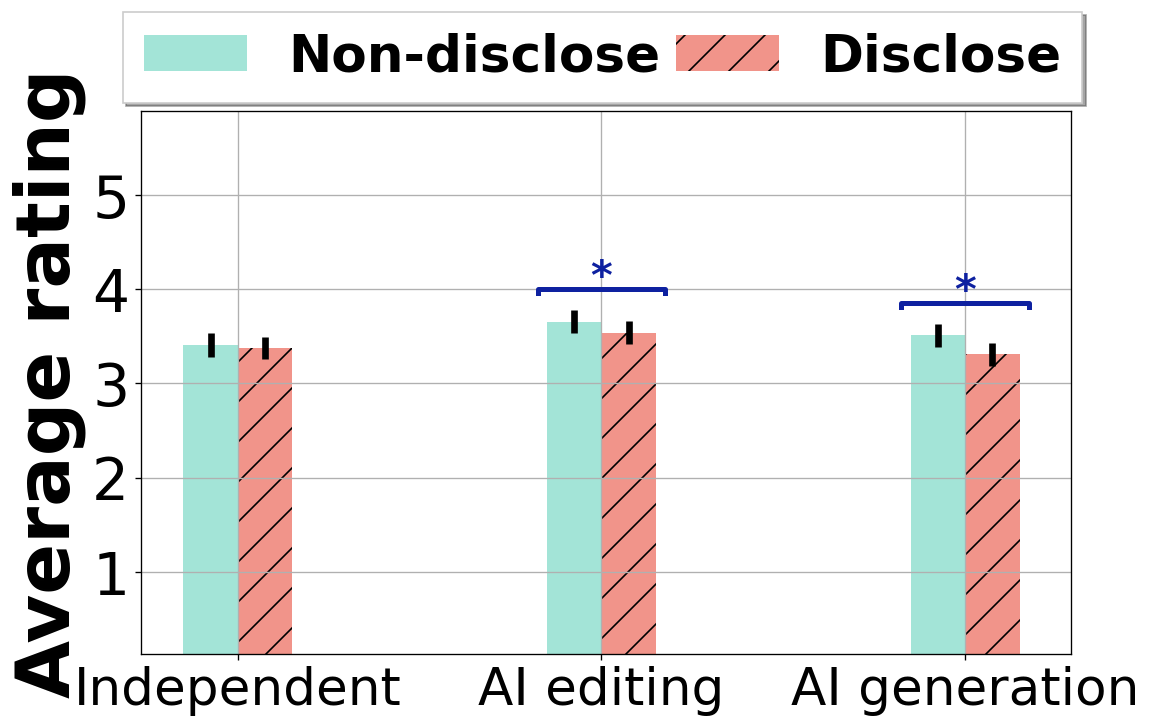}\label{fig:creat_story}}
  \vspace{-5pt}
  \caption{Comparing the {\em average ratings} of creativity of articles generated under the independent, AI editing, or AI generation writing modes, when the use and type of AI assistance during the writing process was or was not revealed to raters. Error bars represent the 95\% confidence intervals of the mean values. 
$\textsuperscript{*}$ and $\textsuperscript{***}$ denote significance levels of $0.05$ and $0.001$, respectively.}
\label{creat_treatment}
  \vspace{-15pt}
\end{figure}

\begin{figure}[t]
  \centering
  \subfloat[Argumentative essay]{\includegraphics[width=0.24\textwidth]{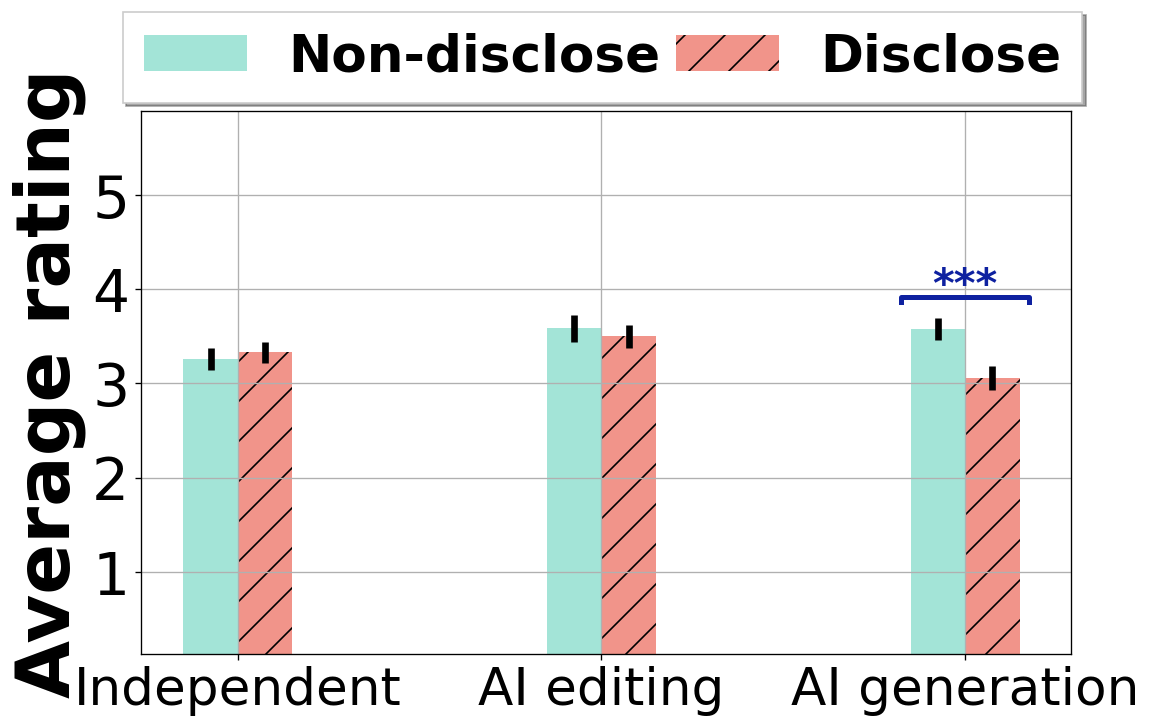}\label{fig:emotion_statement}}
  \hfill
  \subfloat[Creative story]{\includegraphics[width=0.24\textwidth]{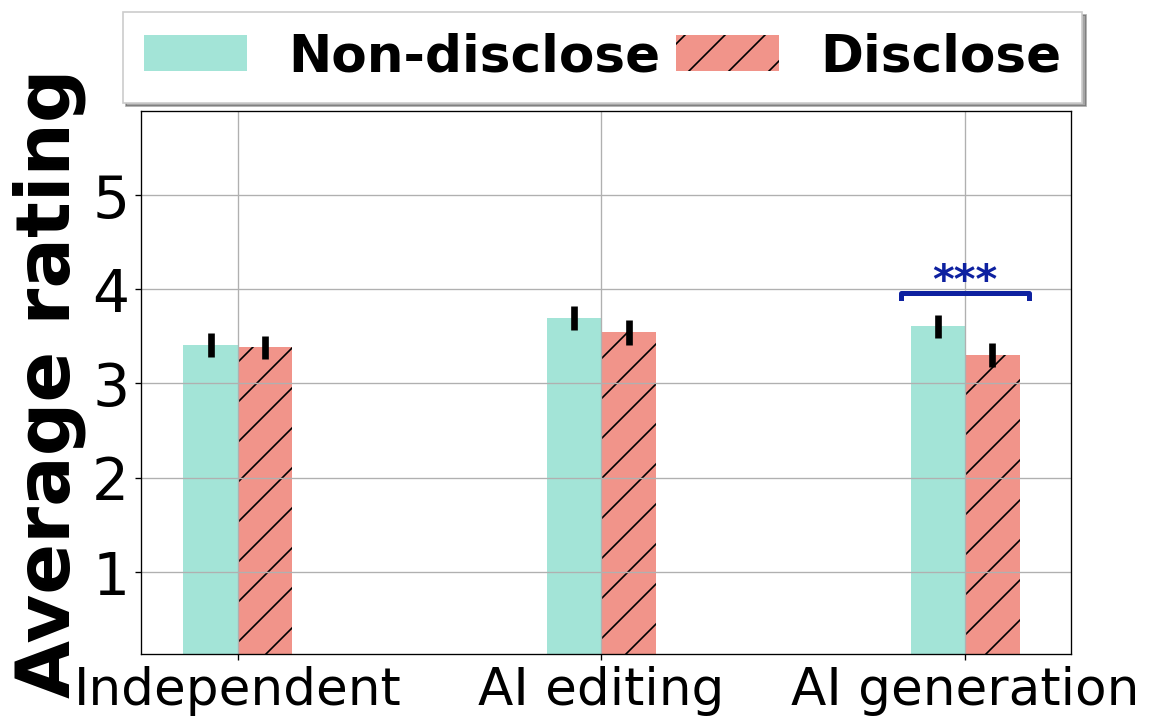}\label{fig:emotion_story}}
  \vspace{-5pt}
  \caption{Comparing the {\em average ratings} of emotion authenticity of articles generated under the independent, AI editing, or AI generation writing modes, when the use and type of AI assistance during the writing process was or was not revealed to raters. Error bars represent the 95\% confidence intervals of the mean values. $\textsuperscript{***}$ denote significance levels of  $0.001$.}
\label{emotion_treatment}
  \vspace{-15pt}
\end{figure}

Figures~\ref{recommend_treatment}--\ref{emotion_treatment}
compare participants' average willingness to shortlist an article, and their detailed evaluations on the article's grammar and vocabulary, organization, originality, creativity, and emotion authenticity, for argumentative essays and creative stories generated under the three writing modes. In general, we find that disclosing ChatGPT's content generation assistance during the writing process significantly decreases people's willingness to shortlist an article and their ratings on all five aspects of evaluations, for both argumentative essays and creative stories ($p<0.05$ for all comparisons between ``{Disclose}'' and ``{Non-Disclose}'' for articles written under the ``AI generation'' mode). In addition, knowing that ChatGPT provides editing assistance during the writing process also leads to people's decreased willingness to shortlist a story, as well as their decreased ratings on a story's organization and creativity ($p<0.05$). 

\section{Impacts of Disclosing AI Assistance on the
Dispersion in the Quality Evaluation (Additional Results) }
\label{dispersion_sm}

\begin{figure}[t]
  \centering
  \subfloat[Argumentative essay]{\includegraphics[width=0.24\textwidth]{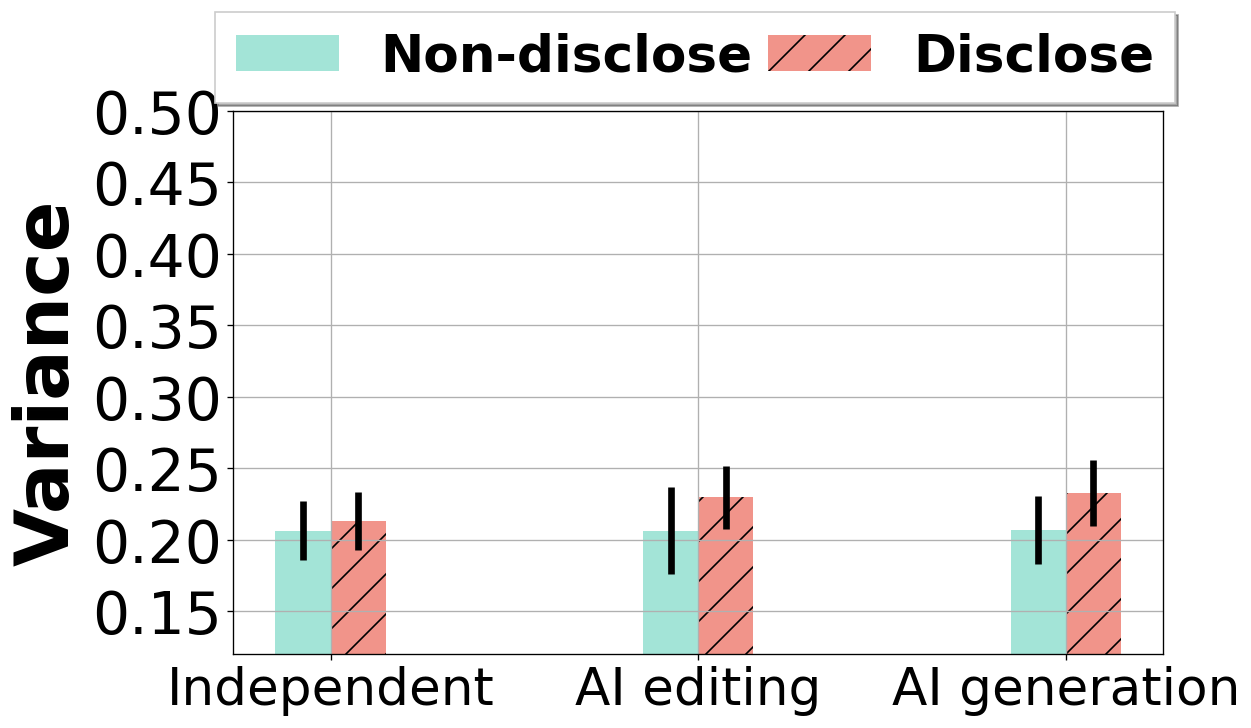}\label{fig:rec_variance_statement}}
  \hfill
  \subfloat[Creative story]{\includegraphics[width=0.24\textwidth]{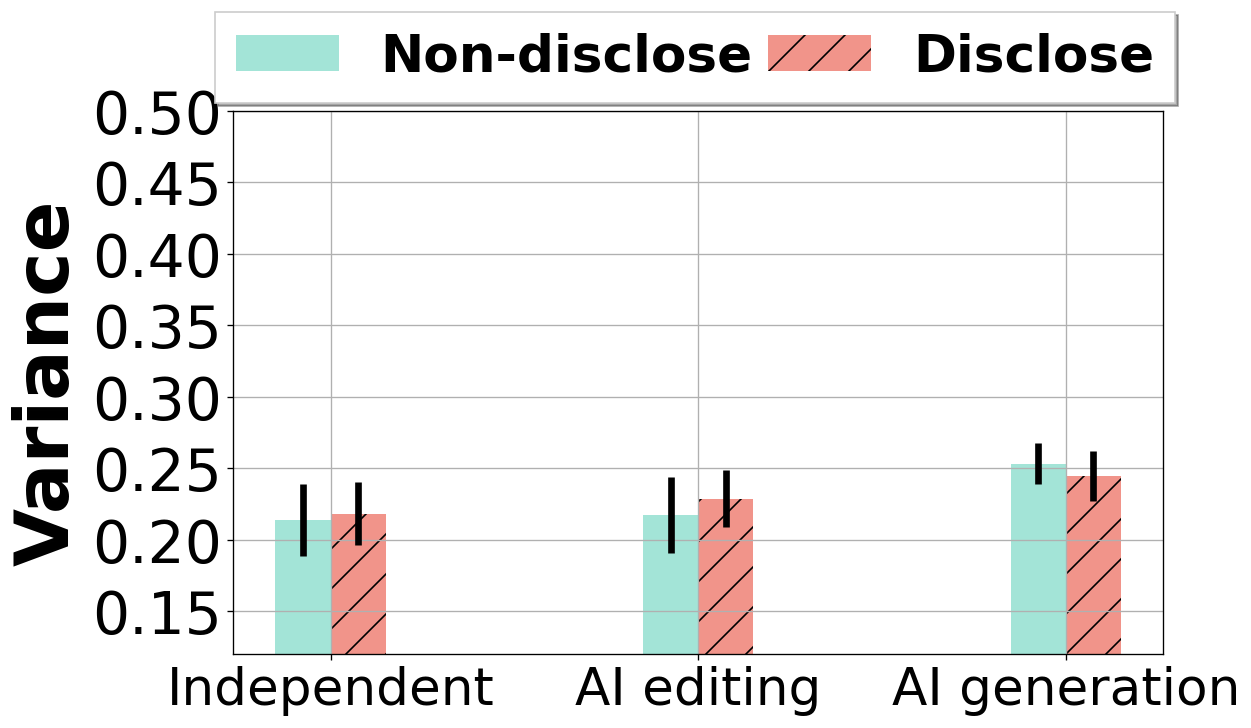}\label{fig:rec_variance_story}}
  \vspace{-5pt}
  \caption{Comparing the {\em variance} in raters' decisions in shortlisting an article that was generated under the independent, AI editing, or AI generation writing modes, when the use and type of AI assistance during the writing process was or was not revealed to raters.  Error bars represent the 95\% confidence intervals of the variance.  }
\label{rec_variance}
  \vspace{-15pt}
\end{figure}

\begin{figure}[t]
  \centering
  \subfloat[Argumentative essay]{\includegraphics[width=0.24\textwidth]{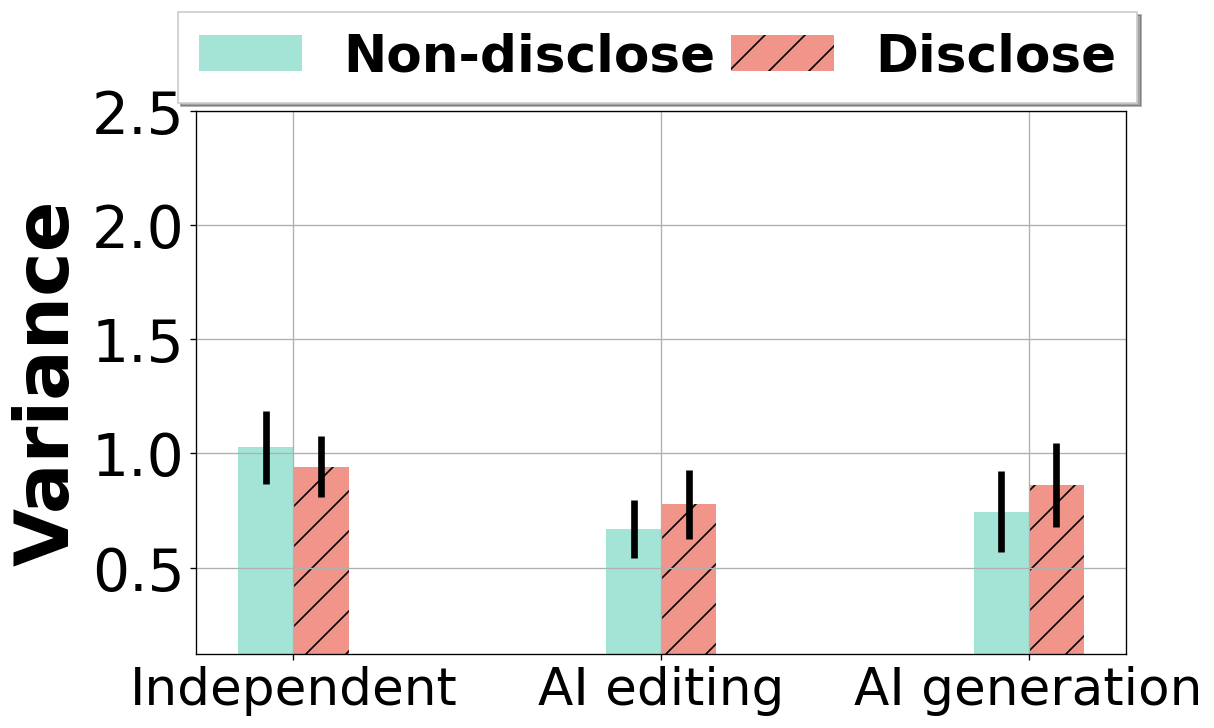}\label{fig:grammar_variance_statement}}
  \hfill
  \subfloat[Creative story]{\includegraphics[width=0.24\textwidth]{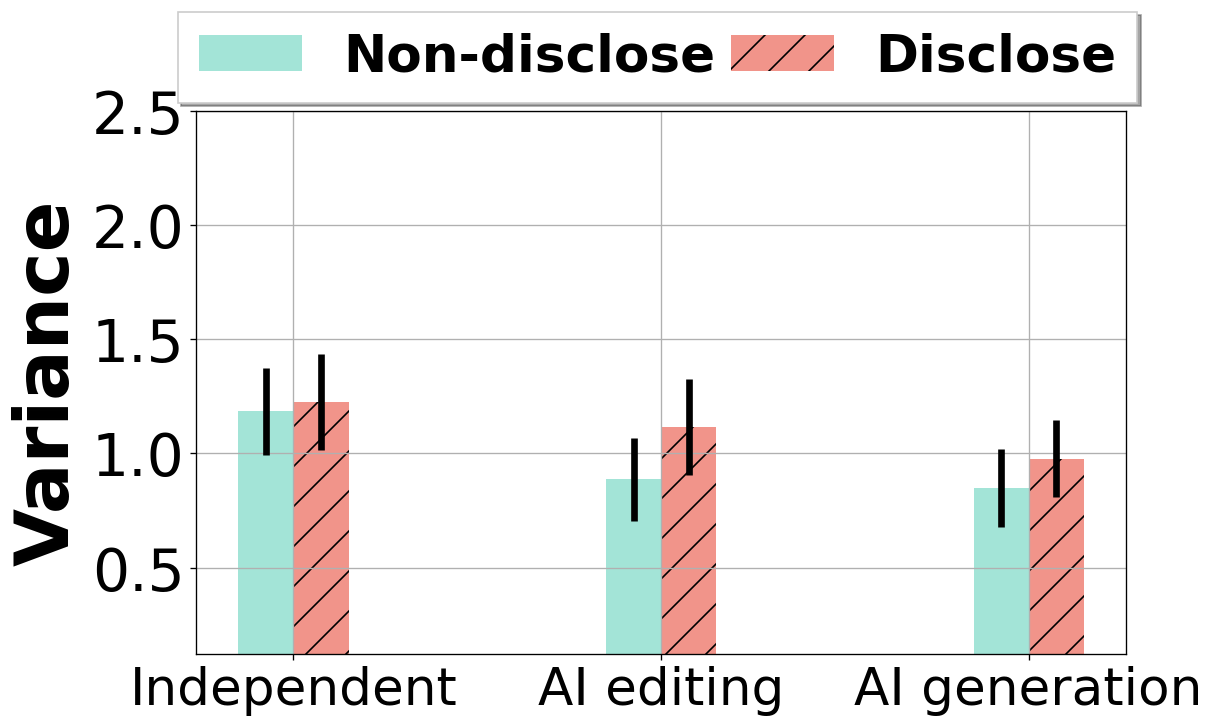}\label{fig:grammar_variance_story}}
  \vspace{-5pt}
  \caption{Comparing the {\em variance} in the ratings of grammar and vocabulary
given to the same article that was generated under the independent, AI editing, or AI generation writing modes, when the use and type of AI assistance during the writing process was or was not revealed to raters.  Error bars represent the 95\% confidence intervals of the variance.  }
\label{grammar_variance}
  \vspace{-15pt}
\end{figure}

\begin{figure}[t]
  \centering
  \subfloat[Argumentative essay]{\includegraphics[width=0.24\textwidth]{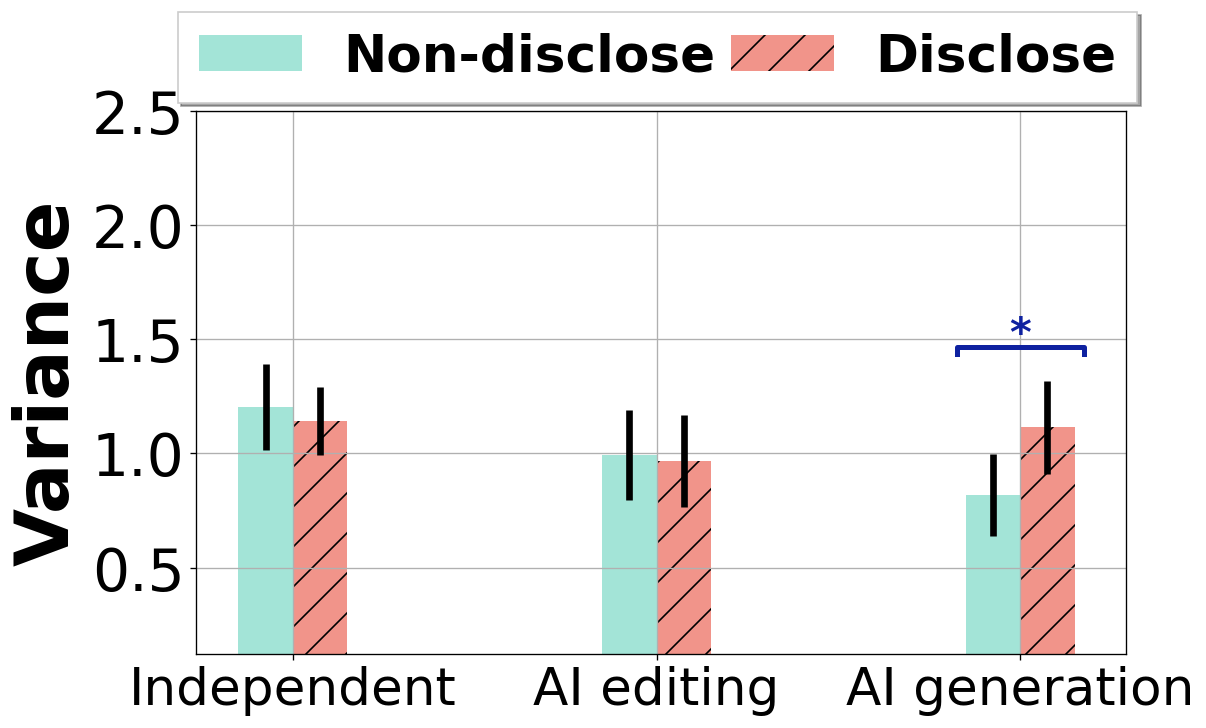}\label{fig:organ_variance_statement}}
  \hfill
  \subfloat[Creative story]{\includegraphics[width=0.24\textwidth]{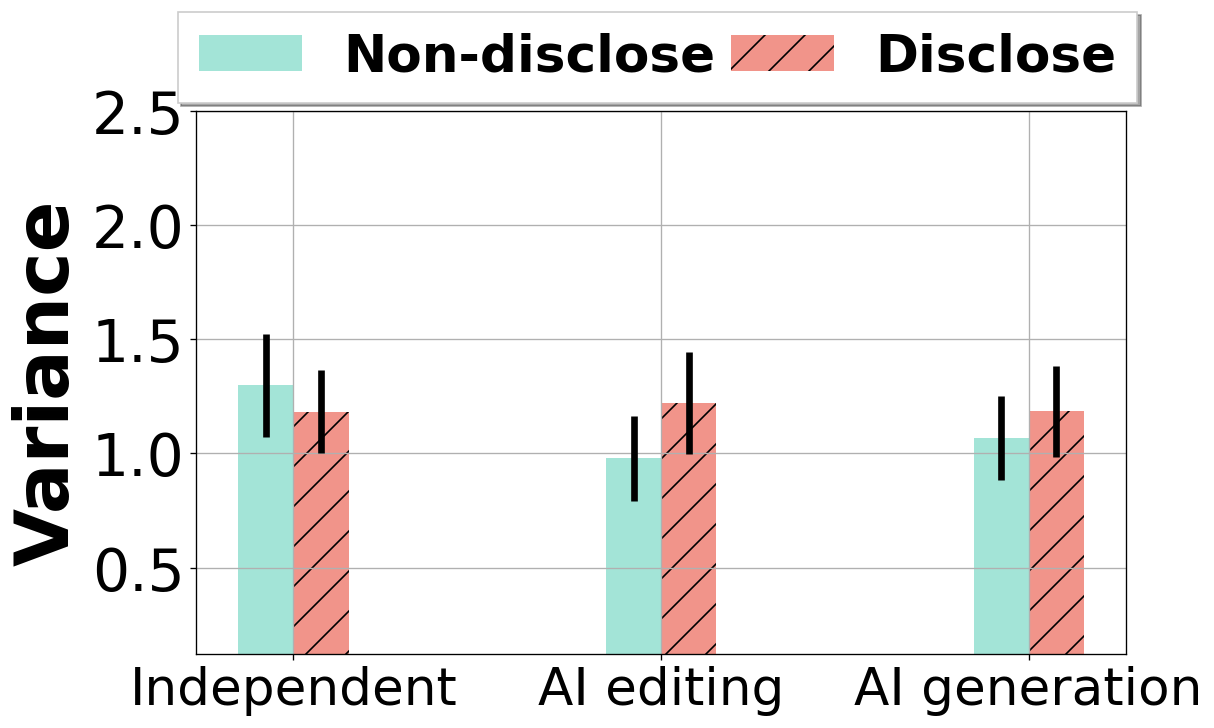}\label{fig:organ_variance_story}}
  \vspace{-5pt}
\caption{
Comparing the {\em variance in} the ratings of organization
given to the same article that was generated under the independent, AI editing, or AI generation writing modes, when the use and type of AI assistance during the writing process was or was not revealed to raters. Error bars represent the 95\% confidence intervals of the variance.  $\textsuperscript{*}$ denotes the significance level of $0.05$.
}\label{organ_variance}
  \vspace{-15pt}
\end{figure}

\begin{figure}[t]
  \centering
  \subfloat[Argumentative essay]{\includegraphics[width=0.24\textwidth]{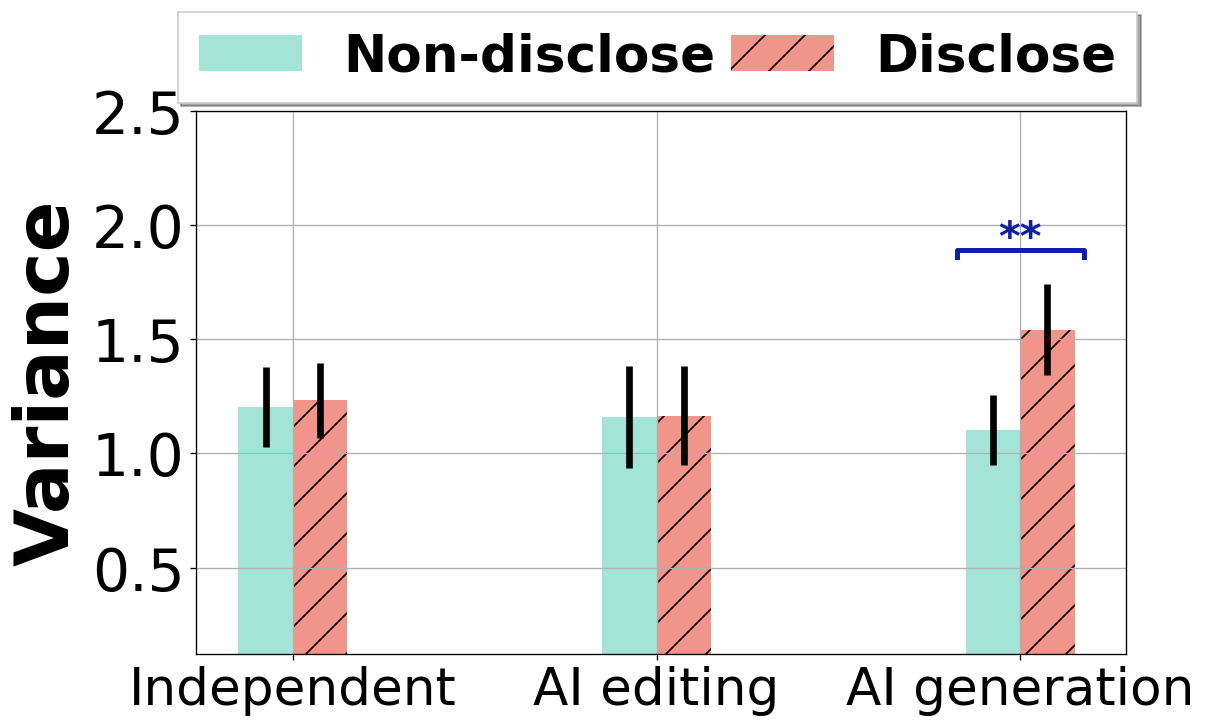}\label{fig:origin_variance_statement}}
  \hfill
  \subfloat[Creative story]{\includegraphics[width=0.24\textwidth]{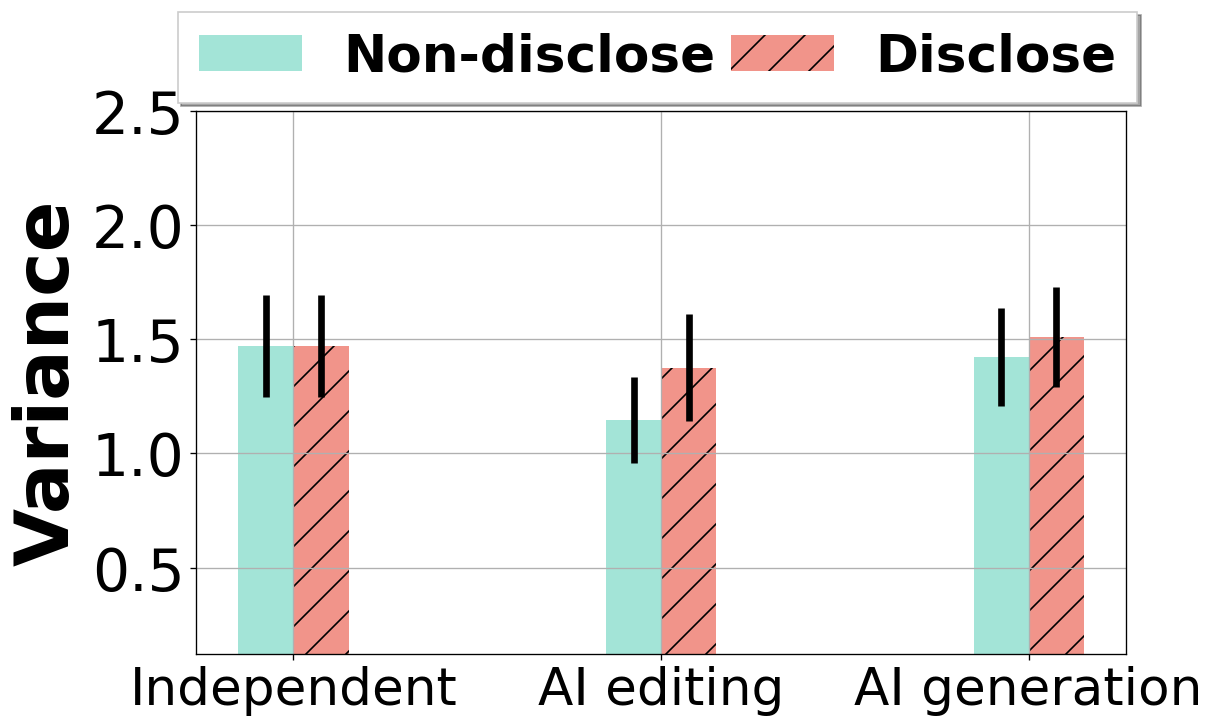}\label{fig:origin_variance_story}}
  \vspace{-5pt}
\caption{
Comparing the {\em variance} in the ratings of originality
given to the same article that was generated under the independent, AI editing, or AI generation writing modes, when the use and type of AI assistance during the writing process was or was not revealed to raters.
Error bars represent the 95\% confidence intervals of the variance. 
$\textsuperscript{**}$ denotes the significance level of $0.01$.
}\label{origin_variance}
  \vspace{-15pt}
\end{figure}

\begin{figure}[t]
  \centering
  \subfloat[Argumentative essay]{\includegraphics[width=0.24\textwidth]{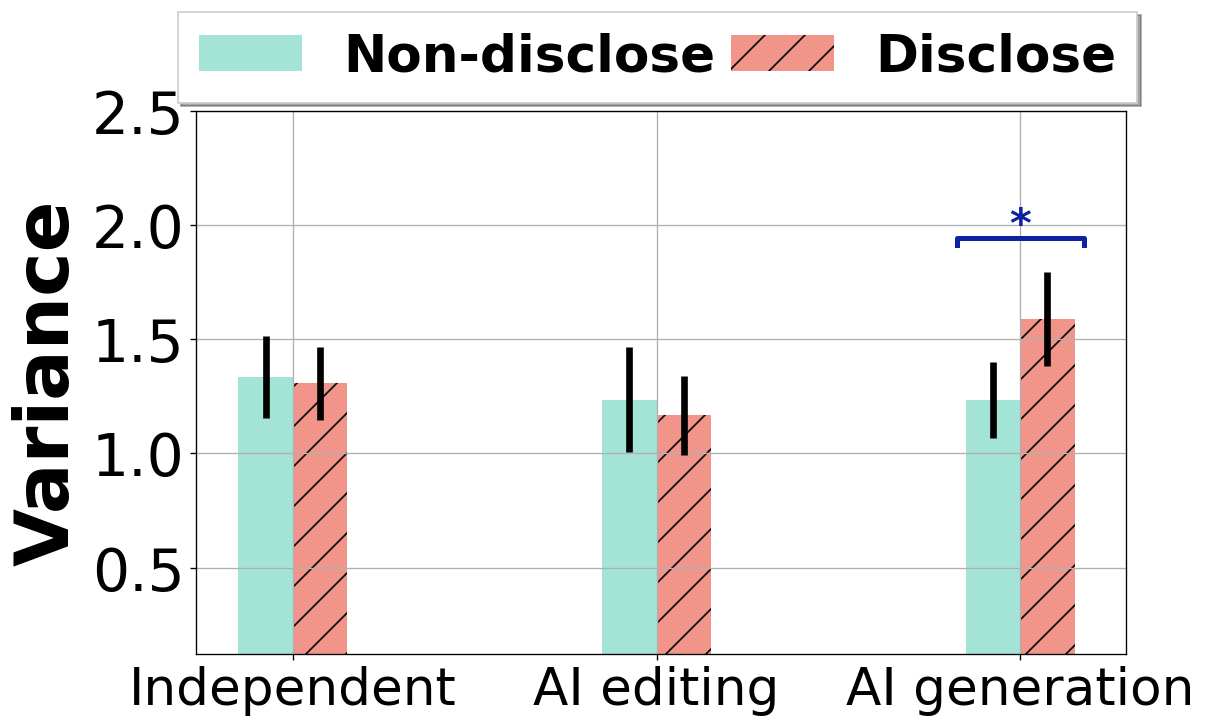}\label{fig:creat_variance_statement}}
  \hfill
  \subfloat[Creative story]{\includegraphics[width=0.24\textwidth]{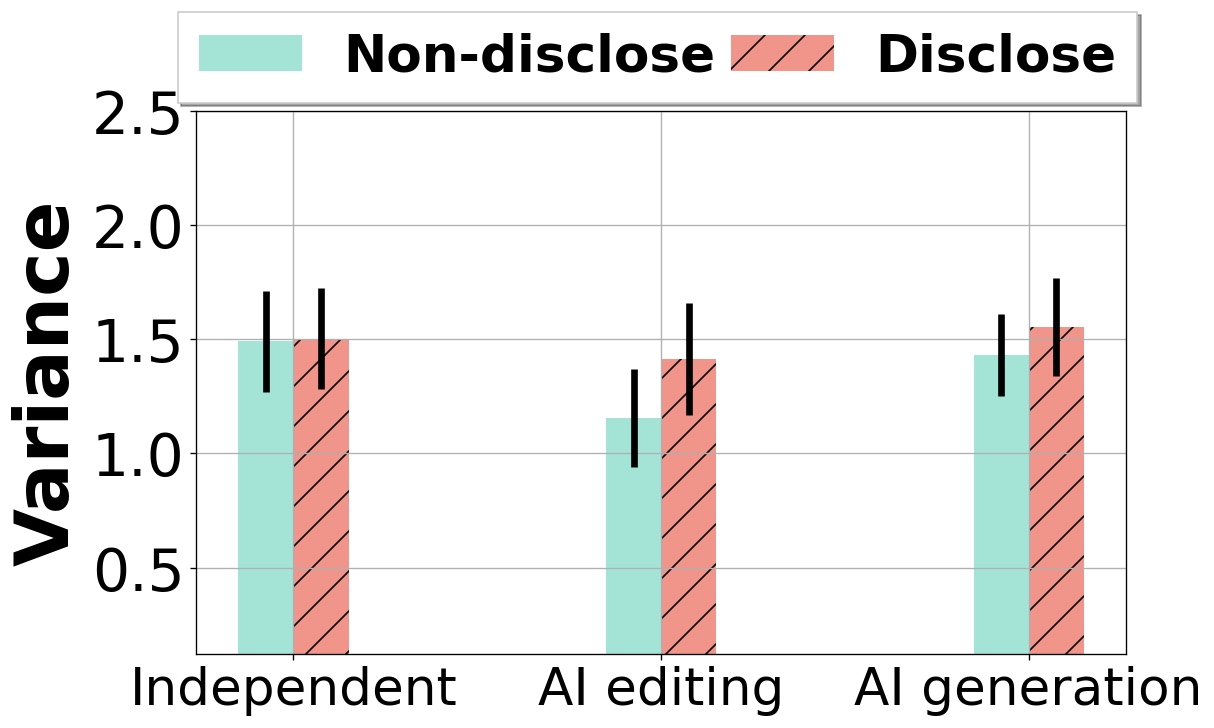}\label{fig:creat_variance_story}}
  \vspace{-5pt}
\caption{
Comparing the {\em variance} in the ratings of creativity
given to the same article that was generated under the independent, AI editing, or AI generation writing modes, when the use and type of AI assistance during the writing process was or was not revealed to raters. Error bars represent the 95\% confidence intervals of the variance.  $\textsuperscript{*}$ denotes the significance level of $0.05$.
}\label{creat_variance}
  \vspace{-15pt}
\end{figure}

\begin{figure}[t]
  \centering
  \subfloat[Argumentative essay]{\includegraphics[width=0.24\textwidth]{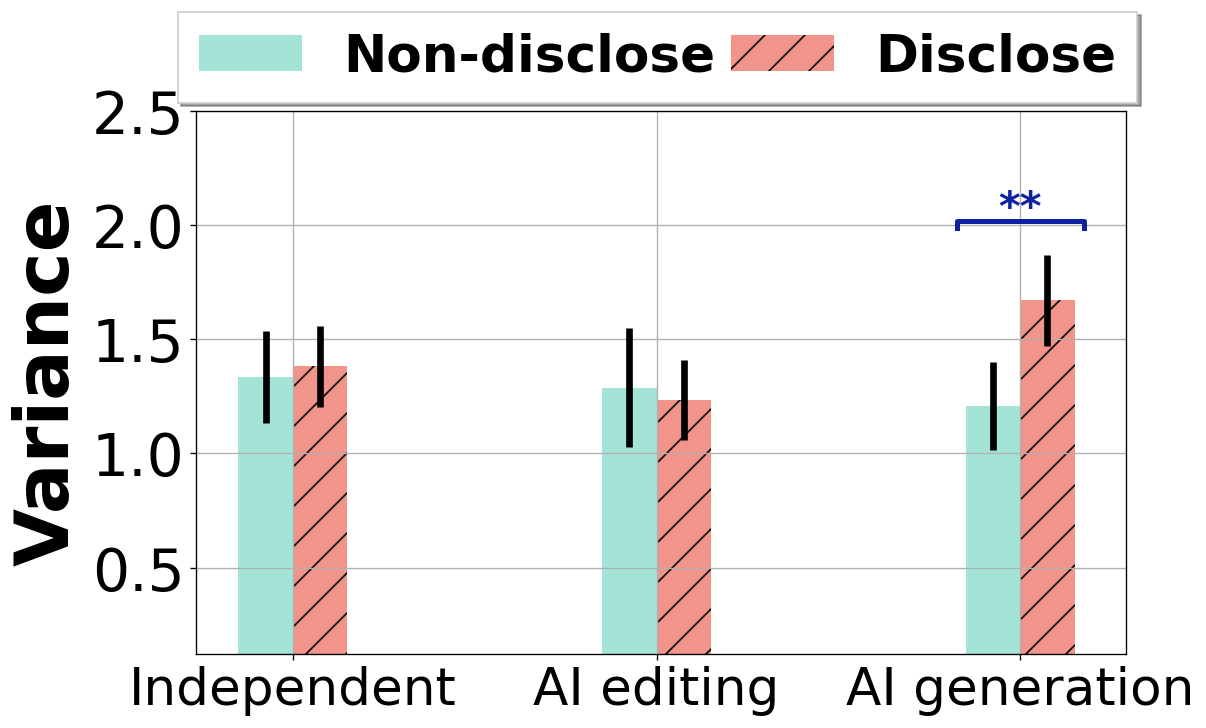}\label{fig:emotion_variance_statement}}
  \hfill
  \subfloat[Creative story]{\includegraphics[width=0.24\textwidth]{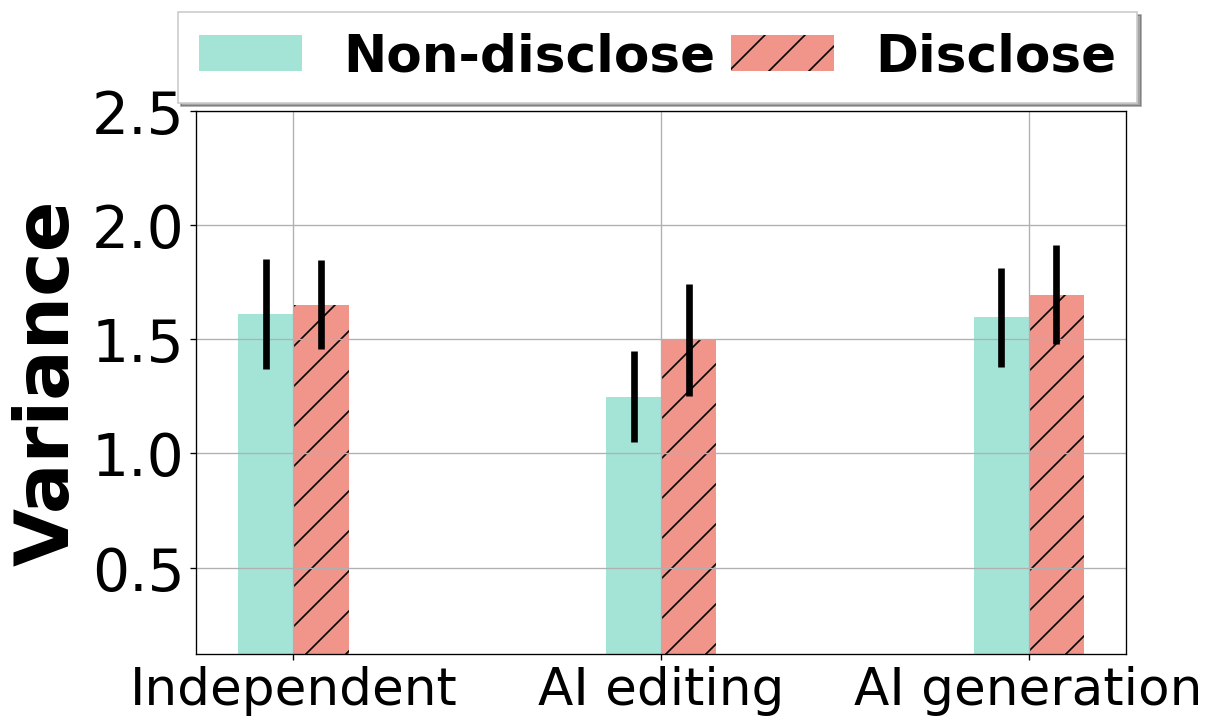}\label{fig:emotion_variance_story}}
  \vspace{-5pt}
\caption{
Comparing the {\em variance} in the ratings of emotion authenticity
given to the same article that was generated under the independent, AI editing, or AI generation writing modes, when the use and type of AI assistance during the writing process was or was not revealed to raters.
Error bars represent the 95\% confidence intervals of the variance.  $\textsuperscript{**}$ denotes the significance level of $0.01$.
}\label{emotion_variance}
  \vspace{-15pt}
\end{figure}

Figures~\ref{rec_variance}--\ref{emotion_variance}
compare the variance in participants' willingness to shortlist an article, and in their detailed evaluations on the grammar and vocabulary, organization, originality, creativity, and emotion authenticity of the same argumentative essay or the same creative story that was generated under the three writing modes. We found that when participants' were informed about that the author used ChatGPT to assist them in generating some content of their argumentative essay, their ratings on the essay's organization, originality, creativity, and emotion authenticity have a significantly higher level of variance than the case when participants were unaware of the AI assistance.

\section{Individual Heterogeneity in the Impacts of AI
Disclosure on the Quality Evaluation (Additional Results)} \label{heter_sm}

\begin{figure}[t]
  \centering
  \subfloat[Argumentative essay]{\includegraphics[width=0.24\textwidth]{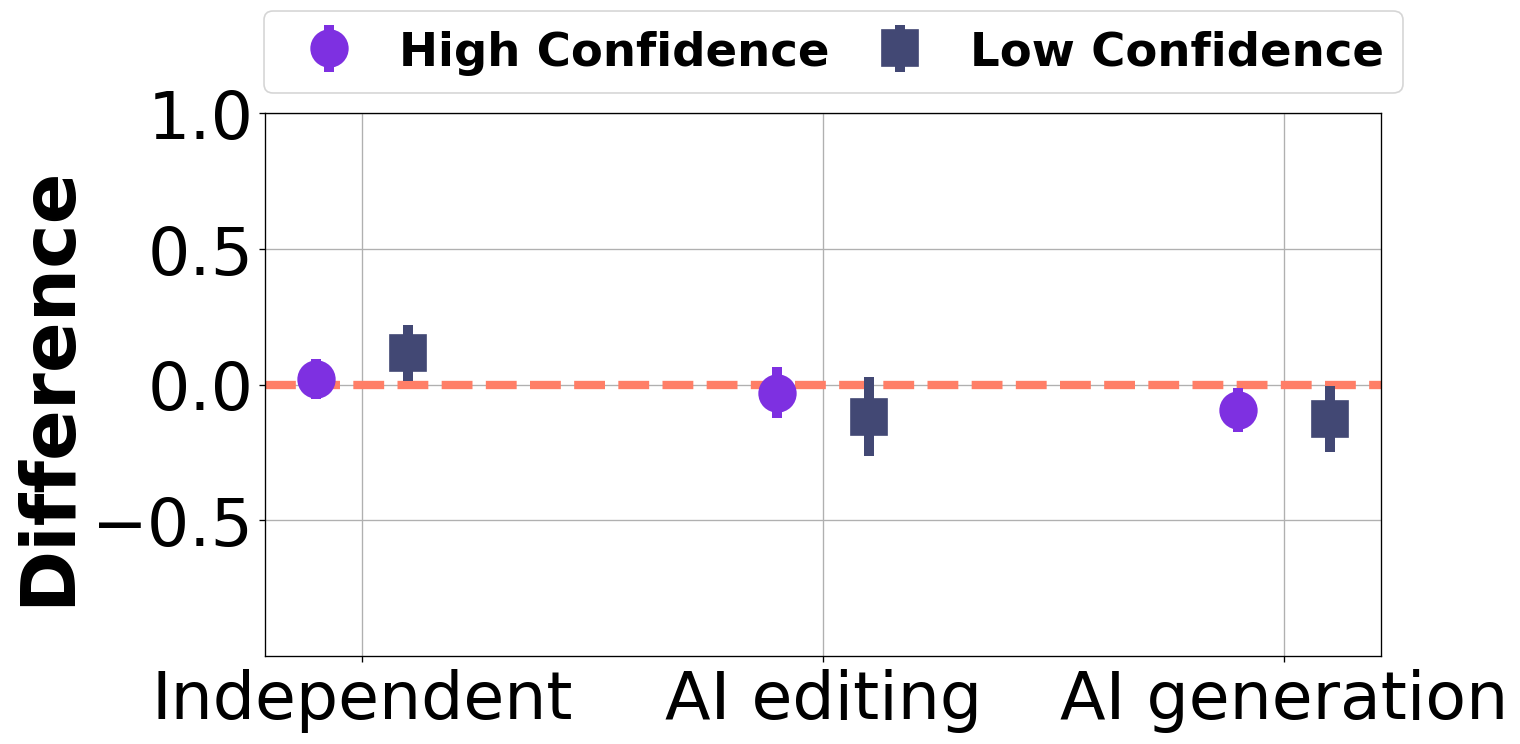}\label{fig:confidence_rec_statement}}
  \hfill
  \subfloat[Creative story]{\includegraphics[width=0.24\textwidth]{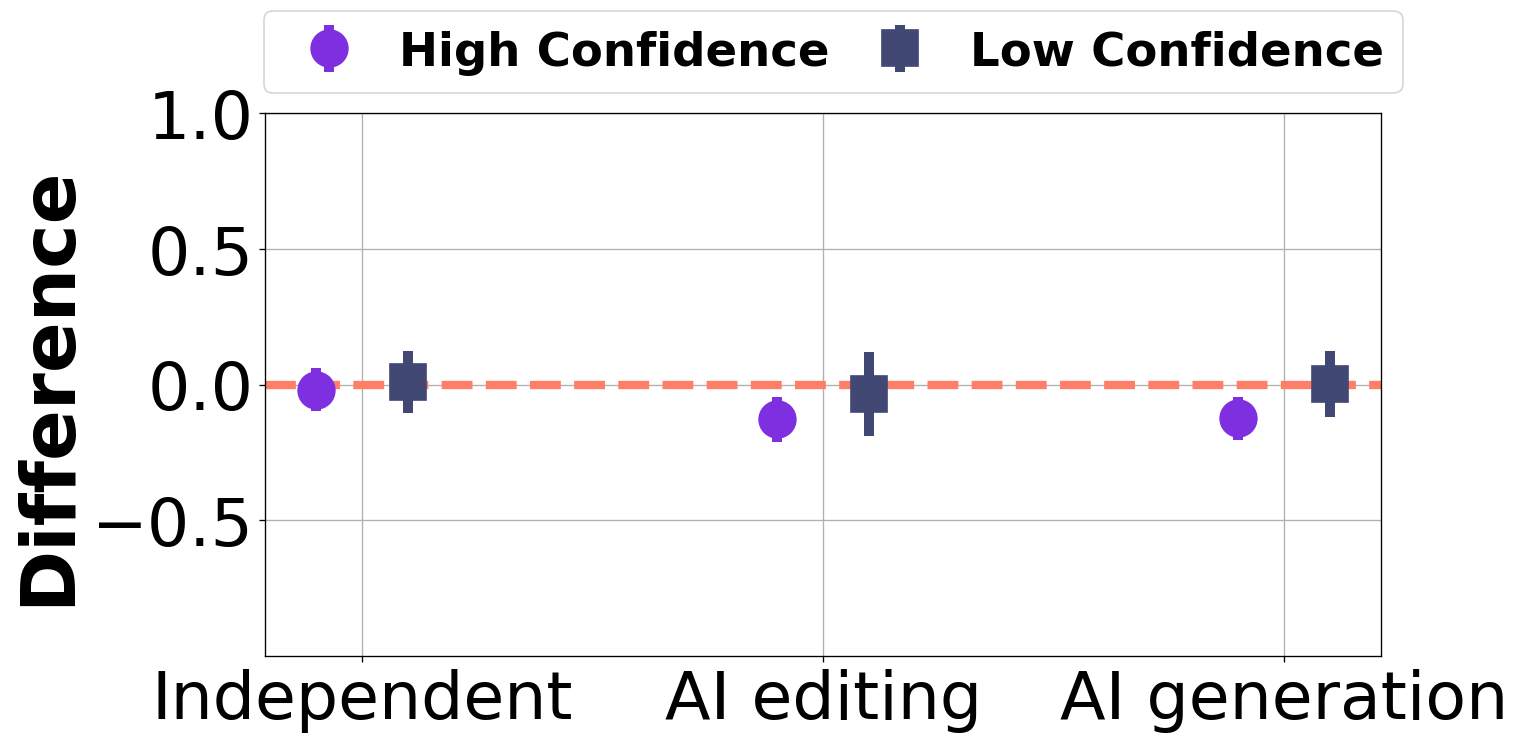}\label{fig:confidence_rec_story}}
  \vspace{-5pt}
\caption{
The difference in an article's shortlisting rates received in the ``{\em Disclose}'' treatment and those received in the ``{\em Non-Disclose}'' treatment,  when the ratings were provided by participants who had high or low {\em confidence in their own writing skills}.  Error bars represent the 95\% bootstrap confidence intervals of the rating difference. An interval below zero means the corresponding group of raters decrease their ratings when the use and type of AI assistance in the writing process was revealed to them.
}\label{confidence_rec}
  \vspace{-15pt}
\end{figure}

\begin{figure}[t]
  \centering
  \subfloat[Argumentative essay]{\includegraphics[width=0.24\textwidth]{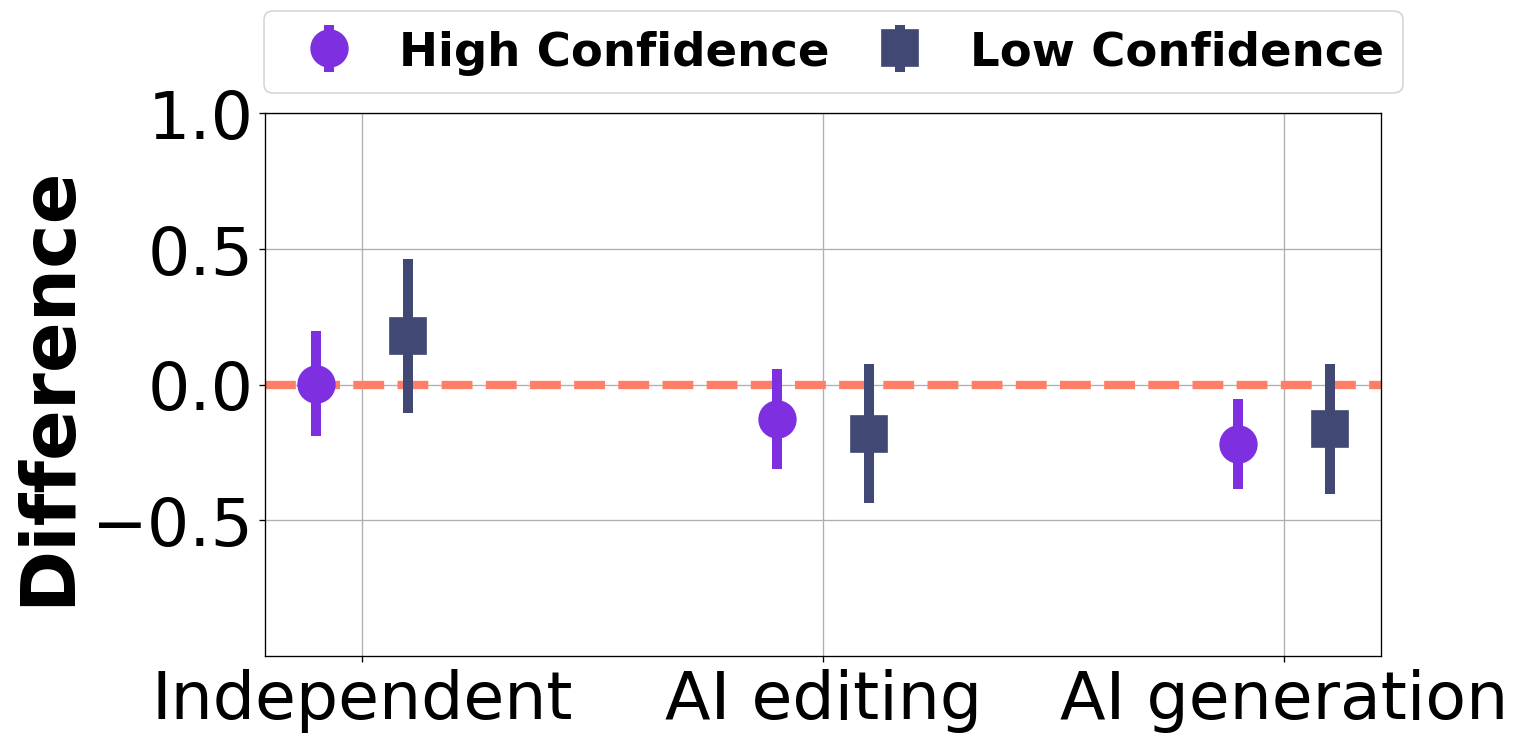}\label{fig:confidence_grammar_statement}}
  \hfill
  \subfloat[Creative story]{\includegraphics[width=0.24\textwidth]{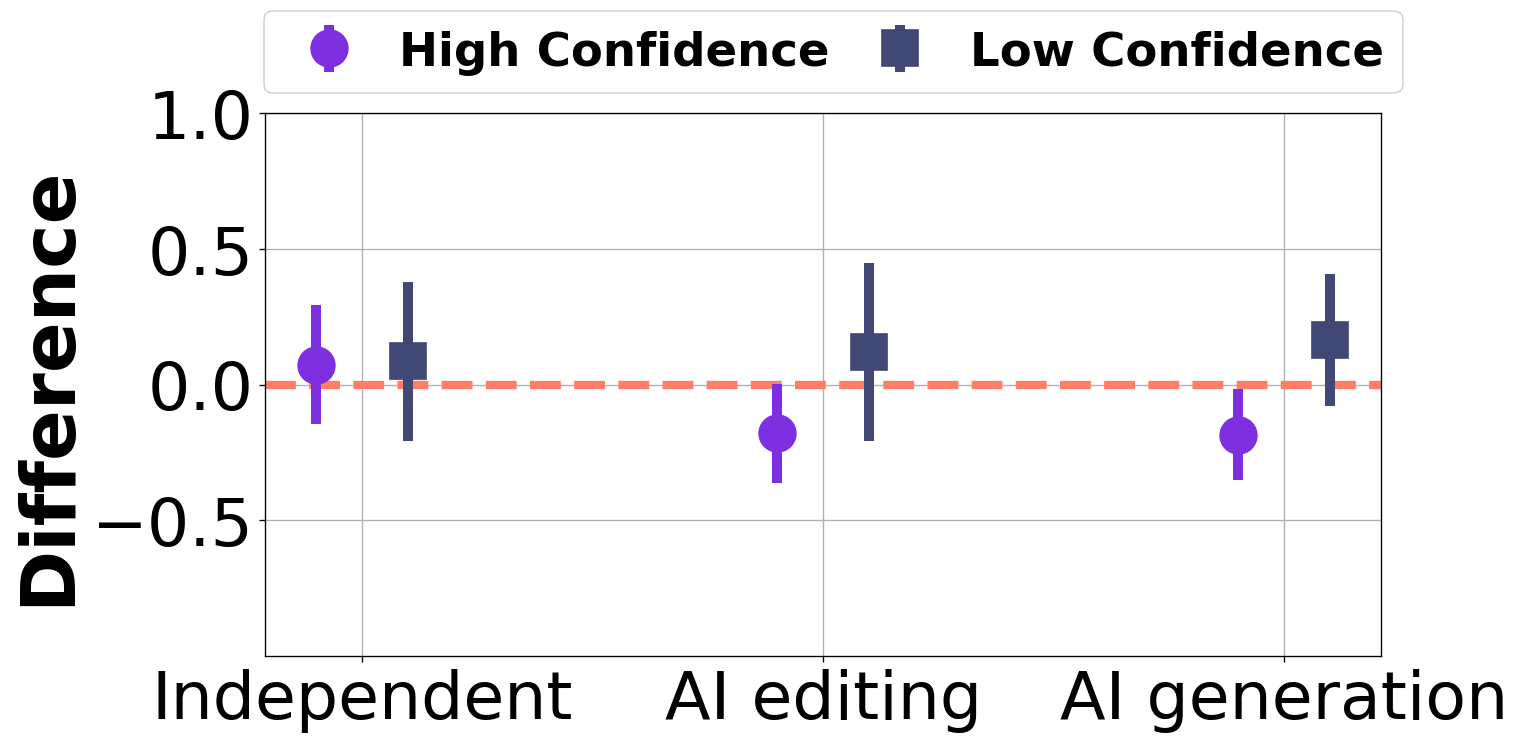}\label{fig:confidence_grammar_story}}
  \vspace{-5pt}
\caption{
The difference in an article's grammar and vocabulary ratings received in the ``{\em Disclose}'' treatment and those received in the ``{\em Non-Disclose}'' treatment,  when the ratings were provided by participants who had high or low {\em confidence in their own writing skills}.  Error bars represent the 95\% bootstrap confidence intervals of the rating difference. An interval below zero means the corresponding group of raters decrease their ratings when the use and type of AI assistance in the writing process was revealed to them.
}\label{confidence_grammar}
  \vspace{-15pt}
\end{figure}

\begin{figure}[t]
  \centering
  \subfloat[Argumentative essay]{\includegraphics[width=0.24\textwidth]{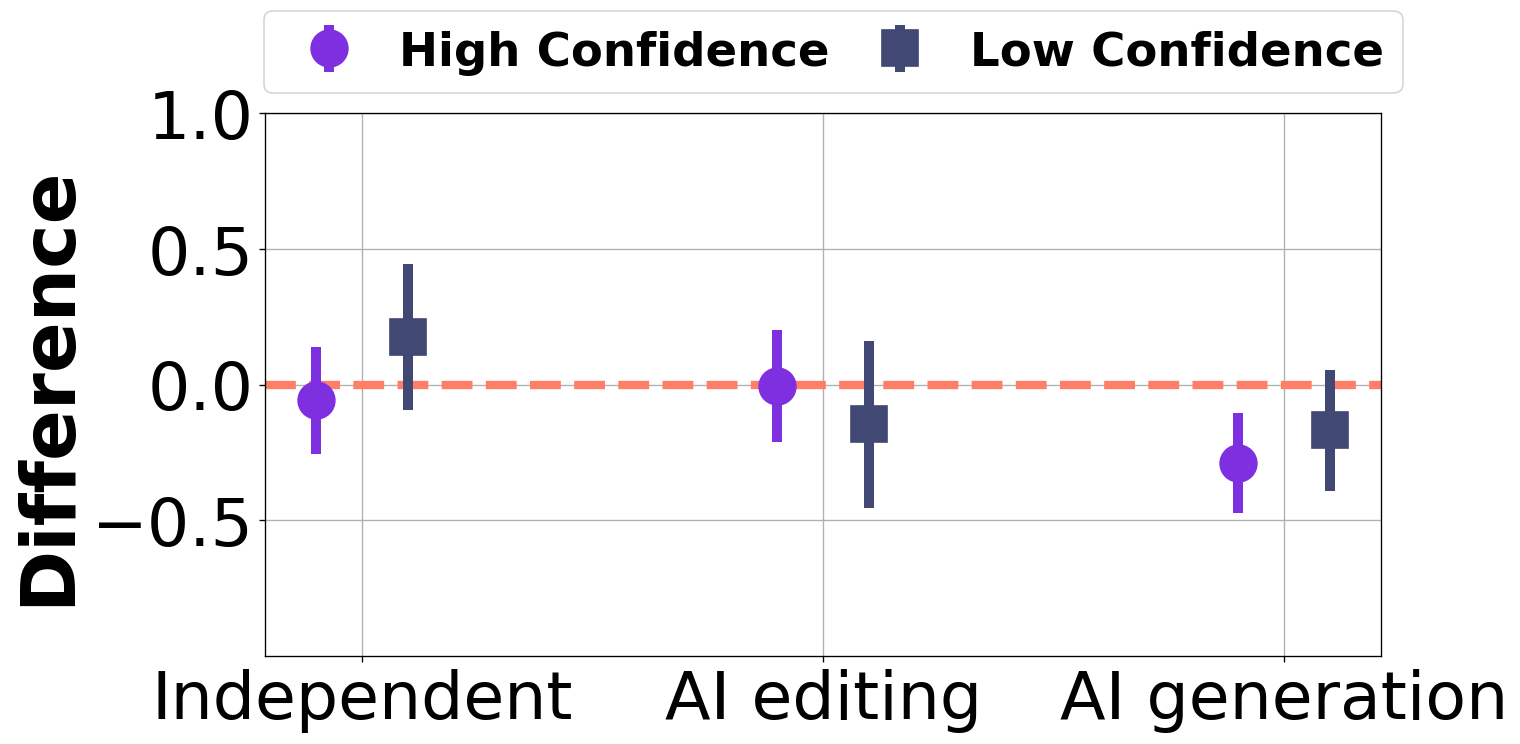}\label{fig:confidence_organization_statement}}
  \hfill
  \subfloat[Creative story]{\includegraphics[width=0.24\textwidth]{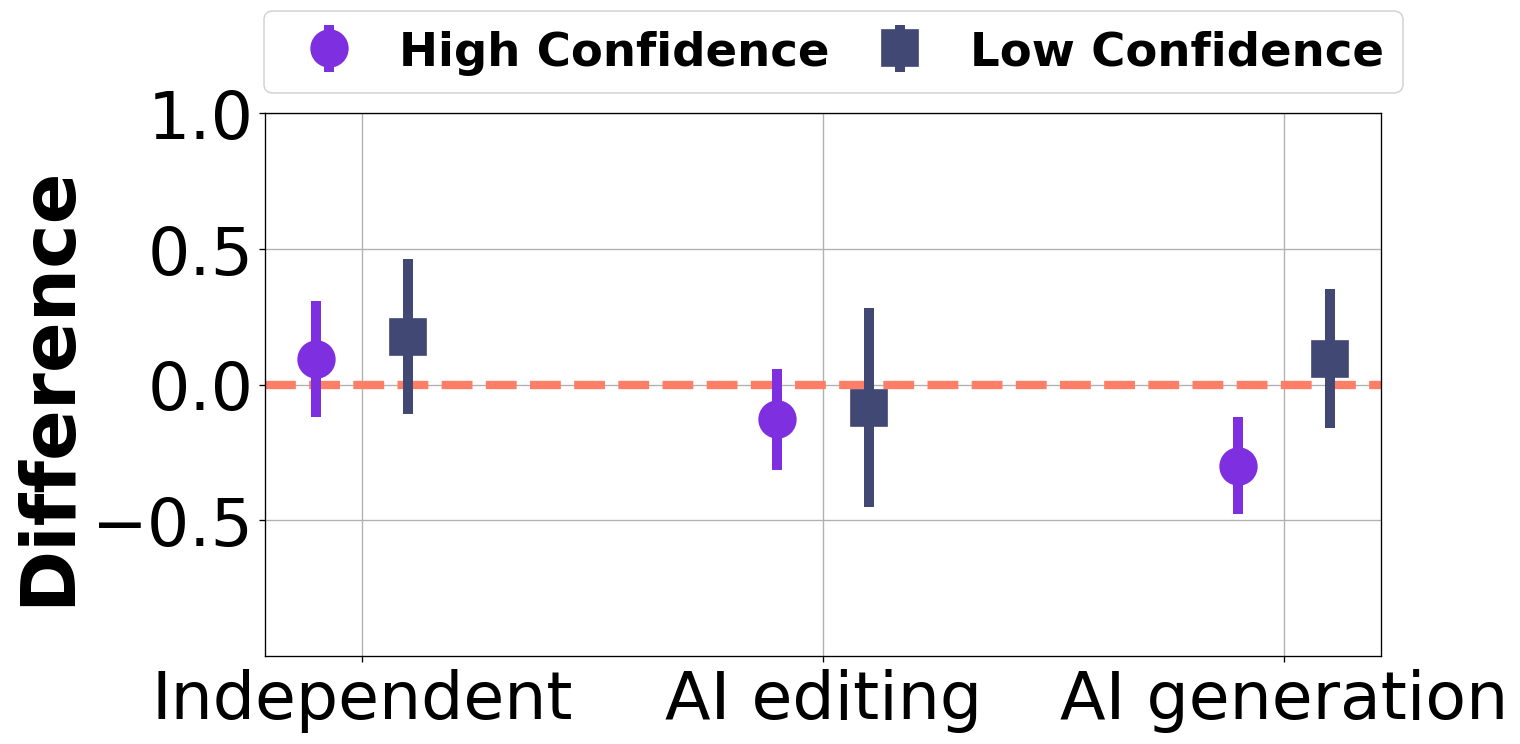}\label{fig:confidence_organization_story}}
  \vspace{-5pt}
\caption{
The difference in an article's organization ratings received in the ``{\em Disclose}'' treatment and those received in the ``{\em Non-Disclose}'' treatment,  when the ratings were provided by participants who had high or low {\em confidence in their own writing skills}.  Error bars represent the 95\% bootstrap confidence intervals of the rating difference. An interval below zero means the corresponding group of raters decrease their ratings when the use and type of AI assistance in the writing process was revealed to them.
}\label{confidence_organ}
  \vspace{-15pt}
\end{figure}

\begin{figure}[t]
  \centering
  \subfloat[Argumentative essay]{\includegraphics[width=0.24\textwidth]{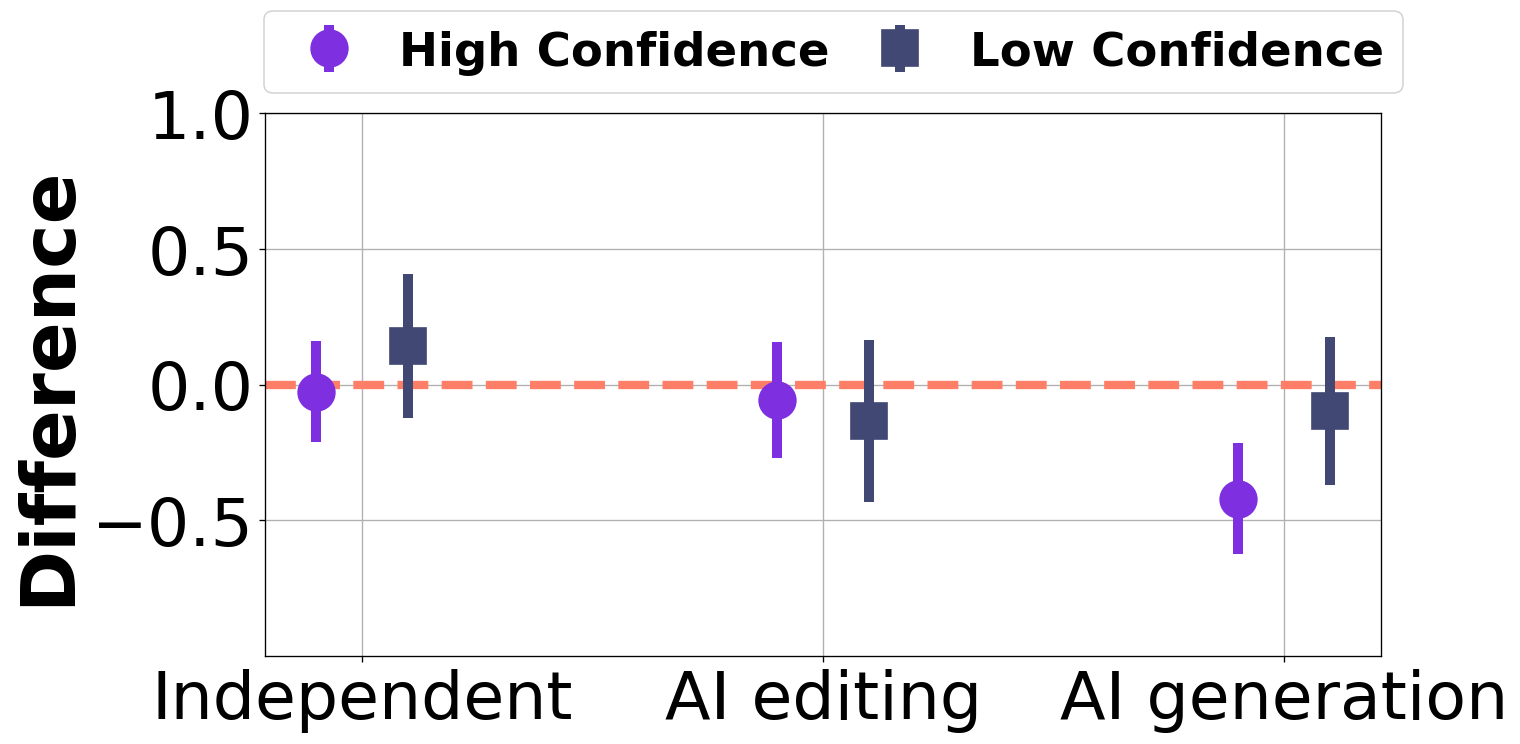}\label{fig:confidence_origin_statement}}
  \hfill
  \subfloat[Creative story]{\includegraphics[width=0.24\textwidth]{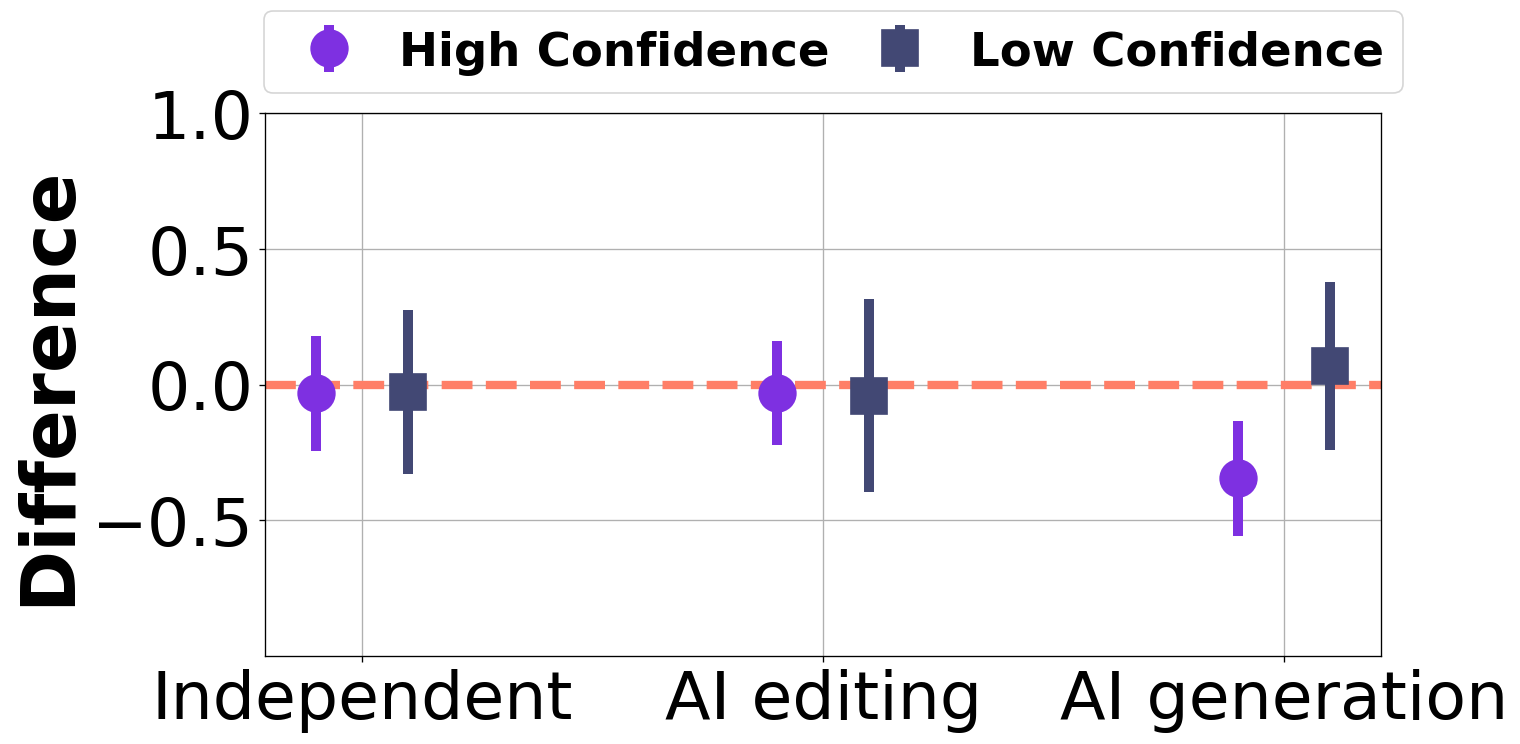}\label{fig:confidence_origin_story}}
  \vspace{-5pt}
\caption{
The difference in an article's originality ratings received in the ``{\em Disclose}'' treatment and those received in the ``{\em Non-Disclose}'' treatment,  when the ratings were provided by participants who had high or low {\em confidence in their own writing skills}.  Error bars represent the 95\% bootstrap confidence intervals of the rating difference. An interval below zero means the corresponding group of raters decrease their ratings when the use and type of AI assistance in the writing process was revealed to them.
}\label{confidence_origin}
  \vspace{-15pt}
\end{figure}

\begin{figure}[t]
  \centering
  \subfloat[Argumentative essay]{\includegraphics[width=0.24\textwidth]{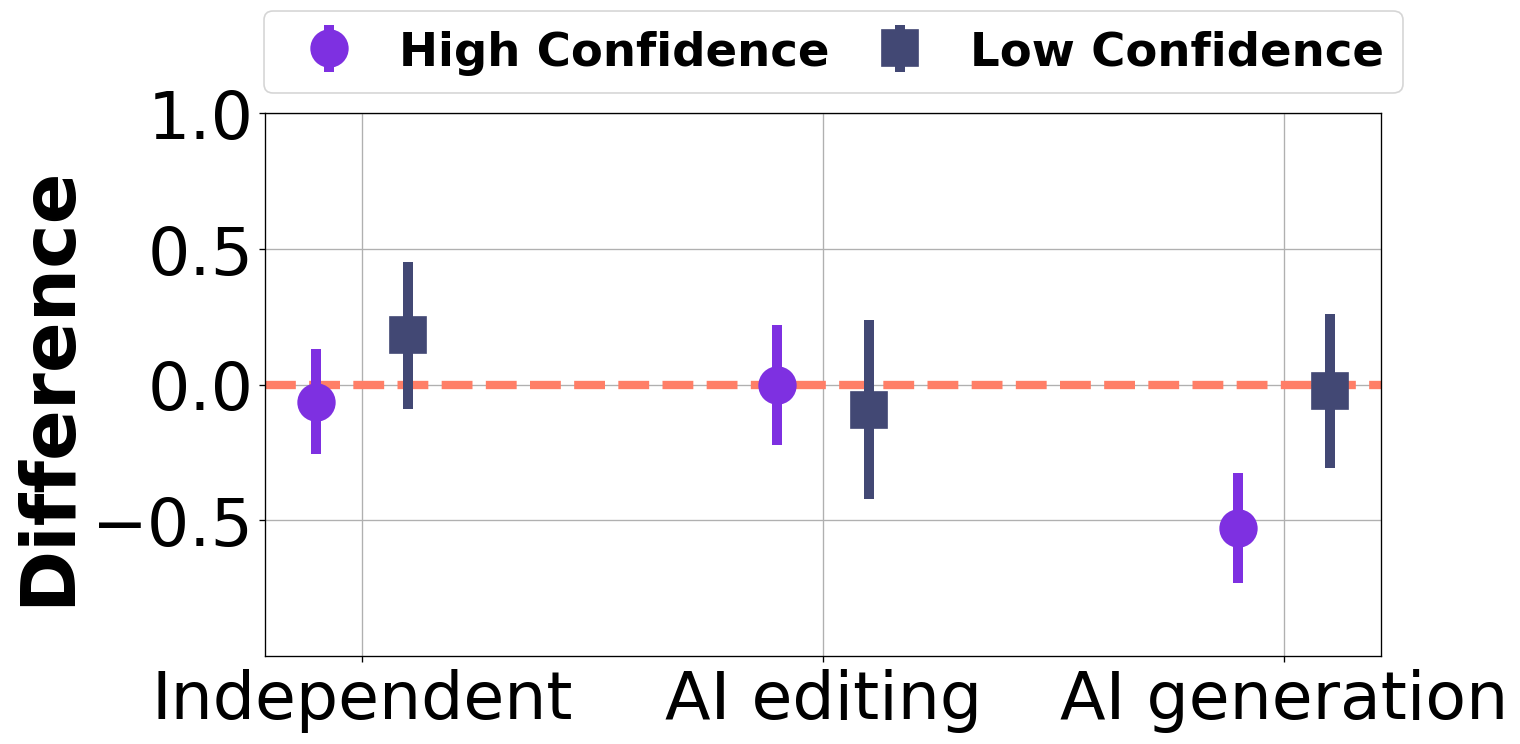}\label{fig:confidence_creat_statement}}
  \hfill
  \subfloat[Creative story]{\includegraphics[width=0.24\textwidth]{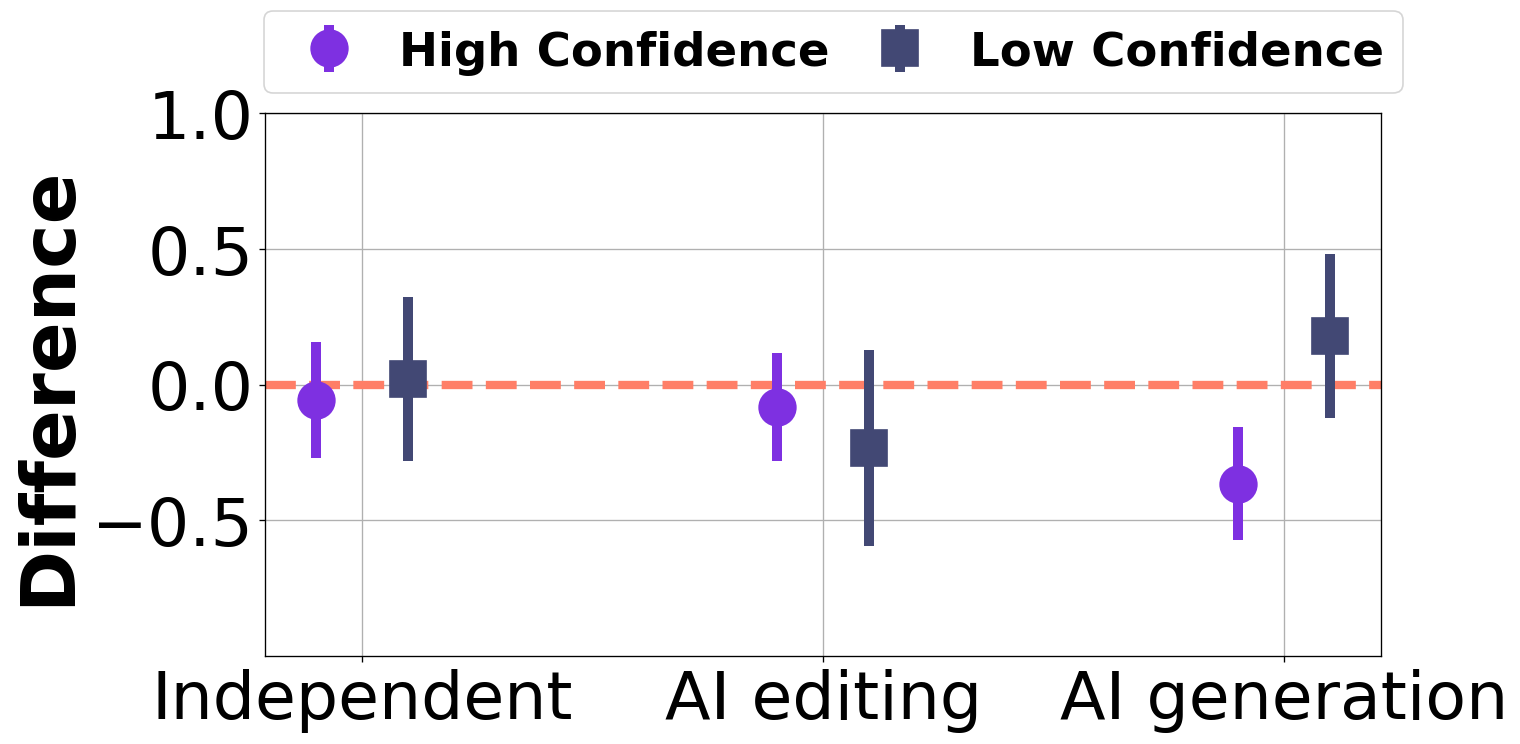}\label{fig:confidence_creat_story}}
  \vspace{-5pt}
\caption{
The difference in an article's  creativity ratings received in the ``{\em Disclose}'' treatment and those received in the ``{\em Non-Disclose}'' treatment,  when the ratings were provided by participants who had high or low {\em confidence in their own writing skills}.  Error bars represent the 95\% bootstrap confidence intervals of the rating difference. An interval below zero means the corresponding group of raters decrease their ratings when the use and type of AI assistance in the writing process was revealed to them.
}\label{confidence_creat}
  \vspace{-15pt}
\end{figure}

\begin{figure}[t]
  \centering
  \subfloat[Argumentative essay]{\includegraphics[width=0.24\textwidth]{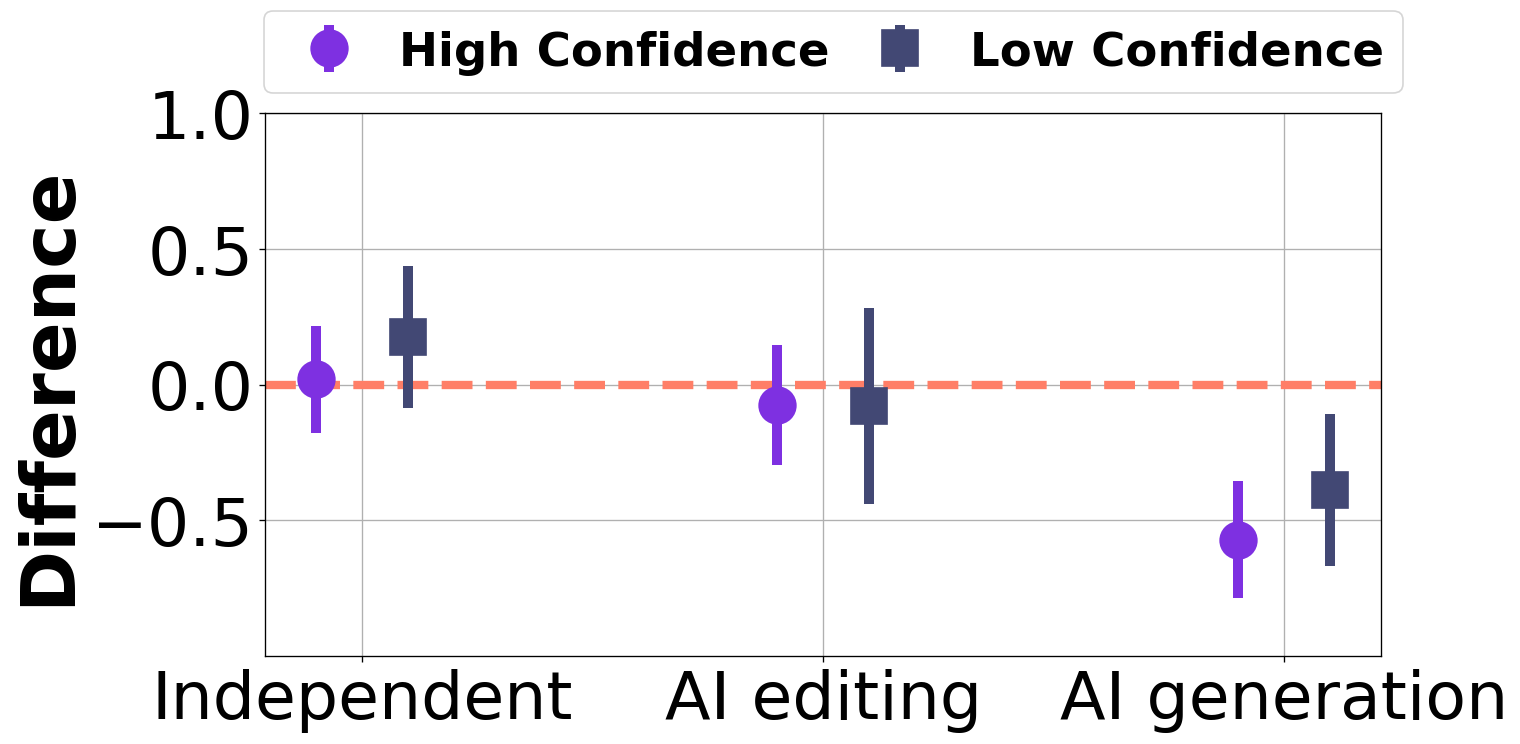}\label{fig:confidence_emotion_statement}}
  \hfill
  \subfloat[Creative story]{\includegraphics[width=0.24\textwidth]{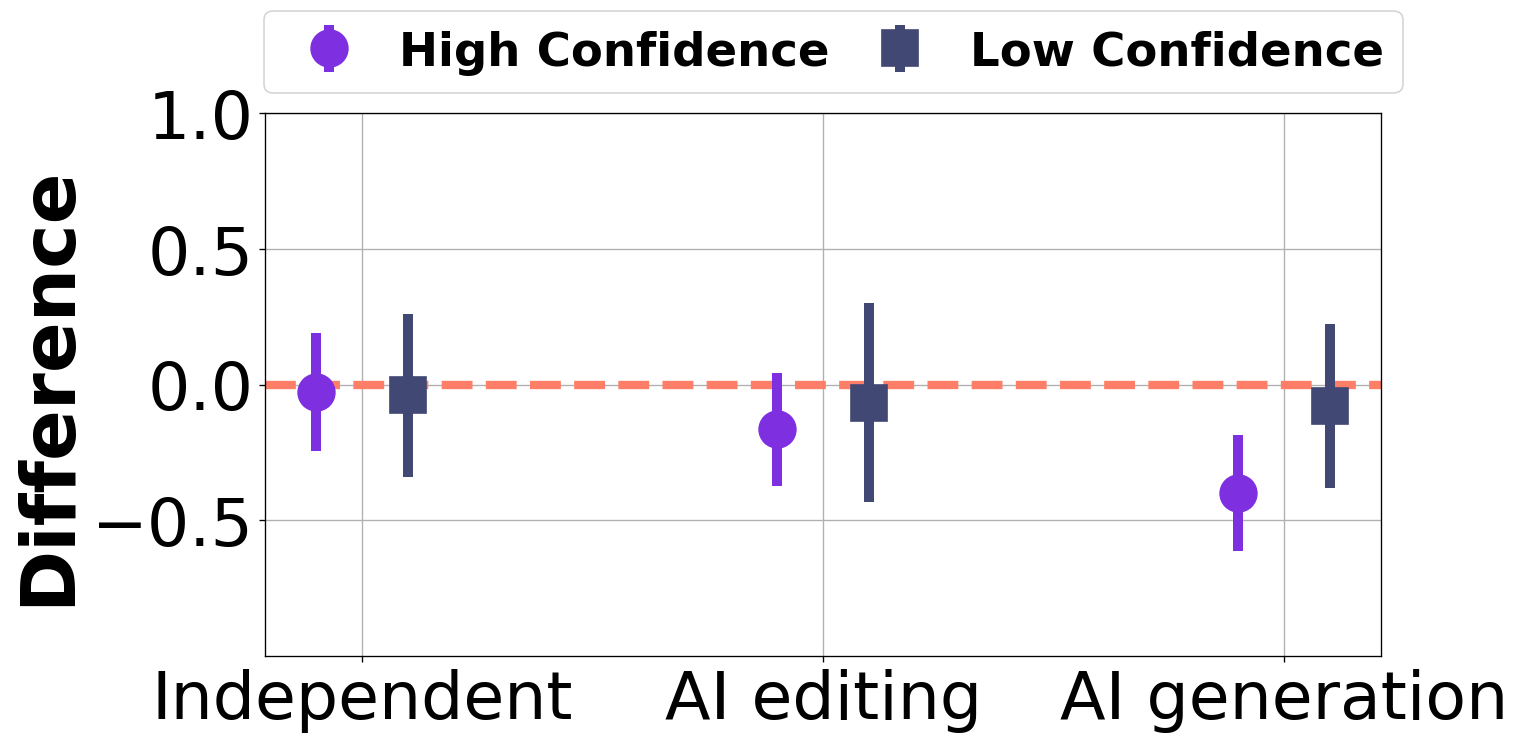}\label{fig:confidence_emotion_story}}
  \vspace{-5pt}
\caption{
The difference in an article's emotion authenticity ratings received in the ``{\em Disclose}'' treatment and those received in the ``{\em Non-Disclose}'' treatment,  when the ratings were provided by participants who had high or low {\em confidence in their own writing skills}.  Error bars represent the 95\% bootstrap confidence intervals of the rating difference. An interval below zero means the corresponding group of raters decrease their ratings when the use and type of AI assistance in the writing process was revealed to them.
}\label{confidence_emotion}
  \vspace{-15pt}
\end{figure}

\begin{figure}[t]
  \centering
  \subfloat[Argumentative essay]{\includegraphics[width=0.24\textwidth]{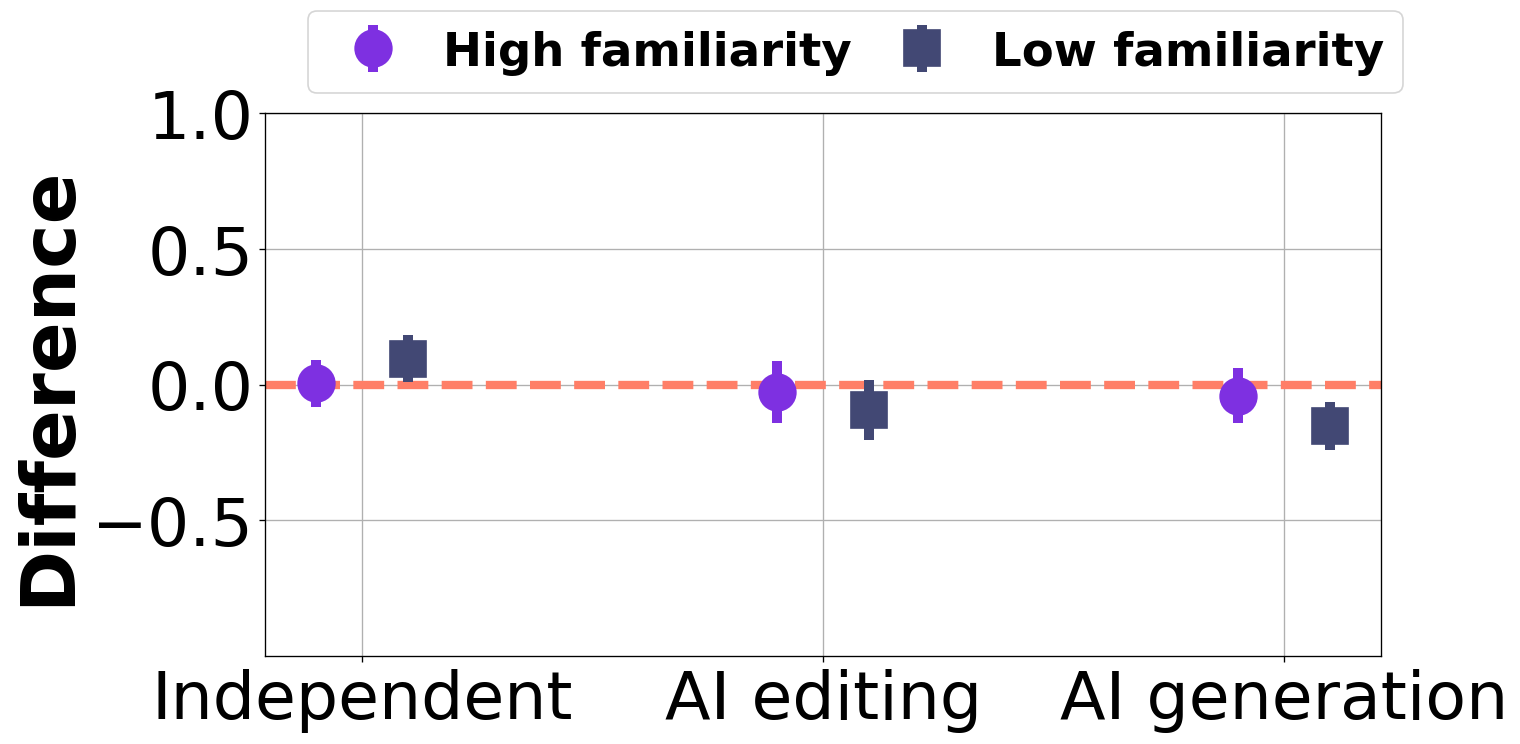}\label{fig:fam_rec_statement}}
  \hfill
  \subfloat[Creative story]{\includegraphics[width=0.24\textwidth]{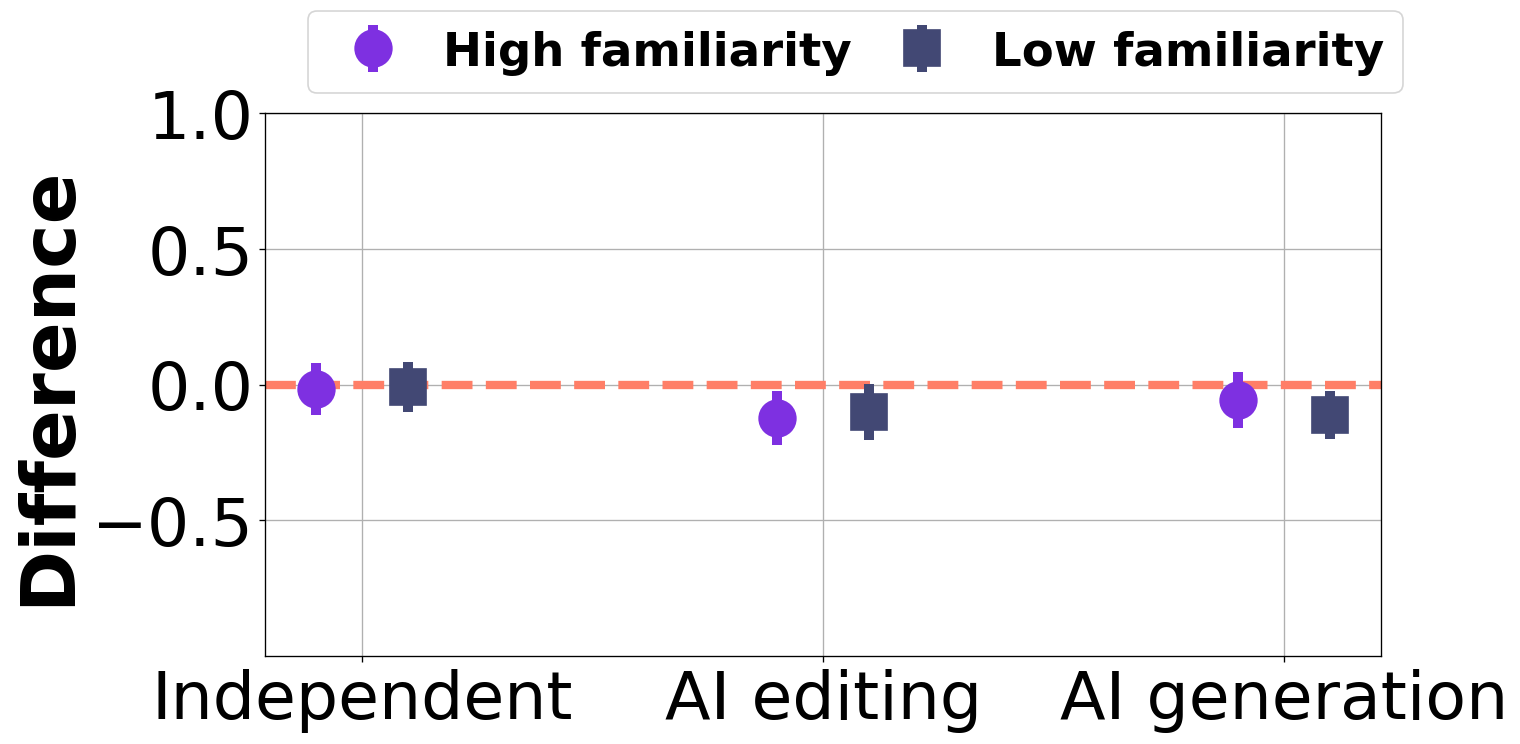}\label{fig:fam_rec_story}}
  \vspace{-5pt}
\caption{
The difference in an article's shortlisting rates received in the ``{\em Disclose}'' treatment and those received in the ``{\em Non-Disclose}'' treatment, when the ratings were provided by participants who had high or low {\em familiarity with ChatGPT}. Error bars represent the 95\% bootstrap confidence intervals of the rating difference. An interval below zero means the corresponding group of raters decrease their ratings when the use and type of AI assistance in the writing process was revealed to them. 
}\label{fam_rec}
  \vspace{-15pt}
\end{figure}

\begin{figure}[t]
  \centering
  \subfloat[Argumentative essay]{\includegraphics[width=0.24\textwidth]{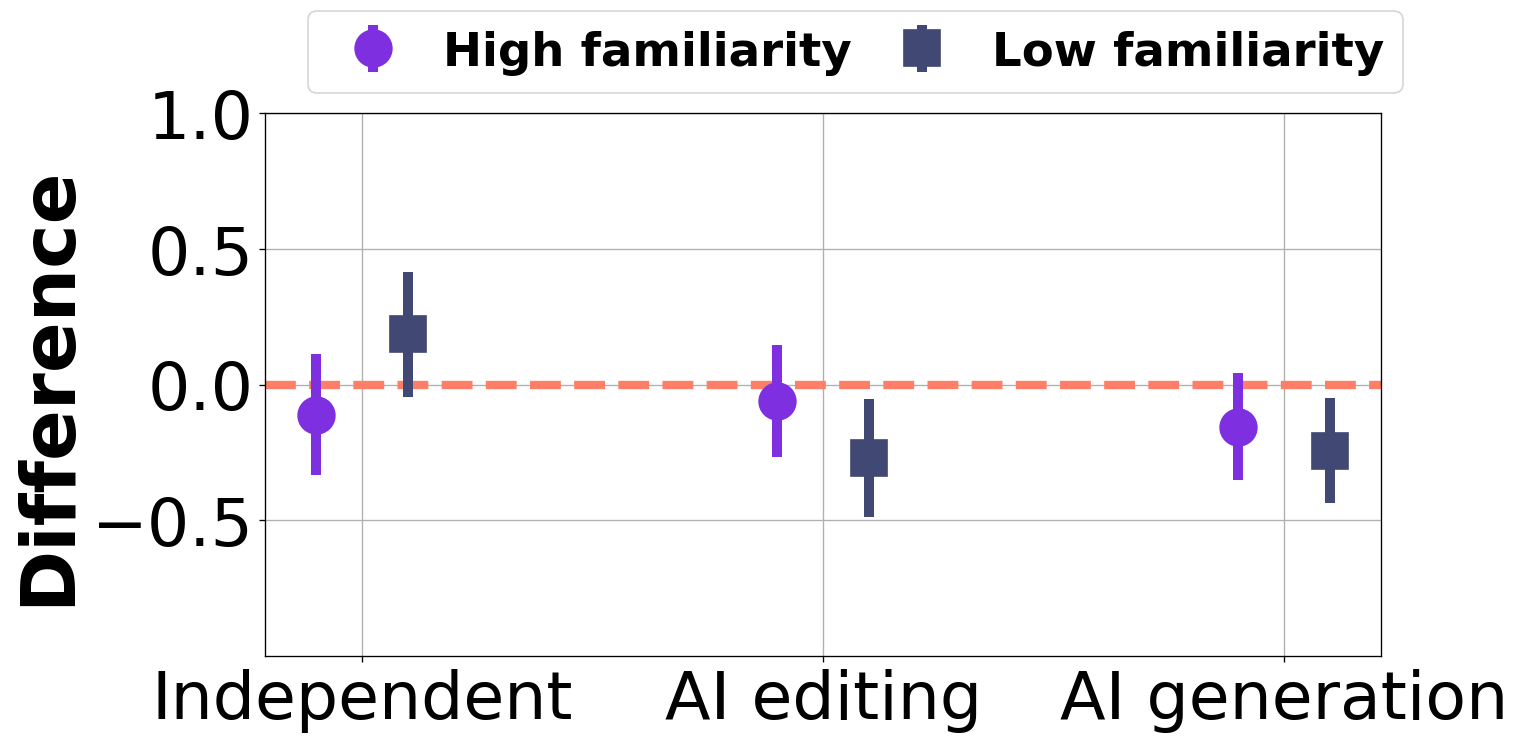}\label{fig:fam_grammar_statement}}
  \hfill
  \subfloat[Creative story]{\includegraphics[width=0.24\textwidth]{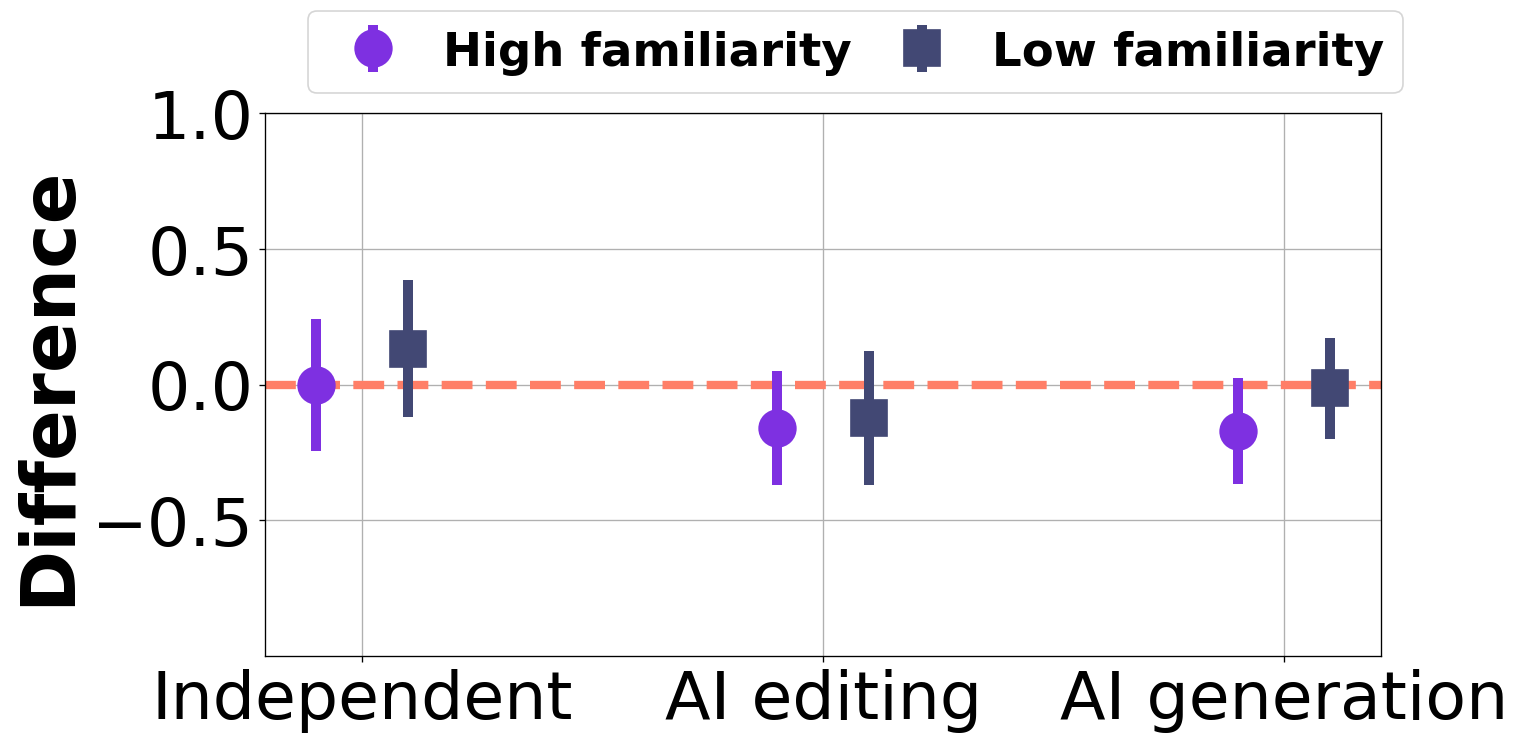}\label{fig:fam_grammar_story}}
  \vspace{-5pt}
\caption{
The difference in an article's  grammar and vocabulary ratings received in the ``{\em Disclose}'' treatment and those received in the ``{\em Non-Disclose}'' treatment, when the ratings were provided by participants who had high or low {\em familiarity with ChatGPT}. Error bars represent the 95\% bootstrap confidence intervals of the rating difference. An interval below zero means the corresponding group of raters decrease their ratings when the use and type of AI assistance in the writing process was revealed to them. 
}\label{fam_grammar}
  \vspace{-15pt}
\end{figure}

\begin{figure}[t]
  \centering
  \subfloat[Argumentative essay]{\includegraphics[width=0.24\textwidth]{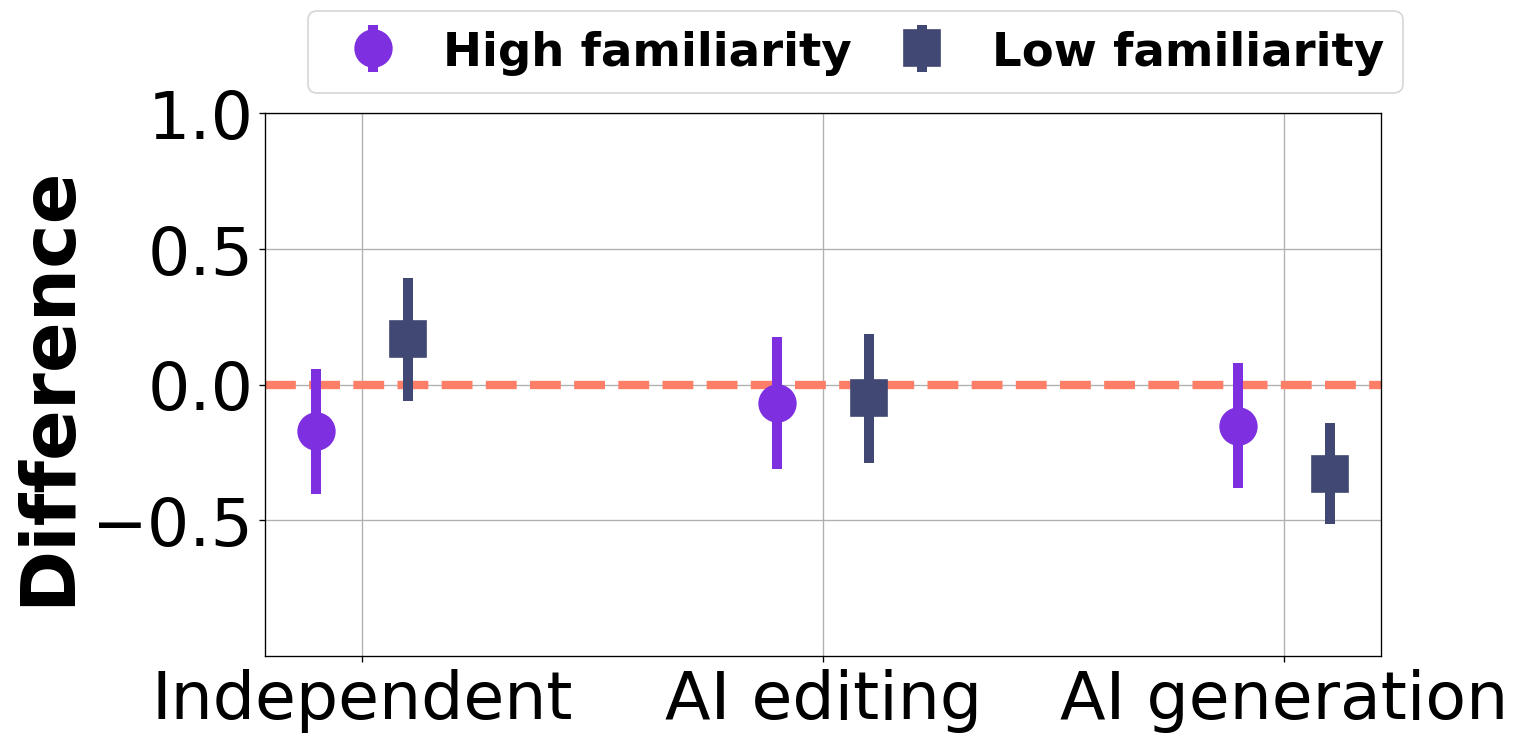}\label{fig:fam_organization_statement}}
  \hfill
  \subfloat[Creative story]{\includegraphics[width=0.24\textwidth]{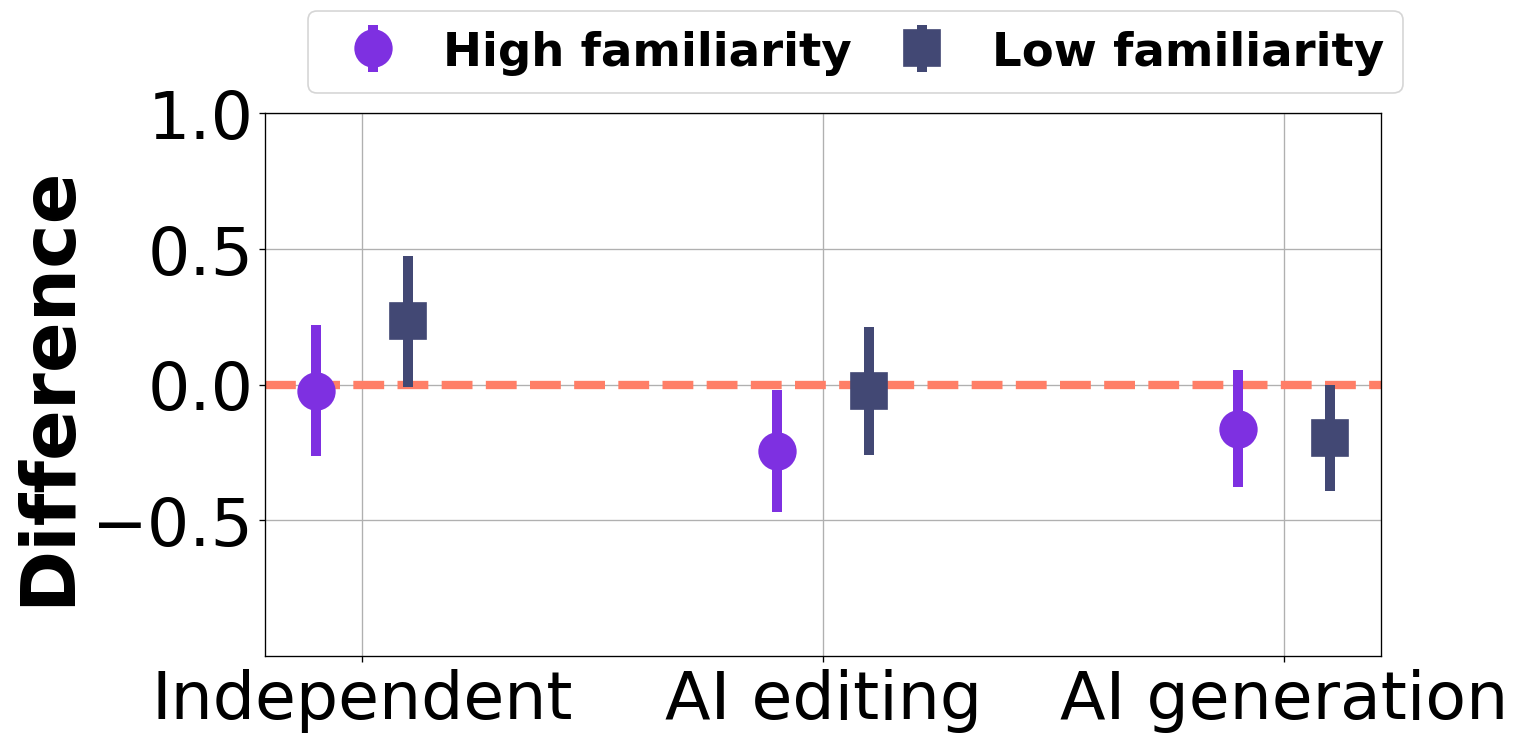}\label{fig:fam_organization_story}}
  \vspace{-5pt}
\caption{
The difference in an article's organization ratings received in the ``{\em Disclose}'' treatment and those received in the ``{\em Non-Disclose}'' treatment, when the ratings were provided by participants who had high or low {\em familiarity with ChatGPT}. Error bars represent the 95\% bootstrap confidence intervals of the rating difference. An interval below zero means the corresponding group of raters decrease their ratings when the use and type of AI assistance in the writing process was revealed to them. 
}\label{fam_organ}
  \vspace{-15pt}
\end{figure}

\begin{figure}[t]
  \centering
  \subfloat[Argumentative essay]{\includegraphics[width=0.24\textwidth]{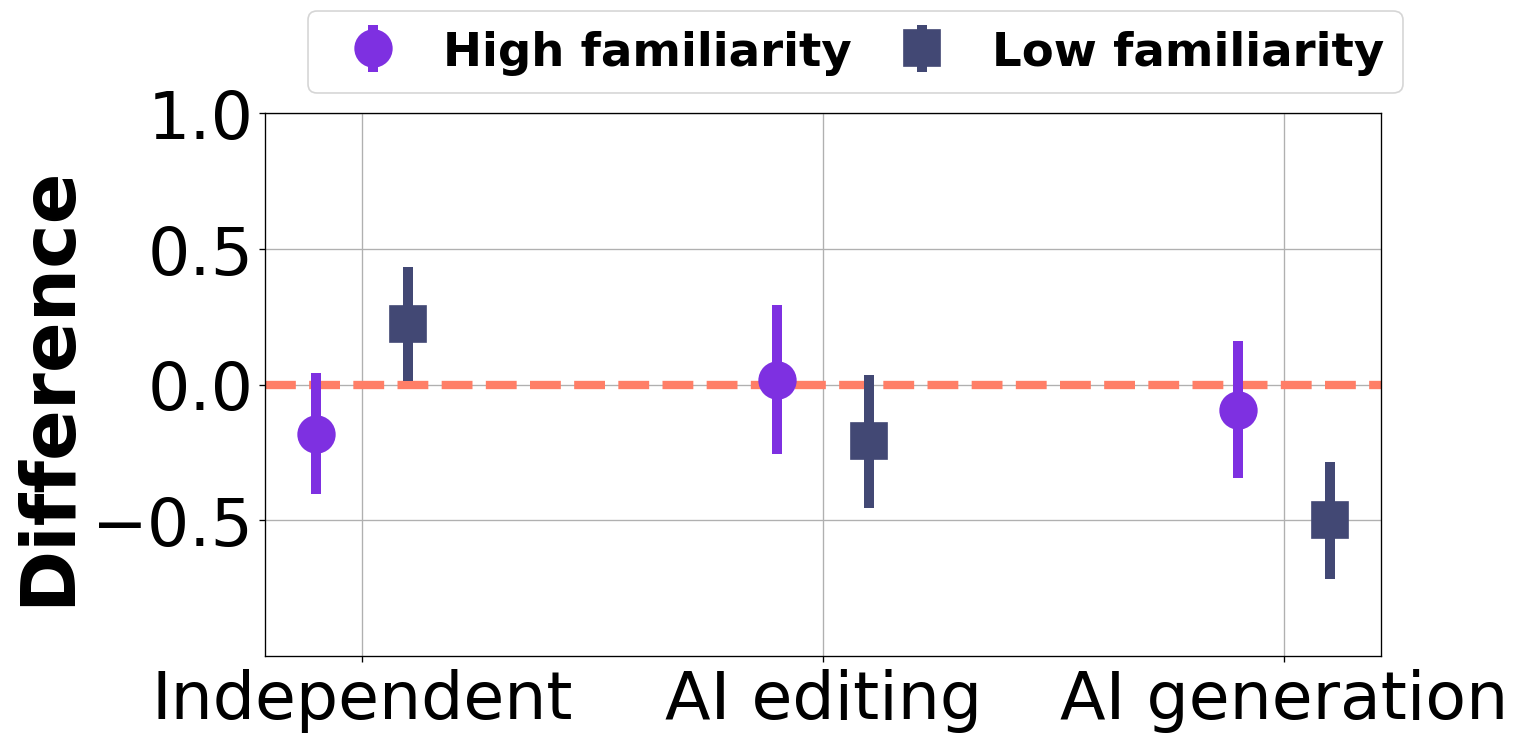}\label{fig:fam_origin_statement}}
  \hfill
  \subfloat[Creative story]{\includegraphics[width=0.24\textwidth]{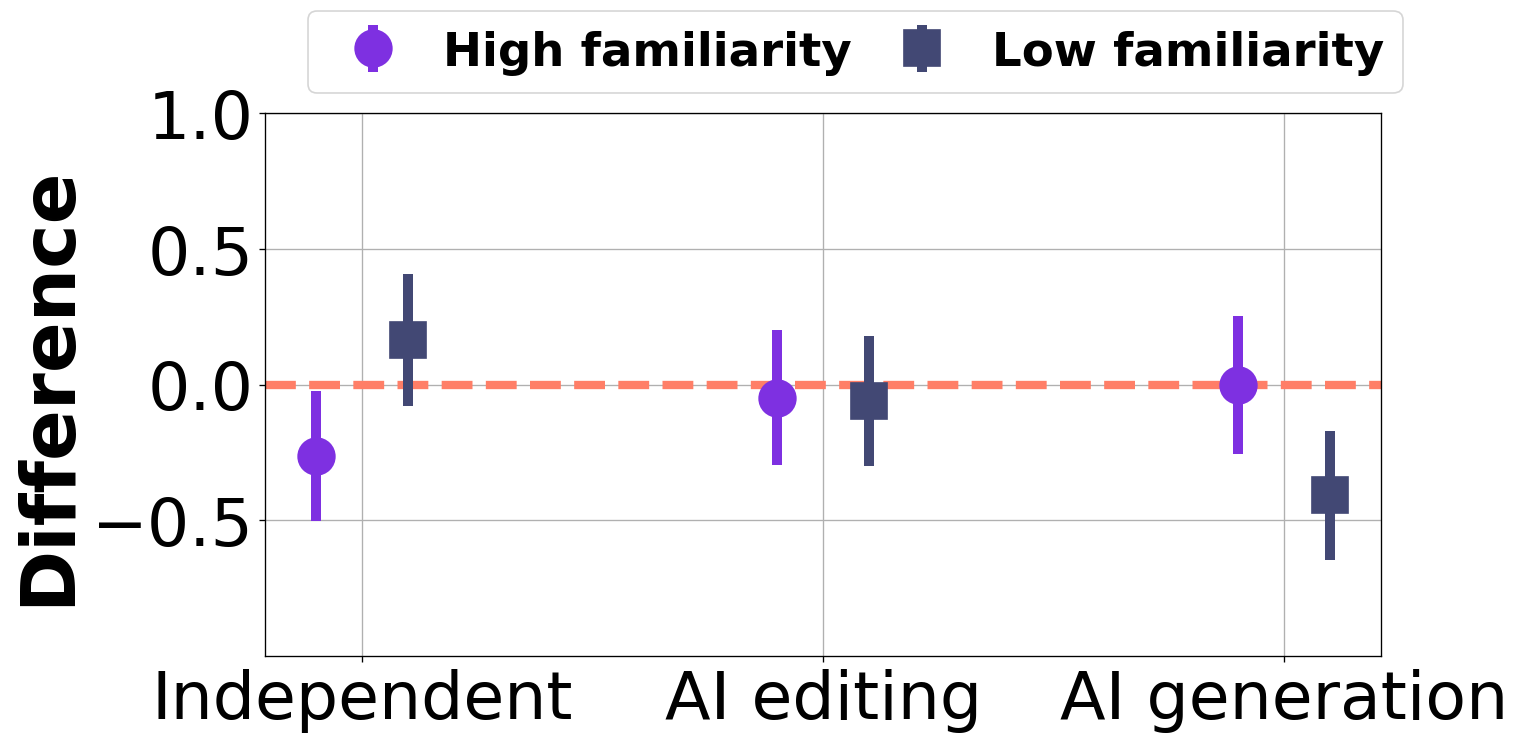}\label{fig:fam_origin_story}}
  \vspace{-5pt}
\caption{
The difference in an article's originality ratings received in the ``{\em Disclose}'' treatment and those received in the ``{\em Non-Disclose}'' treatment, when the ratings were provided by participants who had high or low {\em familiarity with ChatGPT}. Error bars represent the 95\% bootstrap confidence intervals of the rating difference. An interval below zero means the corresponding group of raters decrease their ratings when the use and type of AI assistance in the writing process was revealed to them. 
}\label{fam_origin}
  \vspace{-15pt}
\end{figure}

\begin{figure}[t]
  \centering
  \subfloat[Argumentative essay]{\includegraphics[width=0.24\textwidth]{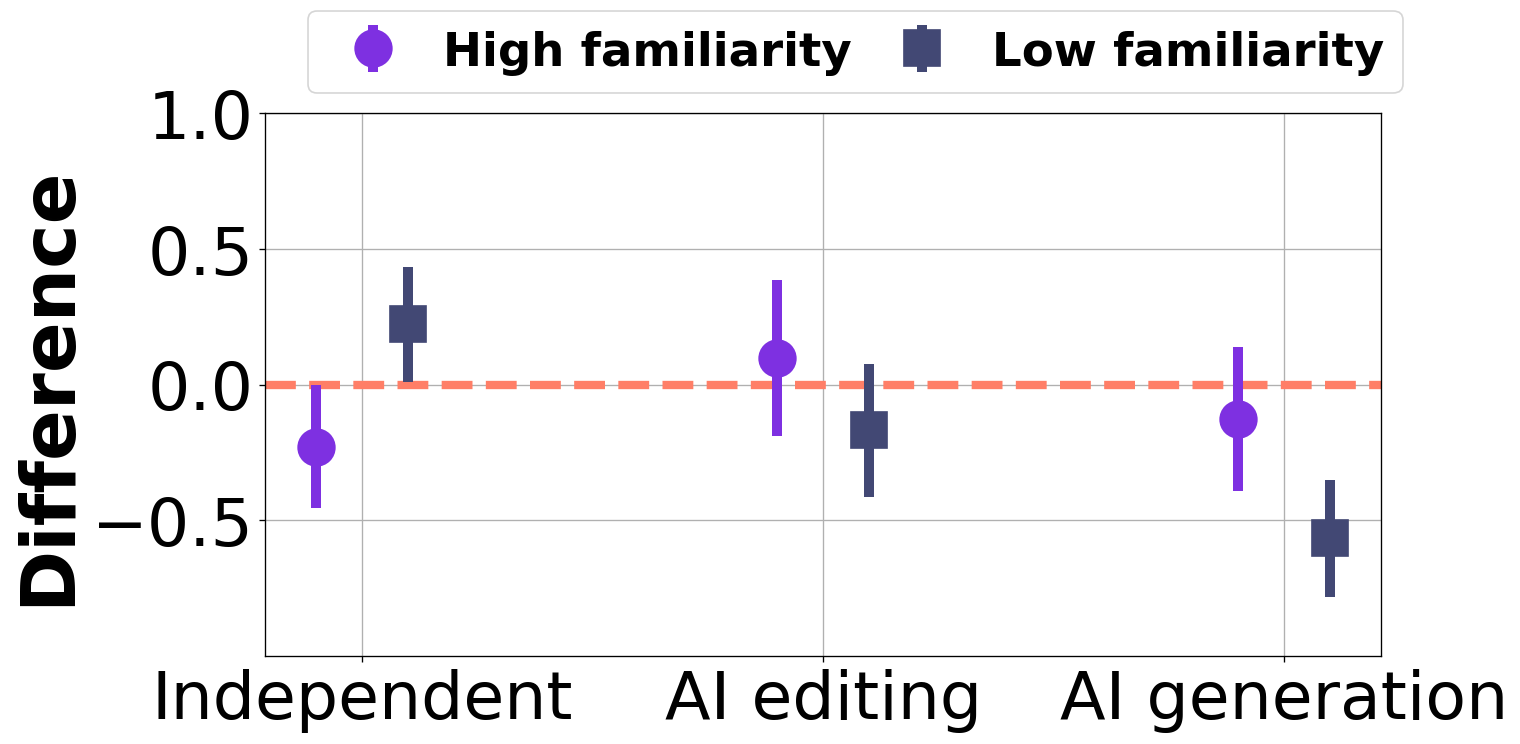}\label{fig:fam_creat_statement}}
  \hfill
  \subfloat[Creative story]{\includegraphics[width=0.24\textwidth]{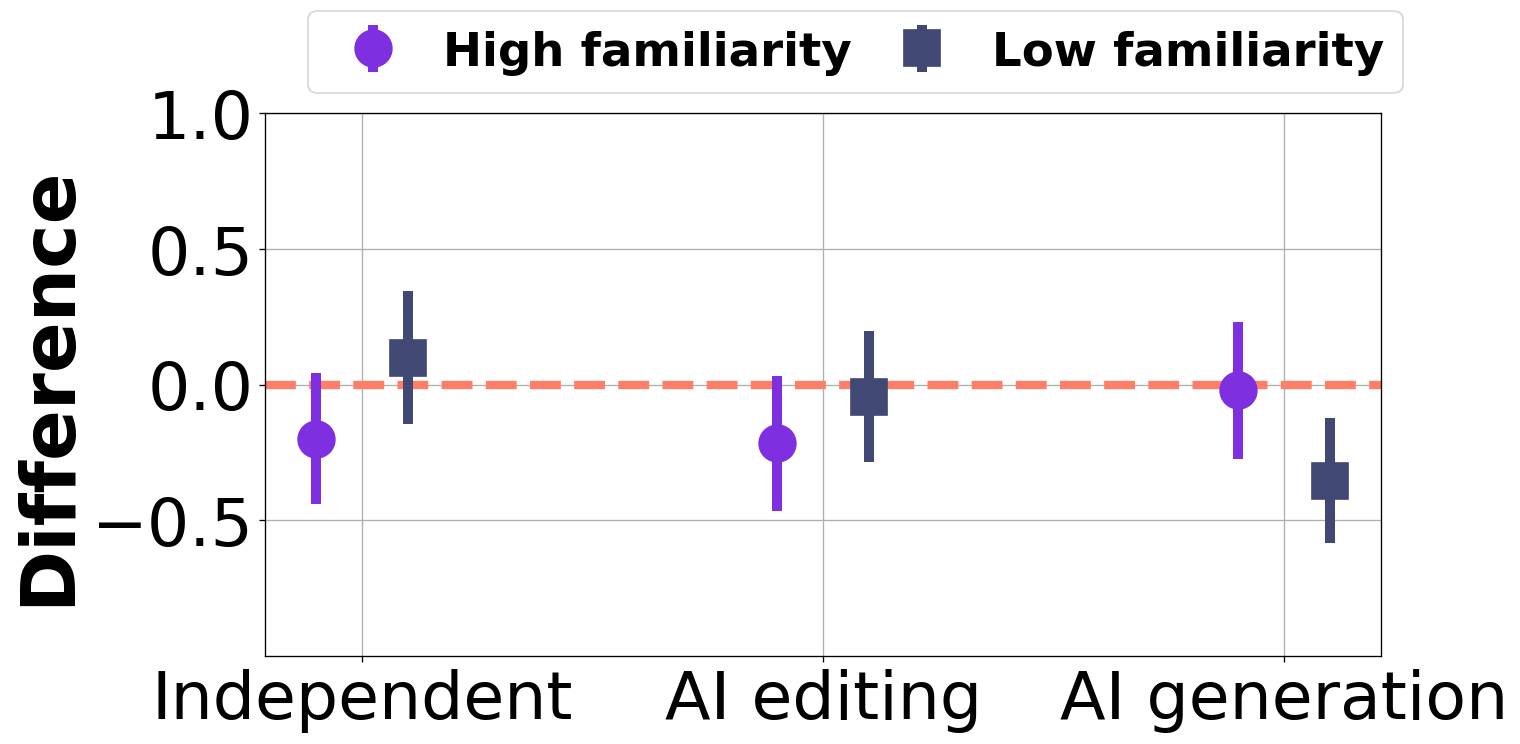}\label{fig:fam_creat_story}}
  \vspace{-5pt}
\caption{
The difference in an article's creativity ratings received in the ``{\em Disclose}'' treatment and those received in the ``{\em Non-Disclose}'' treatment, when the ratings were provided by participants who had high or low {\em familiarity with ChatGPT}. Error bars represent the 95\% bootstrap confidence intervals of the rating difference. An interval below zero means the corresponding group of raters decrease their ratings when the use and type of AI assistance in the writing process was revealed to them. 
}\label{fam_creat}
  \vspace{-15pt}
\end{figure}

\begin{figure}[t]
  \centering
  \subfloat[Argumentative essay]{\includegraphics[width=0.24\textwidth]{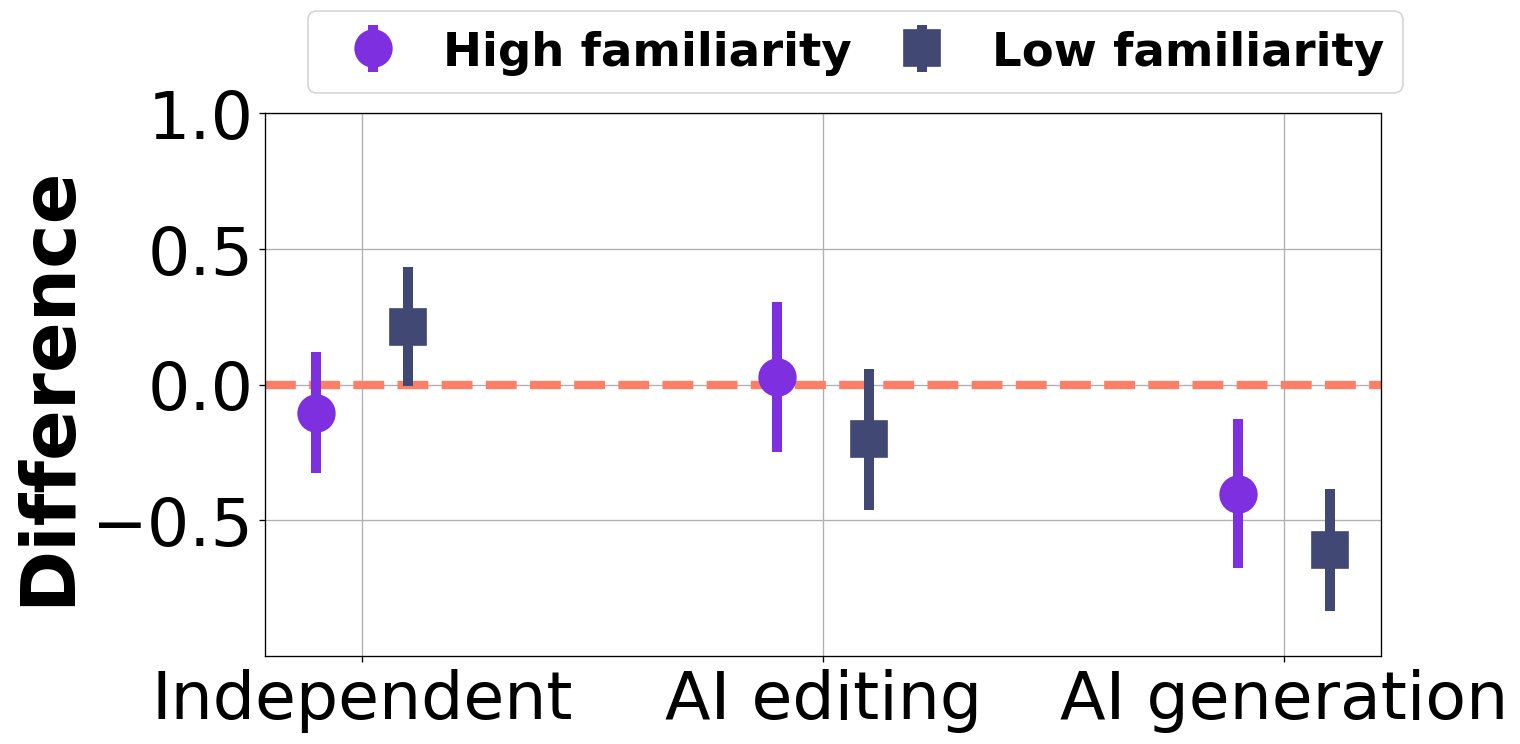}\label{fig:fam_creat_statement}}
  \hfill
  \subfloat[Creative story]{\includegraphics[width=0.24\textwidth]{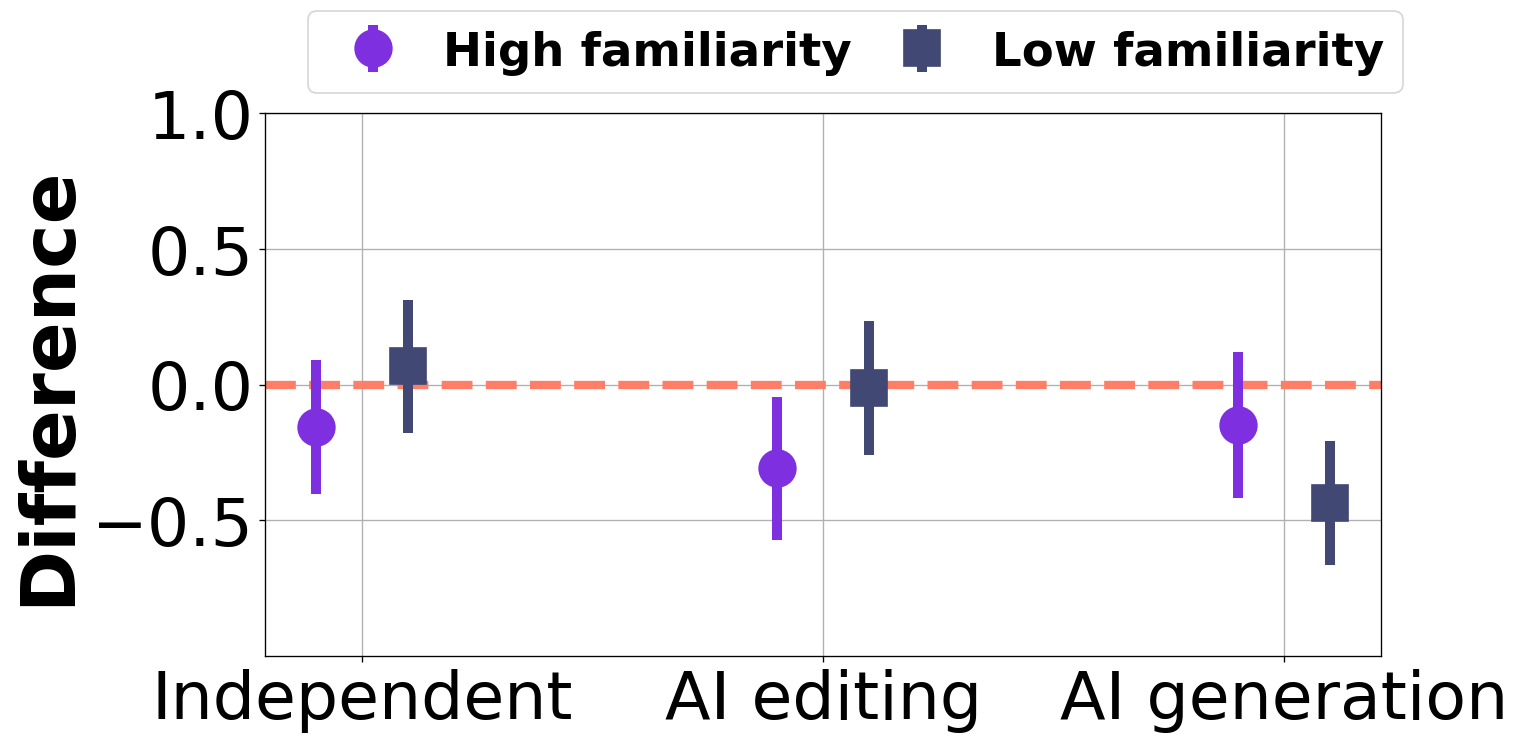}\label{fig:fam_creat_story}}
  \vspace{-5pt}
\caption{
The difference in an article's emotion authenticity ratings received in the ``{\em Disclose}'' treatment and those received in the ``{\em Non-Disclose}'' treatment, when the ratings were provided by participants who had high or low {\em familiarity with ChatGPT}. Error bars represent the 95\% bootstrap confidence intervals of the rating difference. An interval below zero means the corresponding group of raters decrease their ratings when the use and type of AI assistance in the writing process was revealed to them. 
}\label{fam_emotion}
  \vspace{-15pt}
\end{figure}

Figures~\ref{confidence_rec}--\ref{confidence_emotion} show the difference in an article's shortlisting rate, ratings of grammar and vocabulary, ratings of organization, ratings of originality, ratings of creativity, and ratings of emotion authenticity between the ``Disclose'' and 
the ``Non-Disclose'' treatments, when the ratings were provided by participants who had high or low confidence in their own writing skills. In general, participants with high confidence in writing appear to be influenced by the disclosure of AI assistance more in evaluating various aspects of an article than participants with low confidence in writing; they tend to decrease their willingness to shortlist an article or decrease their ratings on various aspects of the article, especially after knowing about the use of AI's content generation assistance during the writing process.

Figures~\ref{fam_rec}--\ref{fam_emotion} show the difference in an article's shortlisting rates, ratings of grammar and vocabulary, ratings of organization, ratings of originality, ratings of creativity, and ratings of emotion authenticity 
between the
``Disclose'' 
and 
``Non-disclose'' treatments, when the ratings were provided by participants who had high or low familiarity with ChatGPT. Compared to participants with high familiarity with ChatGPT, those with low familiarity with ChatGPT are more likely to decrease their willingness to shortlist an article and their ratings on various aspects of the article, if they become aware of that the author of the article utilizes ChatGPT's content generation assistance when writing the article.  

\section{Impacts of Disclosing AI Assistance on the
Ranking of Writing (Additional Result)}\label{rank_sm}

\begin{table*}
\centering
\caption{Within the top $\gamma \%$ of articles for the same writing task (ranked by articles' average overall quality ratings), the exact percentages of articles that were written in each of the three writing modes, with and without disclosing the use and type of AI assistance. }
\label{tab:proportion}
\resizebox{\linewidth}{!}{
\begin{tabular}{ccc:cc:cc:cc:cc} 
\hline
\multirow{2}{*}{Threshold (\%)} & \multicolumn{2}{c:}{10} & \multicolumn{2}{c:}{20} & \multicolumn{2}{c:}{30} & \multicolumn{2}{c:}{40} & \multicolumn{2}{c}{50}   \\ 
\cdashline{2-11}[1pt/1pt]
                                & Non-disclose & Disclose & Non-disclose & Disclose & Non-disclose & Disclose & Non-disclose & Disclose & Non-disclose & Disclose  \\ 
\hline
Independent                     & 12.1\%       & 30.1\%   & 21.3\%       & 35.3\%   & 22.3\%       & 34.1\%   & 24.8\%       & 31.5\%   & 28.3\%       & 30.1\%    \\
AI editing                      & 24.3\%       & 45.4\%   & 26.5\%       & 36.8\%   & 28.8\%       & 31.7\%   & 25.8\%       & 32.1\%   & 27.1\%       & 32.6\%    \\
AI generation                   & 63.6\%       & 24.5\%   & 52.2\%       & 27.9\%   & 48.9\%       & 34.2\%   & 49.4\%       & 36.3\%   & 44.6\%       & 37.3\%    \\
\hline
\end{tabular}
}
\end{table*}
Table~\ref{tab:proportion} reports the average percentages of articles  generated under the three writing modes for argumentative essay tasks whose overall quality ratings fall within the top $\gamma \%$ ($\gamma\in\{10, 20, \cdots, 50\}$) threshold,  when the use of AI assistance was or was not revealed to raters.

\begin{figure}[t]
  \centering
  \subfloat[Argumentative essay]{\includegraphics[width=0.24\textwidth]{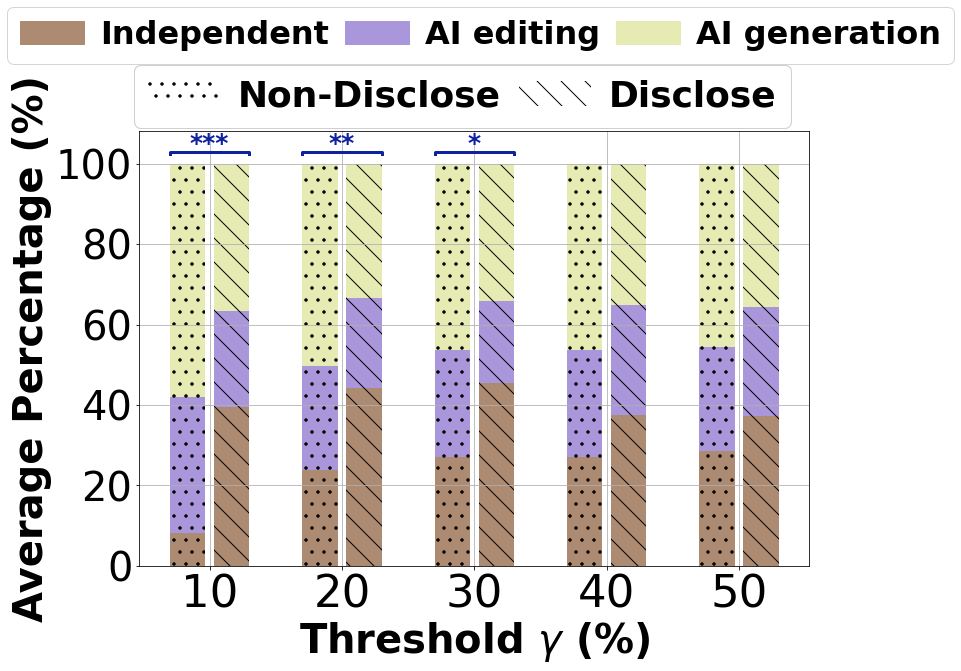}\label{fig:per_statement_rec}}
  \hfill
  \subfloat[Creative story]{\includegraphics[width=0.24\textwidth]{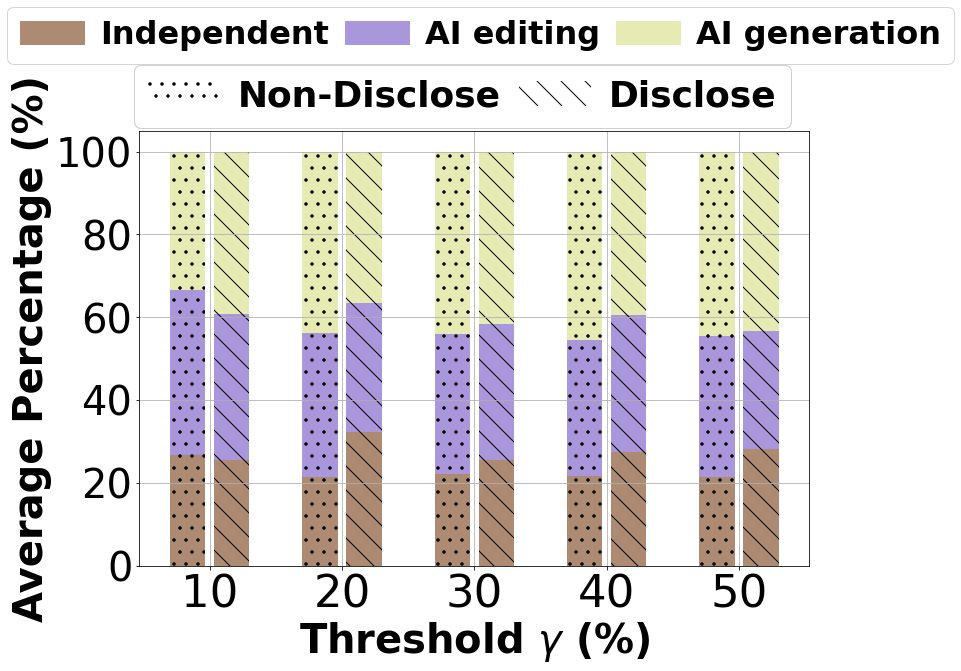}\label{fig:per_story_rec}}
  \vspace{-5pt}
\caption{Within the top $\gamma \%$ of the articles written about the same statement or prompt (determined by the probability for raters to shortlist the article), the percentages of articles that were written in each of the three writing modes,   when the use of AI assistance was or was not reveal to raters. $\textsuperscript{*}$, $\textsuperscript{**}$ and $\textsuperscript{***}$ denote the significance level of $0.05$, $0.01$, and $0.001$, respectively .
}\label{percentage_rec}
  \vspace{-15pt}
\end{figure}

Figure~\ref{percentage_rec} shows the percentages of articles that were produced from each of the three writing modes within the top $\gamma$\% of the articles written about the same statement or prompt (determined by the probability for raters to shortlist an article), when the use of AI assistance was or was not reveal to raters. We found that revealing the usage and type of AI assistance significantly decreases the proportions of highly-ranked articles that are generated with ChatGPT's content generation assistance for argumentative essays, while the impacts on the ranking of creative stories are minimal. 

\section{Impacts of Disclosing AI Assistance on Authorship Attribution
}\label{sec:authorship}

To understand the extent to which participants attribute authorship of the final articles to the human authors (i.e., crowd workers in our Phase 1 study) across the three writing modes, we presented the following statements in the exit survey of Phase 2 and asked participants to rate how much they agreed with each statement on a 5-point Likert scale from 1 (strongly disagree) to 5 (strongly agree) \cite{jago2023made}:
\squishlist
  \setlength\itemsep{0pt} 
    \setlength\parskip{0pt} 
    \setlength\parsep{0pt}  
    \item ``I see the crowd workers as responsible for those articles.'' 
    \item 
    ``I acknowledge the crowd workers as the creators of those articles.''
    \item 
    `` I give credit to the crowd workers for those articles.''
    \item 
    ``I see those articles as the product of the crowd workers’ efforts.''
\squishend
Based on participants' responses, we examine how the disclosing AI assistance in the writing process may change people's evaluation of human writers' ownership in their writing.

\begin{figure}[t]
\centering

\includegraphics[width=0.85\linewidth]{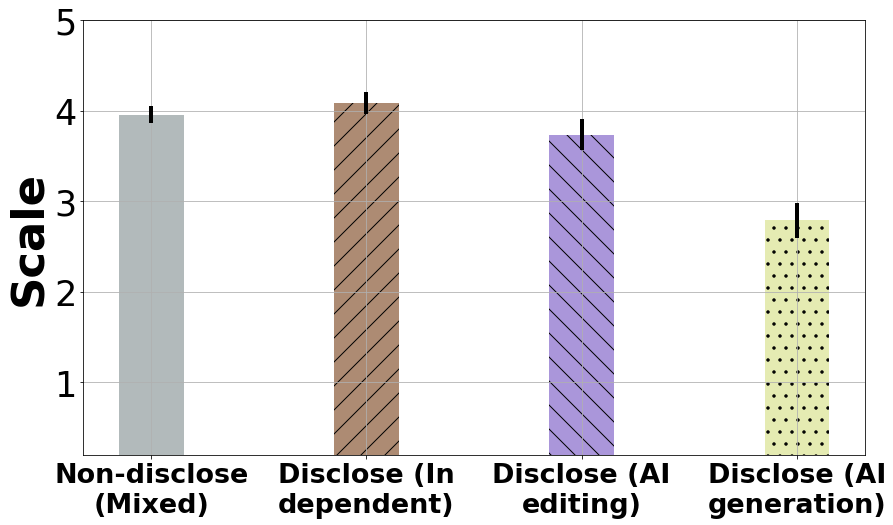}
\vspace{-5pt}
\caption{Comparing people's {\em authorship attribution} of writings to human writers under the ``{\em Non-Disclose}'' and ``{\em Disclose}'' treatments.
Error bars represent the 95\% confidence intervals of the means. 
}\label{ownership}
\end{figure}

Figure~\ref{ownership}  compares participants' {\em authorship attribution} when they were not informed about how the writing was generated (i.e., participants assigned to the ``{\em Non-Disclose}'' treatment) and when they were aware of the specific writing mode the authors took to produce the writing (i.e., participants assigned to the ``{\em Disclose}'' treatment). 
Note that in the ``{\em Non-Disclose}'' treatment, since the articles that participants evaluated were randomly selected from all the articles we collected in Phase 1 regardless of their writing modes, participants' attribution of human writers' authorship of the articles under this treatment was aggregated across the three writing modes. 
A one-way ANOVA test shows that there is a significant difference in participants' attribution of the authorship to human writers across different conditions ($p<0.001$). A post-hoc pairwise comparison further suggests that participants in the ``{\em Disclose}'' treatment who reviewed articles that were produced with AI's content generation assistance attributed a significantly lower level of authorship to the human writers compared to those in all other conditions ($p<0.001$ for all comparisons). 
Moreover, when the use and type of AI assistance were revealed, participants also attribute a lower level of ownership to human writers when they knew the writers used AI for editing purposes during the writing process. This attribution was significantly lower compared to when participants knew the writers completed the writing 
without any AI assistance ($p=0.016$).

\end{document}